\documentclass[journal]{IEEEtran}

\usepackage{xcolor}
\usepackage{tcolorbox}
\tcbuselibrary{breakable}
\usepackage{longtable}
\usepackage{booktabs}
\usepackage{amsmath}
\usepackage{orcidlink}
\usepackage[normalem]{ulem} 
\usepackage{tikz}
\usetikzlibrary{positioning, arrows.meta}

\usepackage[linesnumbered,algo2e,boxed]{algorithm2e}
\graphicspath{{figs/}}
\usepackage[figuresright]{rotating}
\usepackage{textcomp,subfigure,multirow,upgreek}
\usepackage{booktabs}
\usepackage{color,mdwlist}
\usepackage{tikz}
\usepackage[edges]{forest}
\definecolor{hidden-draw}{RGB}{20,68,106}
\definecolor{hidden-pink}{RGB}{255,245,247}

\makeatletter
\DeclareRobustCommand\onedot{\futurelet\@let@token\@onedot}
\def\@onedot{\ifx\@let@token.\else.\null\fi\xspace}

\makeatother

\usepackage{ragged2e}
\usepackage{tikz}
\usepackage{array}

\usepackage{tabularx}
\newcolumntype{Y}{>{\centering\arraybackslash}X}
\newcolumntype{Z}{>{\raggedleft\arraybackslash}X}

\usepackage{graphicx,fontawesome}
\graphicspath{{figs/}}

\usepackage{caption}
\usepackage{wrapfig}
\usepackage{stfloats}
\usepackage{enumitem}
\usepackage[misc]{ifsym}
\setitemize{noitemsep,topsep=0pt,parsep=0pt,partopsep=0pt}

\definecolor{colorfei}{HTML}{C3375A}

\definecolor{colorting}{HTML}{1a512e}

\definecolor{best}{RGB}{210, 235, 255}
\definecolor{second}{RGB}{255, 240, 210} 
\definecolor{hotPink}{RGB}{200, 40, 120}

\usepackage{float} 
\usepackage{makecell}
\DeclareCaptionFont{blue}{\color{blue}}

\ifCLASSINFOpdf
\else
\fi
\definecolor{mypink}{RGB}{255,20,147}

\begin{document}
%
\title{A Survey on Wi-Fi Sensing Generalizability: Taxonomy, Techniques, Datasets, and\\ Future Research Prospects}
%
%
%

\author{Fei Wang\orcidlink{0000-0002-0750-6990},~\IEEEmembership{Member,~IEEE,}
        Tingting Zhang\orcidlink{0009-0007-2157-4360},
        Wei Xi\orcidlink{0000-0001-9348-2982},~\IEEEmembership{Member,~IEEE,}
        Han Ding\orcidlink{0000-0002-5274-7988},~\IEEEmembership{Senior~Member,~IEEE,}\\
        Ge Wang\orcidlink{0000-0002-9058-8543}~\IEEEmembership{Member,~IEEE,}
        Di Zhang\orcidlink{0000-0002-2679-7535},
        Yuanhao Cui\orcidlink{0000-0001-6323-8559},~\IEEEmembership{Member,~IEEE,}
        Fan Liu\orcidlink{0000-0002-5299-9317},~\IEEEmembership{Senior~Member,~IEEE}\\
        Jinsong Han\orcidlink{0000-0001-5064-1955},~\IEEEmembership{Senior~Member,~IEEE},
        Jie Xu\orcidlink{0000-0002-4854-8839},~\IEEEmembership{Fellow,~IEEE,}
        Tony Xiao Han\orcidlink{0009-0007-8423-7345},~\IEEEmembership{Senior~Member,~IEEE}\\
        \textcolor{mypink}{Sensing Dataset Platform: \url{http://www.sdp8.org/}}
        \\ \textcolor{mypink}{Survey Page: \url{https://github.com/aiotgroup/awesome-wireless-sensing-generalization}}
\thanks{
Accepted by IEEE Communications Surveys \& Tutorials 2026\\
This work was supported in part by the National Natural Science Foundation of China under Grant 62572383, Grant 62372400, Grant 62372365, Grant 62472346, Grant 62471424, and Grant U25A20390; in part by China Association
for Science and Technology (CAST) Young Talent Support Program under Grant YESS2023066; and in part by Shenzhen Fundamental Research Program under Grant JCYJ20250604141209012.
}
\thanks{Fei Wang and Tingting Zhang are with the School of Software Engineering and the State Key Laboratory of Human-Machine Hybrid Augmented Intelligence, Xi'an Jiaotong University, Xi'an 710049, China (emails: feynmanw@xjtu.edu.cn; tt\_zhang@stu.xjtu.edu.cn).}
\thanks{Wei Xi, Han Ding, and Ge Wang are with the School of Computer Science and Technology, Xi'an Jiaotong University, Xi'an 710049, China (emails: dinghan@xjtu.edu.cn; gewang@xjtu.edu.cn; xiwei@xjtu.edu.cn).} 
\thanks{Di Zhang and Yuanhao Cui are with Information and Communication Engineering,Beijing University of Posts and Telecommunications, Beijing 100876, China (e-mails: amandazhang@bupt.edu.cn; cuiyuanhao@bupt.edu.cn).}
\thanks{Fan Liu is with the School of Information Science and Engineering, Southeast University, Nanjing 210096, China (email: fan.liu@seu.edu.cn). }
\thanks{Jinsong Han is with the College of Computer Science and Technology, Zhejiang University, Hangzhou 310058, China (email: hanjinsong@zju.edu.cn). }
\thanks{Jie Xu is with the School of Science and Engineering (SSE), the Shenzhen Future Network of Intelligence Institute (FNii-Shenzhen), and the Guangdong Provincial Key Laboratory of Future Networks of Intelligence, The Chinese University of Hong Kong, Shenzhen, Guangdong 518172, China (e-mail: xujie@cuhk.edu.cn). }
\thanks{Tony Xiao Han is with the Wireless Technology Lab, Huawei Technologies Co., Ltd., Shenzhen 518129, China (e-mail: tony.hanxiao@huawei.com)}
}

\markboth{IEEE Communications Surveys \& Tutorials,~Vol.~00, No.~0, February~2026}%
{Fei Wang \MakeLowercase{\textit{et al.}}: A Survey on Wi-Fi Sensing Generalizability: Taxonomy, Techniques, Datasets, and Future Research Prospects}
%


\maketitle

\begin{abstract} Wi-Fi sensing has emerged as a powerful non-intrusive technology for recognizing human activities, monitoring vital signs, and enabling context-aware applications using commercial wireless devices. However, the performance of Wi-Fi sensing often degrades when applied to new users, devices, or environments due to significant domain shifts. To address this challenge, researchers have proposed a wide range of generalization techniques aimed at enhancing the robustness and adaptability of Wi-Fi sensing systems. In this survey, we provide a comprehensive and structured review of over 200 papers published since 2015, categorizing them according to the Wi-Fi sensing pipeline:  experimental setup, signal preprocessing, feature learning, and model deployment. We analyze key techniques, including signal preprocessing, domain adaptation, meta-learning, metric learning, data augmentation, cross-modal alignment, federated learning, and continual learning. Furthermore, we summarize publicly available datasets across various tasks, such as activity recognition, user identification, indoor localization, and pose estimation, and provide insights into their domain diversity. We also discuss emerging trends and future directions, including large-scale pretraining, integration with multimodal foundation models, and continual deployment. To foster community collaboration, we introduce the Sensing Dataset Platform~(SDP) (\textcolor{mypink}{\url{http://www.sdp8.org/}}) for sharing datasets and models. This survey aims to serve as a valuable reference and practical guide for researchers and practitioners dedicated to improving the generalizability of Wi-Fi sensing systems. Notably, while this paper focuses on Wi-Fi sensing, it is important to emphasize that the methodologies discussed in the feature learning and model deployment stages, e.g., domain alignment, metric learning, meta-learning, federated learning, and continual learning, are equally applicable to a broader range of wireless sensing generalization challenges, such as those encountered in millimeter radar-based human sensing. Given the rapid evolution of this field, we will continuously maintain and update relevant resources at \textcolor{mypink}{\url{https://github.com/aiotgroup/awesome-wireless-sensing-generalization}}.
\end{abstract}


%
\IEEEpeerreviewmaketitle

\section{Introduction}\label{sec:introduction}

\begin{figure}[t]
    \centering
    \includegraphics[width=1\linewidth]{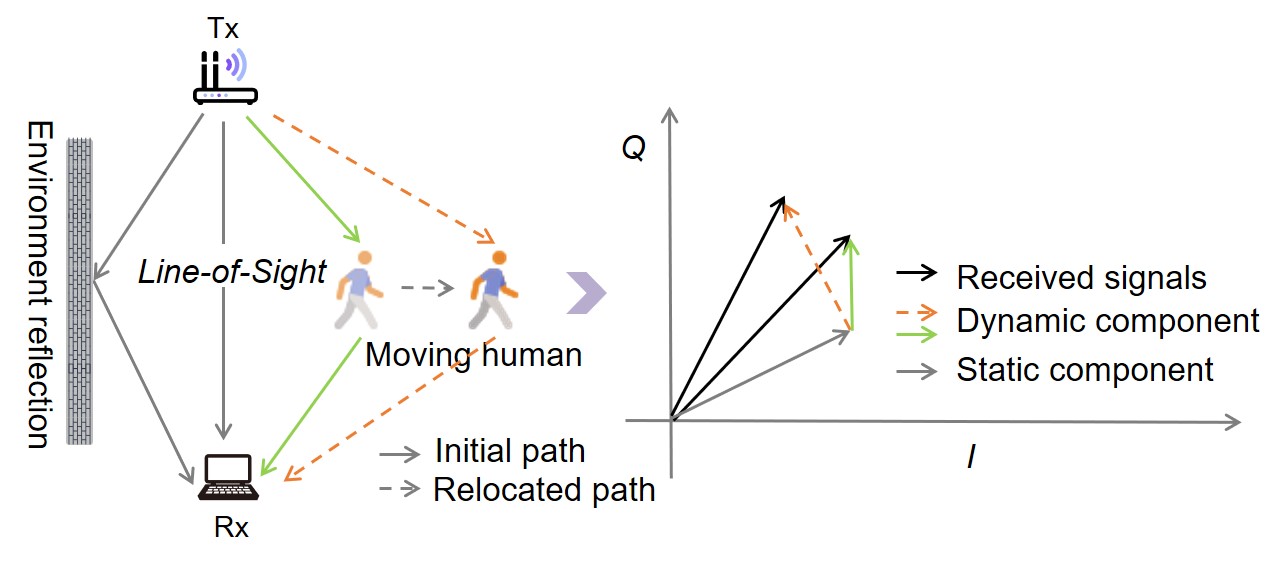}
    \caption{Wi-Fi signals emitted from a transmitter propagate through both static environmental structures and dynamic human bodies before reaching the receiver. Human presence and movements alter the signal propagation paths, leading to measurable changes in the received signal. These variations can be leveraged for Wi-Fi sensing to interpret human movements.}
    \label{fig:wifi-sensing-principle}
\end{figure}

Wi-Fi sensing has emerged as a transformative technology that leverages ubiquitous Wi-Fi infrastructures for tasks beyond communication. By analyzing the variations in Wi-Fi signals such as Channel State Information (CSI) and Received Signal Strength Indicator (RSSI), researchers have demonstrated its potential in a wide range of applications, including 
activity recognition
~\cite{jiang2018towards,
wang2015understanding,
tan2019multitrack,
zhu2017r,
wang2014eyes,
chen2018wifi,
adib2013see}, 
gesture recognition~\cite{li2016csi,
pu2013whole,
li2016wifinger,
ali2015keystroke,
venkatnarayan2018multi,
abdelnasser2015wigest,
ma2018signfi,
wang2019joint,sun2015widraw},
indoor localization~\cite{li2016dynamic,
xie2015precise,
vasisht2016decimeter,
qian2017widar,
li2017indotrack,
xie2019md,
qian2018widar2,
kotaru2015spotfi}, 
user authentication and identification~\cite{korany2019xmodal,
wang2019continuous,
wang2019wipin,
shi2017smart,
zeng2016wiwho,
2-157zhang2020gaitid,
wang2016gait},
health monitoring~\cite{zeng2020multisense,
wang2017tensorbeat,
zeng2018fullbreathe,
wang2016human,
zheng2016smokey,
zeng2019farsense}, 
fallen detection~\cite{wang2016rt,
palipana2018falldefi,
wang2016wifall,
ding2020wifi,
hu2021defall},
pose estimation~\cite{jiang2020towards,
wang2019can,
2-154wang2019person,
ren2022gopose,
wang2024multi,
wang2022wi,
qiansparse,
yan2024person},
crowd counting~\cite{zou2017freecount,
zou2018device,
choi2022wi,
jiang2023pa,
xi2014electronic}, and more. Many survey papers have summarized this thriving research field, reviewing existing studies from perspectives such as methods and tasks~\cite{tan2022commodity,
hernandez2022wifi,
zhang2020device,
ma2019wifi,
liu2019wireless,
li2021deep,
yousefi2017survey}.



While Wi-Fi sensing has shown great promise across a variety of applications, its effectiveness fundamentally relies on how wireless signals interact with the surrounding environment and the human body. Fig.~\ref{fig:wifi-sensing-principle} illustrates how human presence and motion influence the propagation of Wi-Fi signals. As shown in the figure, Wi-Fi devices are typically equipped with multiple transmit and receive antennas. Transmit antennas emit radio signals that propagate through the environment and reach the receive antennas via different paths, including line-of-sight, reflections from walls and furniture, and reflections from the human body. The presence and motion of a human subject alter the signal correlations on time, frequency or space, resulting in changes to key signal properties such as Received Signal Strength (RSS), Angle of Arrival (AoA), and Time of Flight (ToF). By analyzing these signal changes, it is possible to infer meaningful information about human activities. Both handcrafted feature-based methods and data-driven deep learning models have been developed to map variations in Wi-Fi signals to human presence, location, motion, respiration, and more, thus enabling a wide range of Wi-Fi sensing capabilities.

\begin{figure*}[t]
    \centering
    \subfigure[Device heterogeneity]{
        \begin{minipage}[t]{0.26\linewidth}
            \centering
            \includegraphics[width=\linewidth]{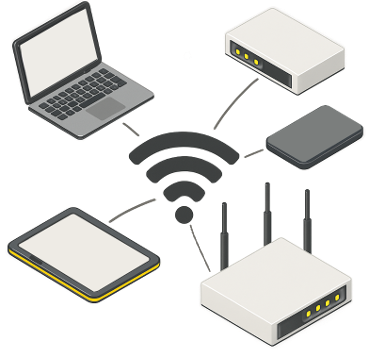}
        \end{minipage}
    }%
    \hfill
    \subfigure[Human body shape diversity]{
        \begin{minipage}[t]{0.41\linewidth}
            \centering
            \includegraphics[width=\linewidth]{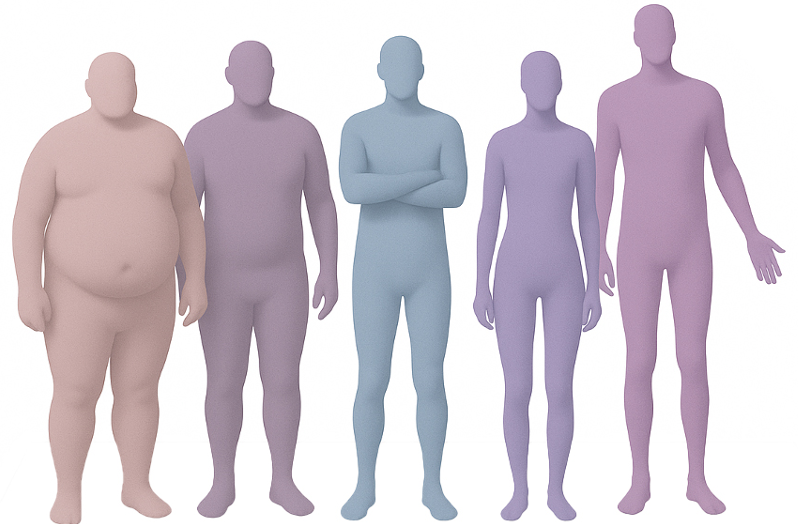}
        \end{minipage}
    }%
    \hfill
    \subfigure[Environment diversity]{
        \begin{minipage}[t]{0.27\linewidth}
            \centering
            \includegraphics[width=\linewidth]{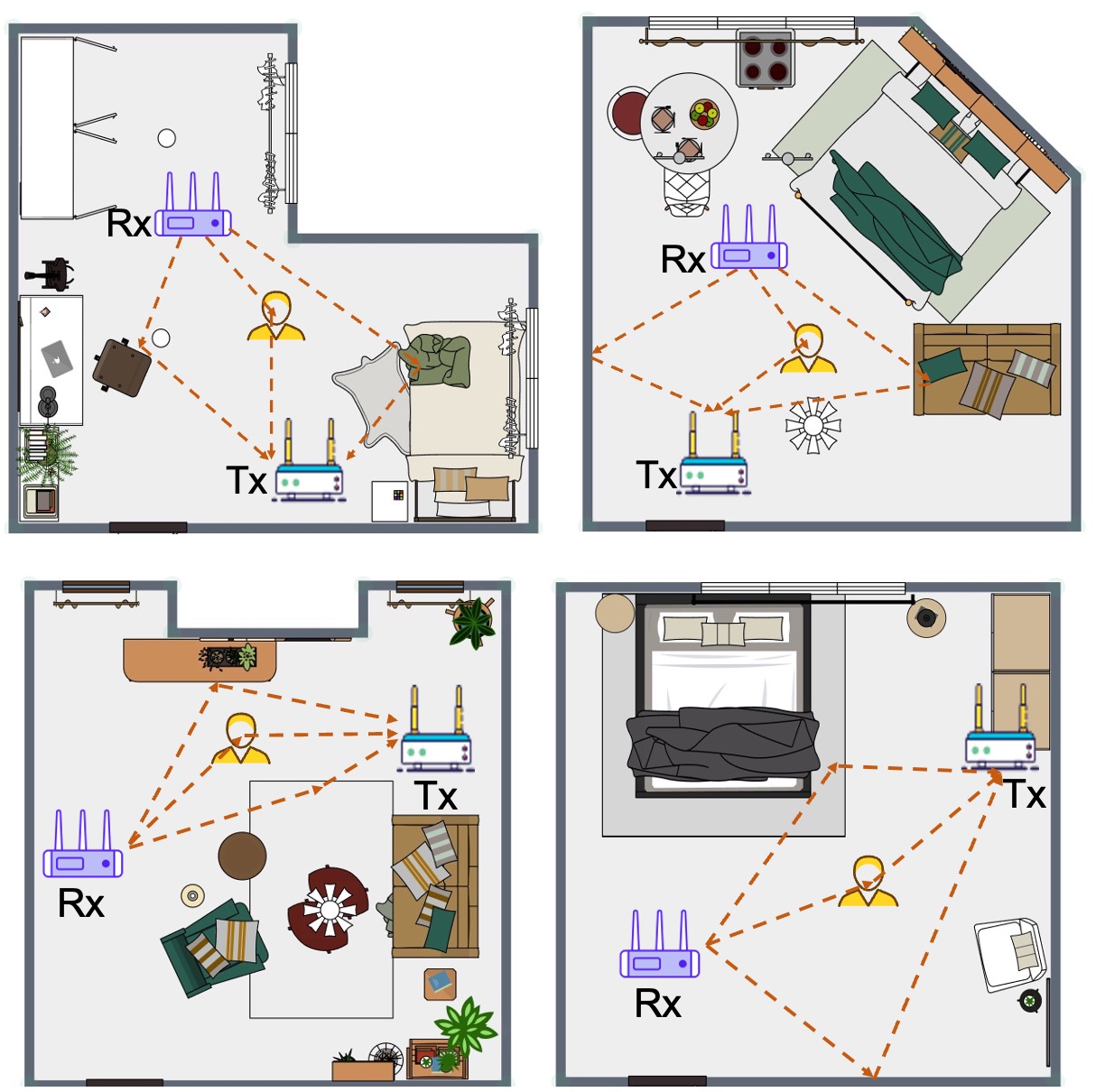}
        \end{minipage}
    }%
    \caption{Wi-Fi sensing generalizability is primarily hindered by three key factors, i.e., device heterogeneity, human body diversity, and environmental diversity.}
    \label{fig:wifi-sensing-diversity}
\end{figure*}

\begin{figure*}[t]
    \centering
    \includegraphics[width=1\textwidth]{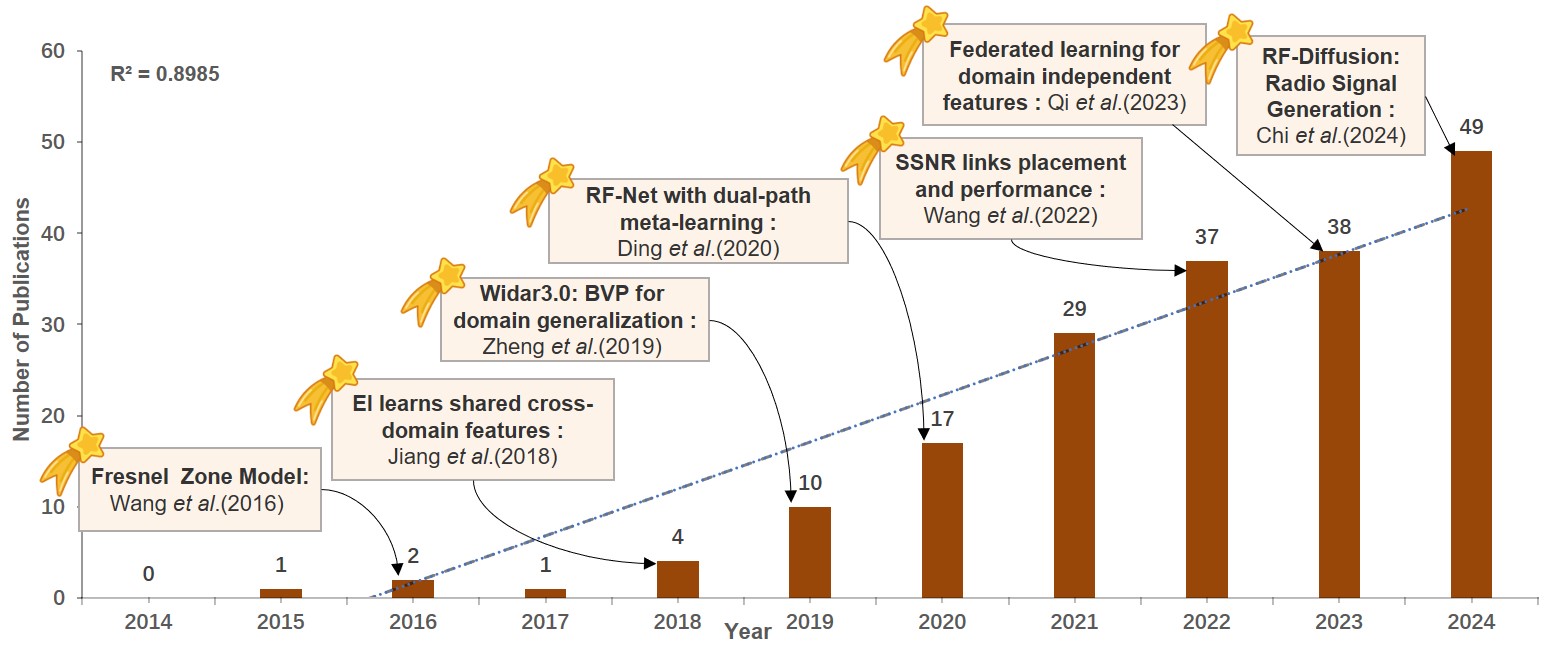}
    \caption{Growth of research in Wi-Fi sensing generalizability: from a handful of studies between year of 2015 and 2018 to a surge of publications since 2019. We employed a linear regression to fit the growth trend. In the figure, the $R^2$ score indicates the correlation between the estimated results and the ground truth, where a value closer to 1 denotes a higher degree of model fitting. (We selected one representative paper per year based on the highest citation count, ensuring each choice employs a technical approach distinct from those featured in previous years.) }
    \label{fig:wifi-sensing-over-years}
\end{figure*}

However, one critical challenge remains unresolved: Wi-Fi sensing generalization---the overall capability of Wi-Fi sensing systems to perform effective inference in unseen domains, such as new users, devices, or environments. To ensure clarity, the related term Wi-Fi sensing generalizability in this survey denotes the inherent property of a Wi-Fi sensing system design that enables such adaptability.

This performance gap between research prototypes and practical systems arises from the inherent variability in Wi-Fi sensing conditions, which introduces considerable domain shifts across deployment scenarios. As illustrated in Fig.~\ref{fig:wifi-sensing-diversity}, generalization is primarily hindered by three key factors, including device heterogeneity, human body diversity, and environmental diversity. First, device heterogeneity introduces substantial signal variation. Wi-Fi signals originate from a wide range of devices, including routers, laptops, and mobile phones, that differ in protocol versions, operating channels, bandwidth, antenna design, physical orientation, and chipset configurations. These variations affect the measured CSI and RSS, often rendering models trained on one device ineffective on another. Second, human body diversity introduces non-negligible discrepancies in signal interactions. Variations in height, weight, body composition, and even clothing materials can alter the propagation and absorption of Wi-Fi signals. As a result, two individuals performing the same action at the same location may produce significantly different signal patterns, posing challenges to consistent activity recognition. Third, environmental diversity in indoor scenes further exacerbates generalization difficulties. Differences in room layouts, wall materials, furniture placement, and device positions reshape the signal propagation paths and multipath profiles, leading to domain-specific variations in signal features such as RSS, AoA, and ToF. These changes severely limit the Wi-Fi sensing systems from generalizing beyond their training conditions.

\definecolor{hidden-draw}{RGB}{177, 177, 177}
\tikzstyle{my-box}=[
    rectangle,
    draw=hidden-draw,
    rounded corners,
    text opacity=1,
    minimum height=1.5em,
    minimum width=5em,
    inner sep=2pt,
    align=center,
    fill opacity=.5,
    line width=0.8pt,
]
\tikzstyle{leaf}=[my-box, minimum height=1.5em,
    fill=hidden-pink!80, text=black, align=left,font=\footnotesize,
    inner xsep=2pt,
    inner ysep=4pt,
    line width=0.8pt,
]

\begin{figure*}[!th]
    \centering
    \resizebox{0.96\textwidth}{!}{
        \begin{forest}
            forked edges,
            for tree={
                grow=east,
                reversed=true,
                anchor=base west,
                parent anchor=east,
                child anchor=west,
                base=center,
                font=\large,
                rectangle,
                draw=hidden-draw,
                rounded corners,
                align=left,
                minimum width=4em,
                edge+={darkgray, line width=1pt},
                s sep=3pt,
                inner xsep=2pt,
                inner ysep=3pt,
                line width=0.8pt,
                ver/.style={rotate=90, child anchor=north, parent anchor=south, anchor=center},
            },
            where level=1{text width=6em,align=left,fit=band,font=\normalsize,}{},
            where level=2{text width=8.2em,align=left,fit=band,font=\small,}{},
            where level=3{text width=9em,font=\small,}{},
            where level=4{text width=8em, font=\small,}{},
            [
                Wi-Fi Human Sensing  Generalization Categories, ver
                 [
                    Experimental \\ Setup \\ (\S \ref{sec:hardware_Setup})
                    , fill=orange!5
                    [
                        Distributed Antennas \\ (\S \ref{sec:distributed-antennas}), fill=orange!10,text width=9em
                        [
                                WiNDR~\cite{2-10qin2024direction}{, } WiCross~\cite{2-78qin2023cross}{, } WiSDAR~\cite{2-11wang2018spatial}
                                , leaf,fill=orange!10, text width=42em
                        ]
                    ]
                    [
                       Distributed Devices \\ (\S \ref{sec:distributed-devices}),fill=orange!10,text width=9em
                        [
                           ReWiS~\cite{2-26bahadori2022rewis}{, } Widar 3.0~\cite{1-45zheng2019zero}{, } OneFi~\cite{1-44xiao2021onefi}{, } WiTraj~\cite{2-147wu2021witraj}
                                , leaf,fill=orange!10, text width=42em
                        ] 
                    ]
                    [
                       Scaling Up Dataset \\ (\S \ref{sec:scaling-up-dataset}),fill=orange!10,,text width=9em
                        [
                           XRF55~\cite{wang2024xrf55}{, } Widar 3.0~\cite{1-45zheng2019zero}{, } Zhang et al.\cite{2-91zhang2023csi}{, }MM-Fi\cite{yang2024mm}{, } CSI-Bench~\cite{zhu2025csi}
                                , leaf,fill=orange!10, text width=42em
                        ] 
                    ]
                ]
                [
                    Signal \\Preprocessing \\ (\S \ref{sec:signal_preprocessing}), fill=violet!5
                    [
                        Signal Analysis   \\ Methods  (\S \ref{sec:signal-analysis-methods}), fill=violet!10,text width=9em
                        [
                                WiHand~\cite{2-138lu2019towards}{, } CLAR~\cite{2-32zhou2022device}{, } Gait-Enhance\cite{2-79yang2023gait}{, } Zhuo et al.\cite{2-74zhuo2023position}{, } 
                                Abuhoureyah et al.~\cite{2-158abuhoureyah2024multi}, leaf, fill=violet!15, text width=42em
                        ]
                    ]
                    [
                       Advanced Handcrafted \\ Indicators   (\S \ref{sec:advanced-wifi-indicators}), fill=violet!10, text width=9em
                        [
                                WiHGR~\cite{2-12meng2021wihgr}{, } PhaseAnti~\cite{2-41huang2021phaseanti}{, } DPSense~\cite{2-102gao2022towards}{, }HandGest~\cite{2-149zhang2022handgest}{, }
                                Yu et.al.~\cite{2-105yu2023towards}{, }AF-ACT~\cite{2-7zhang2022csi-based}{, } RoSeFi~\cite{2-43peng2023rosefi}{, }\\ Wang et.al.~\cite{2-110wang2024understanding}{, }WiGesFree\cite{2-17ding2024robust}{, }LiKey~\cite{2-82peng2024likey}{, }
                                Bu et.al.~\cite{2-156bu2018wi}{, }
                                Zhang et.al.~\cite{2-159zhang2023enhancing}
                                , leaf, fill=violet!15, text width=42em
                        ]
                    ]
                    [
                       Motion Indicators  \\   (\S \ref{sec:motion-indicators}), fill=violet!10,text width=9em
                        [
                                Sharp~\cite{2-163meneghello2022sharp}{, } OneSense~\cite{1-48zhao2024one}{, } Yin et al.\cite{2-98yin2021towards}{, } Wi-learner~\cite{2-89feng2022wi}{, } Niu et al.~\cite{1-43niu2021understanding}{, } and LAGER~\cite{1-32Chen_LAGER}{, }WiLife~\cite{2-115liwilife}{, }\\ Mini-Batch Alignment~\cite{2-66van2023mini,2-70van2022insights}{, } WiTraj~\cite{2-147wu2021witraj}{, } Widar 3.0~\cite{1-45zheng2019zero}. Zhang et al.~\cite{2-100zhang2023unsupervised}{, } WiTransformer~\cite{2-65yang2023witransformer}{, } \\Shi et al.~\cite{2-99shi2023location}{, } Bulugu~\cite{2-86bulugu2023gesture}. GaitID~\cite{2-157zhang2020gaitid}{, } OneFi~\cite{1-44xiao2021onefi}{, } XFall~\cite{2-117chi2024xfall}{, } WiGesture~\cite{2-124gao2021towards}{, } Shi et al.~\cite{2-142shi2020towards}{, }\\ Wang et al.~\cite{2-46wang2022position}{, } Widir~\cite{wu2016widir}{, } Wang et al.~\cite{wang2016human}{, } WiNDR\cite{2-10qin2024direction}{, } WiDIGR~\cite{2-23zhang2019widigr}{, } Wi-PIGR~\cite{2-27zhang2021wi}{, }\\ WiDetect~\cite{2-148zhang2019widetect}{, } Zhu et al.~\cite{2-63zhu2024wifi}, leaf, fill=violet!15, text width=42em
                        ]
                    ]
                    [
                      Angle Indicators \\ (\S \ref{sec:aoa-indicators}), fill=violet!10,text width=9em
                        [
                                 Li et al.~\cite{2-13li2020location}{, }EasyTrack~\cite{2-111han2023position}{, }
                                 3D-ID~\cite{2-76ren2022robust}, leaf, fill=violet!15, text width=42em
                        ]
                    ] 
                ]
                [                
                    Feature \\Learning \\(\S \ref{sec:Feature Learning Stage}),fill=yellow!5
                    [
                        Domain Alignment \\(\S \ref{sec:domain-alignment}),fill=yellow!10,text width=9em
                        [
                            Domain Alignment with \\ Domain Discriminator,fill=yellow!10
                                [
                                    WiHARAN~\cite{1-17wang2021environment}{, 
                                    }DA-HAR~\cite{2-69sheng2023har}{, 
                                    }Zhang et al.~\cite{1-5zhang2021privacy}{, 
                                    }i-Sample~\cite{1-18zhou2024sample}{, 
                                    }JADA~\cite{2-2zou2018joint}\\CrossGR~\cite{1-9li2021crossgr}{, 
                                    }Wi-Adaptor~\cite{2-30zhang2021wi}
                                    , leaf,fill=yellow!15, text width=31.5em
                                ]
                        ]
                        [
                            Domain Alignment with \\ Domain Classifier,fill=yellow!10
                            [
                                EI~\cite{jiang2018towards}{, }Khattak~\cite{2-72khattak2022cross}{, }WiCAR~\cite{1-11wang2019wicar}{, }GESFI~\cite{2-168zhang2024objective}{, }ADA~\cite{2-55zinys2021domain}{, }
                                        DATTA~\cite{2-155strohmayer2024datta}{, }WiSR~\cite{1-29liu2023wisr}\\
                                        Yin et al.\cite{2-98yin2021towards}{, }
                                        INDG-Fi~\cite{2-48yang2024domain}{, }
                                        WiCross~\cite{2-78qin2023cross}{, }
                                        Berlo et al.~\cite{2-68berlo2023use}{, }DINN~\cite{1-31ZhouSubject-independent}{, }freeGait~\cite{2-45yan2024freegait}\\
                                        DAFI~\cite{1-7li2021dafi}{, 
                                        }Shi et al.~\cite{2-142shi2020towards}{, 
                                        }Fidora~\cite{1-3chen2022fidora}
                                        , leaf,fill=yellow!15, text width=31.5em
                            ]
                        ]
                        [
                            Domain Alignment with \\ Similarity Computing,fill=yellow!10
                            [
                                   AdaPose~\cite{1-22zhou2024adapose}{, 
                                    }BSWCLoc~\cite{2-96rao2024novel}{, 
                                    }Kang et al.~\cite{1-12kang2021context}{, 
                                    }WiAi-ID~\cite{2-109liang2023wiai}\\
                                    Wi-CHARS~\cite{2-107zhang2023device}{, 
                                    }Mehryar~\cite{2-119mehryar2023domain}{, 
                                    }AdapLoc~\cite{2-151zhou2020adaptive}{, 
                                    }PIAS~\cite{2-77xiao2024pattern}{, 
                                    }CDD~\cite{kang2019contrastive}{, 
                                    }WiCAU~\cite{1-34WiCAU}\\WiSDA~\cite{2-1jiao2024wisda}{, 
                                    }FDAS~\cite{1-30gong2024privacy}{, 
                                    }CDFi~\cite{1-24sheng2024cdfi}{, 
                                    }LAGER~\cite{1-32Chen_LAGER}{, 
                                    }Zhan \& Wu~\cite{2-67zhan2024indoor}{, 
                                    }AirFi~\cite{1-10wang2022airfi}\\
                                    WiLISensing~\cite{2-136ding2020device}
                                    , leaf,fill=yellow!15, text width=31.5em
                        ]
                        ]
                        [
                            Domain Alignment with \\ Generative Adversarial \\ Networks,fill=yellow!10
                            [
                                    MetaGanFi~\cite{2-3zhang2022metaganfi}{, 
                                    }Zhang et al.~\cite{2-61zhang2022cross} {, 
                                    }Wi-Cro~\cite{2-52mao2024wi}{, 
                                    }WiTeacher~\cite{2-39xiao2023mean}
                                    , leaf,fill=yellow!15, text width=31.5em
                            ]
                        ]
                        [
                            Domain Alignment with \\Multi-task Learning,fill=yellow!10
                            [
                                    CrossSense~\cite{1-50zhang2018crosssense}{, 
                                    }Sugimoto et al.~\cite{2-59sugimoto2023towards}{, 
                                    }CSI-MTGN~\cite{1-28zhang2023location}
                                    , leaf,fill=yellow!15, text width=31.5em
                            ]
                        ]
                        [
                            Domain Alignment with \\Cross-modal Embedding,fill=yellow!10
                            [
                                    Wi-Fringe~\cite{2-29islam2020wi}{, 
                                    }XRF55~\cite{wang2024xrf55}{, 
                                    }Wi-Chat~\cite{zhang2025wichat}WiFi2Radar~\cite{1-33Wifi2Radar}
                                    , leaf,fill=yellow!15, text width=31.5em
                            ]
                        ]
                    ]
                    [
                         Component Disentangle\\(\S \ref{sec:component-disentangle}),fill=yellow!10, text width=9em
                        [
                              Elujide et al.~\cite{1-42elujide2022location}{, }UH-Sense~\cite{2-103chen2024task}{, 
                                }Person-in-WiFi~\cite{2-154wang2019person}{, }Wi-Piga~\cite{2-80hao2021wi}{, 
                                }CrossID~\cite{2-18wu2021device}{, 
                                }, leaf,fill=yellow!15, text width=42em
                        ]
                    ]
                    [
                        Metric Learning for \\Zero/Few-shot Learning\\(\S \ref{sec:metric-learning}),fill=yellow!10, text width=9em
                        [
                                 LT-WiOB~\cite{2-28zhou2022enabling}{, 
                                }CATFSID~\cite{2-53wei2024catfsid}{, 
                                }TransferSense~\cite{1-15bu2022transfersense}{, 
                                }UniFi~\cite{2-49liu2024unifi}{, 
                                }WiOpen~\cite{2-54zhang2024wiopen}{, 
                                }NCA~\cite{wu2018improving}{, 
                                }SFTSeg~\cite{2-37xiao2022self}{, 
                                }\\DualConFi\cite{2-167xu2022dual}{, 
                                }CFLH~\cite{2-166wang2024csi}{, 
                                }Xiao et al.~\cite{2-56xiao2024diffusion}{, 
                                }3D-ID~\cite{2-76ren2022robust}{, 
                                }MaP-SGAN~\cite{2-50liang2024map}{, 
                                }CrossFi~\cite{2-44zhao2024crossfi}{, 
                                }Wang et al.~\cite{2-146wang2020learning}{, 
                                }\\WiGR~\cite{2-57hu2021wigr}{, 
                                }LESS~\cite{2-132zhang2023learning}{, 
                                }WiGesID~\cite{2-8zhang2022wi}{, 
                                }AFEE-MatNet~\cite{2-128shi2022environment}{, 
                                }HAR-MN-EF~\cite{1-19shi2020towards}{, 
                                }MatNet-eCSI~\cite{2-21shi2020environment}{, 
                                }Zhou et al.~\cite{1-27zhao2024functional}{, 
                                }\\WiLiMetaSensing~\cite{2-6ding2021wifi-based}{, 
                                }ReWiS~\cite{2-26bahadori2022rewis}{, 
                                }Ding et al.~\cite{2-123ding2021device}{, 
                                }WiONE~\cite{2-126gu2021wione}{, 
                                }CAUTION~\cite{2-130wang2022caution}{, 
                                }FewCS~\cite{2-4yang2023few}{, 
                                }Zhang et al.~\cite{1-6zhang2021wifi}
                                , leaf,fill=yellow!15, text width=42em
                        ]
                    ]
                    [
                        Meta Learning for\\ Few-shot Learning \\(\S \ref{sec:meta-learning}),fill=yellow!10,text width=9em
                        [
                                Huang et al.~\cite{2-47huang2022few}{, 
                                }TOSS~\cite{1-13zhou2022target}{, 
                                }MetaLoc~\cite{2-125gao2023metaloc}{, 
                                }MetaFormer~\cite{1-47sheng2024metaformer}{, 
                                }OneSense~\cite{1-48zhao2024one}{, 
                                }Wi-learner~\cite{2-89feng2022wi}{, 
                                }\\RF-Net~\cite{1-2ding2020rf}{, 
                                }WiGR~\cite{2-75gao2022ml}{, 
                                }MetaGanFi~\cite{2-3zhang2022metaganfi}{, 
                                }Zhang et al.~\cite{2-131zhang2020human}
                                , leaf,fill=yellow!15, text width=42em
                        ]
                    ]
                    [
                       Data Augmentation \\and Data Synthesize \\(\S \ref{sec:data-augmentation}),fill=yellow!10,text width=9em
                        [
                                CrossGR~\cite{1-9li2021crossgr}{, 
                                }Wang et al.~\cite{2-146wang2020learning}{, 
                                }SIDA~\cite{2-51zhang2023sida}{, 
                                }WiARGAN~\cite{2-64huang4160593low}{, 
                                }AFSL-HAR~\cite{2-24wang2021robust}{, 
                                }LESS~\cite{2-132zhang2023learning}{, 
                                }Wi-Dist~\cite{2-71zhang2024toward}{, 
                                }\\CsiGAN~\cite{2-22xiao2019csigan}{, 
                                }Fidora~\cite{1-3chen2022fidora}{, 
                                }Fido~\cite{2-135chen2020fido}{, 
                                }freeLoc~\cite{1-36FreeLoc}{, 
                                }freeGait~\cite{2-45yan2024freegait}{, 
                                }Xiao et al.~\cite{2-56xiao2024diffusion}{, 
                                }Zhou et al.~\cite{1-49zhou2022towards}{, 
                                }CSI-Net~\cite{wang2018csi}{, 
                                }\\BullyDetect~\cite{lan2024bullydetect}{, 
                                }P$^3$ID~\cite{1-37P3ID}{, 
                                }OpenFi~\cite{2-33OpenFi}{, 
                                }Cutout~\cite{devries2017improved}{, 
                                }AirFi~\cite{1-10wang2022airfi}{, 
                                }RFBoost~\cite{2-60hou2024rfboost}{, 
                                }WiSGP~\cite{1-21liu2023generalizing}{, 
                                }WiAG~\cite{2-129virmani2017position}{, 
                                }\\Wi-Cro~\cite{2-52mao2024wi}{, 
                                }DiRA~\cite{wang2025generative2}{,}  
                                SFTSeg~\cite{2-37xiao2022self}{, 
                                }OneFi~\cite{1-44xiao2021onefi}{, 
                                }OneSense~\cite{1-48zhao2024one}{, 
                                }Wang et al.~\cite{2-81wang2024feature}
                                , leaf,fill=yellow!15, text width=42em
                        ]
                    ]
                    [
                        Pseudo Labeling \\(\S \ref{sec:pseudo-labeling}),fill=yellow!10,text width=9em
                        [
                               CLAR~\cite{2-32zhou2022device}{, 
                                }LAGER~\cite{1-32Chen_LAGER}{, 
                                }Zhang et al.~\cite{1-14zhang2023unsupervised}{, 
                                }DA-HAR~\cite{2-69sheng2023har}{, 
                                }Zhang et al.~\cite{2-100zhang2023unsupervised}
                                , leaf,fill=yellow!15, text width=42em
                        ]
                    ]
                ]
                [                
                    Model \\Deployment \\ (\S \ref{sec:deployment}), fill=teal!3
                    [
                        Transfer Learning \\(\S \ref{sec:transfer-learning-with-fine-tuning}), fill=teal!7,text width=9em
                        [
                                XRF55~\cite{wang2024xrf55}{, 
                                }BullyDedect~\cite{lan2024bullydetect}{, 
                                }WiLISensing~\cite{2-136ding2020device}{, 
                                }DADA-AD~\cite{2-19chen2021semi}{, 
                                }Brinke \& Meratnia~\cite{2-161brinke2019scaling}{, 
                                }WiTransfer~\cite{2-95fang2020witransfer}{, 
                                }\\FewSense~\cite{1-25yin2022fewsense}{, 
                                }ResMon~\cite{2-38zheng2023resmon}{, 
                                }RoGER~\cite{2-112bi2024roger}{, 
                                }Hou et al.~\cite{2-127hou2022sample}{, 
                                }DASECount~\cite{2-36hou2022dasecount}
                                , leaf, fill=teal!10,, text width=42em
                        ]
                    ]
                    [
                        Federated Learning \\(\S \ref{sec:federated-learning}), fill=teal!7,text width=9em
                        [
                                WiFederated~\cite{2-5hernandez2021wifederated}{, } pFedBKD~\cite{2-84geng2023personalized}{, } CARING~\cite{1-20li2023towards}{, } Qi et al.~\cite{2-9qi2023resource}{, } Co-WiSensing~\cite{2-20zhang2023cloud}{, } Zhang et al.~\cite{1-5zhang2021privacy}{, }\\ AdaWiFi~\cite{1-35AdaWifi}{, } Zhang et al.~\cite{2-83zhang2023variance}
                                , leaf, fill=teal!10,, text width=42em
                        ]
                    ]
                    [
                        Continual Learning \\(\S \ref{sec:continuous-learning}), fill=teal!7,text width=9em
                        [
                                FewCS~\cite{2-4yang2023few}{, 
                                } WiDrive~\cite{2-134bai2019widrive}{, 
                                } M-WiFi~\cite{2-144soltanaghaei2020robust}{, 
                                } RISE~\cite{2-94zhai2021rise}{, 
                                } CCS~\cite{2-170fu2024ccscontinuouslearningcustomized}{, 
                                } CAREC~\cite{zhang2025carec}{, 
                                }
                                CSI-ARIL~\cite{2-42zhang2023csi}
                                , leaf, fill=teal!10, text width=42em
                        ]
                    ]
                ]
            ]
        \end{forest}
    }
    \caption{We systematically review and categorize nearly a decade of research on Wi-Fi sensing generalization, covering over 200 publications since 2015. Distinct from previous surveys, we organize the literature along a four-stage sensing pipeline—experimental setup, signal preprocessing, feature learning, and model deployment.}
    \label{fig:taxo_of_bench}
\end{figure*}

To enhance the generalizability of Wi-Fi sensing systems, a diverse array of techniques has been proposed. As shown in Fig.~\ref{fig:wifi-sensing-over-years}, research in this field has expanded significantly, evolving from a few studies published between 2015 and 2018 to a surge of recent work focused on improving Wi-Fi generalization. These efforts can be broadly classified into four key stages of Wi-Fi sensing systems: experimental setup, signal preprocessing, feature learning, and model deployment.
\begin{itemize}
    \item Experimental Setup: The primary function is preparation and placement of devices and antennas. There are some approaches that improve generalization by strategically placing antennas around the target area or the human subject~\cite{2-10qin2024direction,2-78qin2023cross}. This spatial diversity helps mitigate the impact of orientation changes and body blocking effects on signal reception. For instance, placing transceivers at multiple angles can reduce performance degradation caused by changes in a subject's facing direction.

    \item Signal Preprocessing: The primary function is extraction of sensing information from Wi-Fi signals, such as denoising and signal transformation. Researchers have explored the use of domain-invariant features that are closely linked to human motion. Examples include Doppler shift and body velocity profile~\cite{1-45zheng2019zero}, which are more resilient to environmental variation than raw CSI or RSS data.

    \item Feature Learning: The primary function is training Wi-Fi sensing models using data. A rich set of machine learning techniques has been employed to bridge domain gaps. These include domain adaptation~\cite{ganin2016domain}, domain generalization~\cite{zhou2022domain}, meta-learning~\cite{finn2017model}, and synthetic data generation, all aiming to learn features that are transferable across devices, subjects, and environments.

    \item Model Deployment: The primary function is deploying trained models to real-world environments. Techniques such as model fine-tuning, federated learning, and continual learning have been used to adapt pre-trained models to new settings with minimal labeled data.
\end{itemize}

\begin{table}[ht]
\centering
\caption{Common Abbreviations in the Article for Quick Retrieval.}
\begin{tabular}{ll}
\toprule
\textbf{Abbreviation} & \textbf{Full Name} \\
\midrule
ADoA & Angular Difference of Arrival \\
AoA & Angle of Arrival \\
AoD & Angle of Departure \\
AP & Access Point \\
BFI & Beamforming Feedback Information \\
BFM & Beamforming Feedback Matrix \\
BVP & Body-coordinate Velocity Profile \\
CNN & Convolutional Neural Network \\
CSI & Channel State Information \\
CWT & Continuous Wavelet Transform \\
DFS & Doppler Frequency Shift \\
DPV & Dynamic Phase Vector \\
GAN & Generative Adversarial Networks \\
ICA & Independent Component Analysis \\
ISAC & Integrated Sensing and Communications \\
MAML & Model-Agnostic Meta-Learning \\
MMD & Maximum Mean Discrepancy \\
NLOS & Non-Line-of-Sight \\
PCA & Principal Component Analysis \\
RKHS& Reproducing Kernel Hilbert Space \\
RNN & Recurrent Neural Network \\
RSS & Received Signal Strength \\
RSSI & Received Signal Strength Indicator \\
STFT & Short-Time Fourier Transform \\
ToF & Time of Flight \\
VAE & Variational Autoencoder \\
\bottomrule
\end{tabular}
\label{tab:abbr}
\end{table}

\begin{figure*}[t]
    \centering
    \subfigure[Distributed Antennas]{
        \begin{minipage}[t]{0.24\linewidth}
            \centering
            \includegraphics[width=\linewidth]{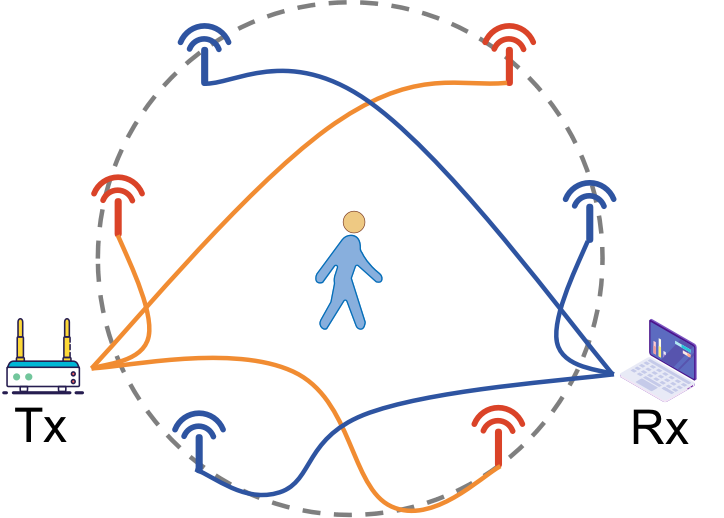}
        \end{minipage}
    }%
    \subfigure[Distributed Devices]{
        \begin{minipage}[t]{0.24\linewidth}
            \centering
            \includegraphics[width=\linewidth]{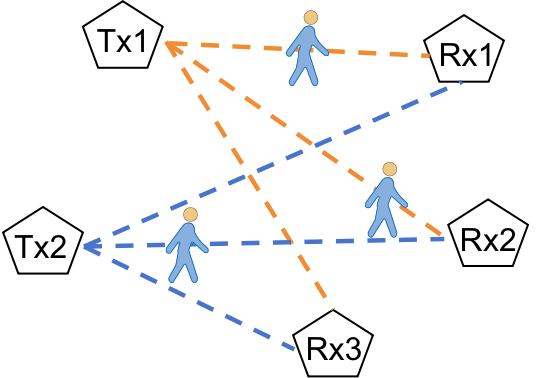}
        \end{minipage}
    }%
    \subfigure[Device Placement]{
        \begin{minipage}[t]{0.24\linewidth}
            \centering
            \includegraphics[width=\linewidth, height=2.8cm]{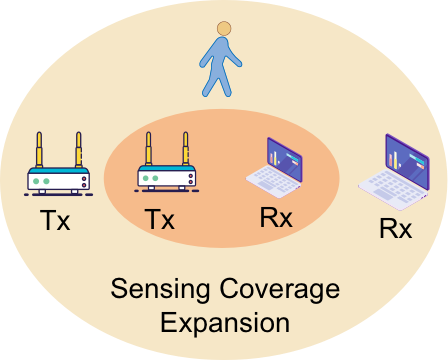}
        \end{minipage}
    }%
    \subfigure[Scaling up dataset]{
        \begin{minipage}[t]{0.24\linewidth}
            \centering
            \includegraphics[width=\linewidth]{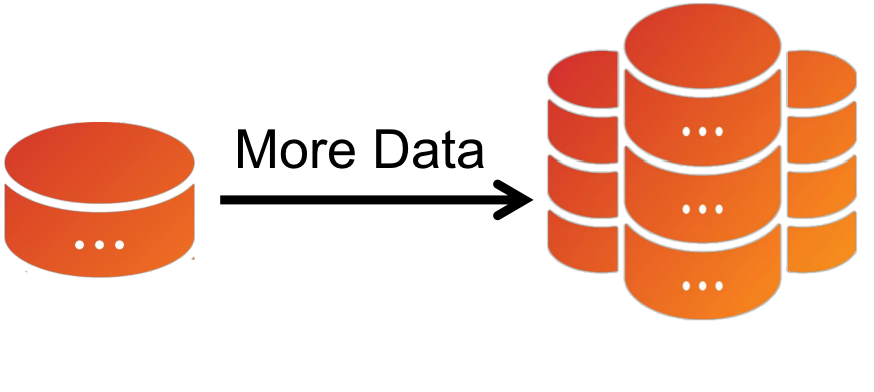}
        \end{minipage}
    }%
    \caption{In the experimental setup stage, Wi-Fi generalization can be enhanced by distributing antennas to mitigate the impact of user orientation, deploying devices more widely or optimizing their placement to increase coverage, and collecting more diverse datasets to support the training of more robust and generalizable models.}
    \label{fig:device-setup}
\end{figure*}

\begin{table*}[htbp]
\centering
\caption{Comparison of Experimental Setup Strategies for Enhanced Generalization.}
\label{tab:hardware_comparison}
\small
\begin{tabularx}{\textwidth}{X X X X X }
\toprule
\textbf{Category} 
& \textbf{Pros (Advantages)} 
& \textbf{Cons (Limitations)} 
& \textbf{Generalization Mechanism} 
& \textbf{Best-fit Scenarios} \\
\midrule

Distributed Antennas  
(\cite{2-10qin2024direction,2-78qin2023cross,2-11wang2018spatial})
& Achieves direction-agnostic sensing with single device pairs; minimal additional hardware cost. 
& Requires manual antenna placement; limited coverage in large-scale areas. 
& Introduces angular diversity by capturing signals from multiple viewpoints, reducing sensitivity to human orientation and body aspect angle. 
& Gesture and activity recognition tasks where user orientation varies but deployment scale is limited. \\ 
\addlinespace

Distributed Devices  
(\cite{2-26bahadori2022rewis, 1-45zheng2019zero,1-44xiao2021onefi, 2-147wu2021witraj})
& Provides high spatial resolution; reduces blind spots and mitigates multipath occlusions. 
& High deployment cost; complex synchronization and calibration across devices. 
& Exploits spatial diversity by observing heterogeneous signal responses across locations, improving robustness to environmental variations. 
& Large indoor spaces or complex layouts requiring robust cross-location generalization. \\ 
\addlinespace

Scaling Up Dataset (\cite{wang2024xrf55, 1-45zheng2019zero,2-91zhang2023csi,yang2024mm, zhu2025csi})
& Enables models to learn implicit invariances; improves robustness and supports few-shot transfer. 
& Labor-intensive data collection; increased storage and training computation cost. 
& Enhances population-level generalization by exposing models to diverse human body types, motion patterns, and interaction styles. 
& Data-driven systems targeting user-independent sensing across diverse populations and activities. \\

\bottomrule
\end{tabularx}
\end{table*}

In the literature, there have been two recent surveys focusing on Wi-Fi sensing generalization. Chen et al.~\cite{chen2023cross} adopt a cross-domain perspective and review over 60 representative studies, emphasizing learning-based techniques such as domain-invariant feature extraction, transfer learning, and few-shot learning. Wang et al.~\cite{wang2024review} focus more narrowly on the few-shot learning paradigm, providing an in-depth discussion of metric learning, transfer learning, and meta-learning approaches for data-scarce scenarios.

While these surveys provide valuable insights into specific aspects of Wi-Fi sensing generalization, they primarily concentrate on learning paradigms and do not offer a system-level view of how generalization challenges arise and are addressed across the entire sensing workflow. In contrast, our survey reviews and synthesizes over 200 publications since 2015 and provides several new insights beyond prior work. First, we organize existing techniques along a four-stage Wi-Fi sensing pipeline, i.e., experimental setup, signal preprocessing, feature learning, and model deployment, revealing how generalization challenges and solutions manifest differently at each stage. Second, to provide actionable guidance, we include ``Key Takeaways and Lessons Learned" for each stage. These sections offer a comparative analysis of different techniques' pros, cons, generalization mechanisms, and best-fit scenarios. Third, by jointly analyzing tools, datasets, learning paradigms, and deployment considerations, our survey moves beyond current techniques to propose a structured research roadmap with short-term and long-term goals for emerging fields like Wi-Fi foundation models and large multimodal models. By providing these multi-dimensional insights—from fundamental principles to practical deployment decisions and future strategic planning—our work offers a more holistic and instructive framework that complements and significantly extends existing literature.

The remainder of this paper is organized as follows. Section~\ref{sec:taxonomy} provides a stage-wise taxonomy of generalization techniques in Wi-Fi sensing, covering key methods and representative studies across experimental setup, data preprocessing, feature learning, and model deployment, respectively. Section~\ref{sec:dataset} presents a comprehensive overview of publicly available Wi-Fi sensing datasets, categorized by sensing tasks, including action recognition, pose estimation, crowd counting, gait recognition, etc. Section~\ref{sec:future} discusses promising future research directions, including multimodal sensing, large-scale model pretraining, and continual learning. Finally, Section~\ref{sec:conclusion} concludes the paper by summarizing key insights on the path forward for building more generalizable Wi-Fi sensing systems. An abbreviation table is provided in Table \ref{tab:abbr}, showcasing the abbreviations of core Wi-Fi signals or methods for quick reference.

\section{Taxonomy}\label{sec:taxonomy}

To enhance the generalization of Wi-Fi sensing systems, researchers must address challenges across various stages, including device deployment, signal processing, feature learning, and model deployment. While some studies concentrate on a single stage, others tackle multiple stages simultaneously, addressing two or three stages at once. To provide a clear and systematic overview of existing research, we categorize the surveyed works based on the Wi-Fi sensing pipeline, as illustrated in Fig.~\ref{fig:taxo_of_bench}.

\subsection{Experimental Setup} \label{sec:hardware_Setup}

At the experimental setup stage, some studies enhance the generalization capability of Wi-Fi sensing by adopting distributed antenna deployment, integrating additional devices, and collecting more extensive datasets, as shown in Fig.~\ref{fig:device-setup}.

To explicitly connect the challenges identified in Fig.~\ref{fig:wifi-sensing-diversity}, we note that each solution is designed to mitigate specific forms of diversity in Wi-Fi sensing systems. 
\begin{itemize}
    \item Device heterogeneity, arising from variations in hardware capabilities, antenna configurations, and radio characteristics across commercial Wi-Fi devices, is primarily addressed through distributed antennas and distributed devices setups, which exploit spatial and hardware diversity to improve robustness.
    
    \item Human body diversity, including differences in body shape, orientation, motion patterns, and carrying conditions, is mitigated by distributed antennas, as multi-view signal observations reduce sensitivity to user orientation and occlusion, and by scaling up datasets, which enables models to generalize across diverse human characteristics. 
    
    \item Environment diversity, caused by changes in room layout, materials, multipath profiles, and deployment locations, is addressed by distributed devices and optimizing device placement that capture spatially heterogeneous channel responses, as well as large-scale and multi-environment datasets, which expose learning models to a broad range of propagation conditions.
\end{itemize}
Together, these experimental setup strategies form a systematic approach to overcoming the key diversity challenges in real-world Wi-Fi sensing.

\subsubsection{Distributed Antennas}\label{sec:distributed-antennas} 

WiNDR~\cite{2-10qin2024direction} and WiCross~\cite{2-78qin2023cross} position three antennas from the transmitter and three antennas from the receiver in a staggered, distributed manner around a 360-degree circle, with the subject performing gestures at the center. This configuration achieves direction-agnostic gesture recognition. Additionally, WiSDAR~\cite{2-11wang2018spatial} tests various distributed antenna topologies, including line, hexagon, square, and random shapes, to explore regions that are most sensitive to human activities, achieving highly accurate and reliable recognition results.

\subsubsection{Distributed Devices}\label{sec:distributed-devices}

Some studies utilize additional Wi-Fi devices to enhance sensing range and improve generalization to human orientation and position. For instance, ReWiS~\cite{2-26bahadori2022rewis}, Widar 3.0~\cite{1-45zheng2019zero}, and OneFi~\cite{1-44xiao2021onefi} adopt configurations of 1 transmitter with 6 receivers and 1 transmitter with 4 receivers, respectively, to extract environment-independent features. WiTraj~\cite{2-147wu2021witraj} leverages 3 receivers placed at different viewing angles to capture human walking and achieve robust trajectory reconstruction. Additionally, Wang et al.~\cite{2-87wang2022placement} propose a theoretical model analyzing how the distance between the transmitter and receiver influences the sensing range. Optimizing the placement of these devices effectively enhances the sensing area and reduces the impact of environmental interference.

\subsubsection{Scaling up dataset}\label{sec:scaling-up-dataset}

Scaling up training datasets, such as collecting large-scale dataset and data augmentation, is another effective approach to enhancing generalization capability, as demonstrated in Widar 3.0~\cite{1-45zheng2019zero}, Zhang et al.~\cite{2-91zhang2023csi}, MM-Fi~\cite{yang2024mm}, XRF55~\cite{wang2024xrf55}, and CSI-Bench~\cite{zhu2025csi}. For example, XRF55~\cite{wang2024xrf55} shows that training on approximately 30,000 samples enables the sensing model to naturally achieve direction and position invariance, while also supporting adaptation to new environments through few-shot fine-tuning.

Table~\ref{tab:hardware_comparison} compares the advantages and limitations of three experimental setup strategies, explains their underlying generalization mechanisms, and summarizes the scenarios in which each approach is best suited. 
Based on this comparison, the analysis of experimental setups provides several strategic insights into building generalizable Wi-Fi sensing systems.

\begin{tcolorbox}[
    colback=hotPink!5!white,
    colframe=hotPink!75!black,
    title=\textbf{Key Takeaways and Lessons Learned: Experimental Setup},
    width=0.48\textwidth,
    boxrule=0.8pt,
    arc=2mm,
    left=2mm,
    right=2mm,
    top=1mm,
    bottom=1mm,
    breakable
]

\begin{itemize}
    \item \textbf{Structural Invariance through Spatial Diversity:} Distributed antenna topologies (e.g., staggered or circular layouts) and multi-device configurations (e.g., 1-TX with 4-6 RXs) provide a physical-layer solution to orientation sensitivity. This ``structural invariance" ensures multi-view signal observations that effectively mitigate body-shadowing and occlusion.
    \item \textbf{Placement as a Robustness Multiplier:} Optimizing the geometric placement of transmitters and receivers is critical for capturing spatially heterogeneous channel responses. Strategic layout design not only maximizes the sensing range but also significantly reduces the impact of environmental multipath interference.
    \item \textbf{Natural Invariance through Data Scaling:} A key lesson from large-scale datasets (e.g., XRF55, CSI-Bench) is the existence of a quantitative threshold where models begin to exhibit ``natural" invariance. Massive data exposure allows models to inherently decouple human motion from specific propagation conditions, facilitating easier adaptation to novel domains.
    \item \textbf{Exploiting Diversity to Counter Heterogeneity:} To overcome device heterogeneity, moving from single-link to multi-link coordinated sensing is essential. Distributed setups exploit spatial and hardware diversity to average out the unique radio artifacts and hardware imperfections of individual commercial Wi-Fi chipsets.

\end{itemize}

\end{tcolorbox}

\begin{figure*}[t]
    \centering
    \includegraphics[width=1\textwidth]{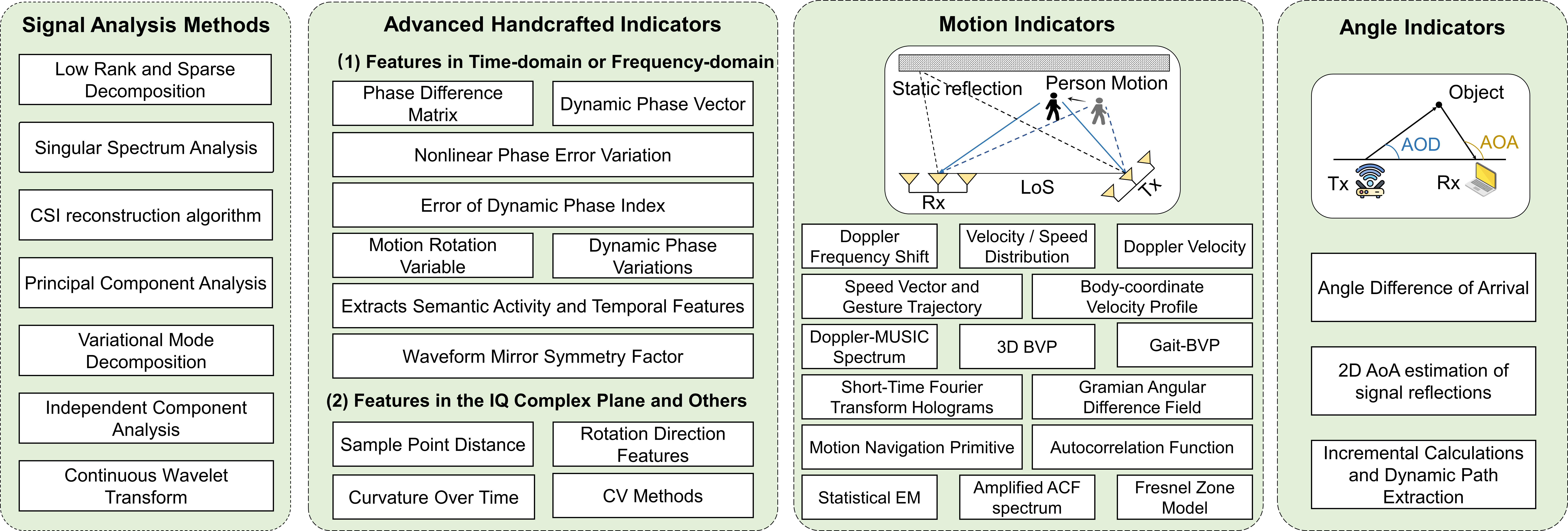}
    \caption{Signal Preprocessing Stage}
    \label{fig:signal-preprocessing}
\end{figure*}


\begin{table*}[htbp]
\centering
\caption{Comparison of Signal Preprocessing Techniques for Wi-Fi Sensing Generalization. }
\label{tab:preprocessing_comparison}
\small
\begin{tabularx}{\textwidth}{XXXXX}
\toprule
\textbf{Category} 
& \textbf{Pros (Advantages)} 
& \textbf{Cons (Limitations)} 
& \textbf{Generalization Mechanism} 
& \textbf{Best-fit Scenarios} \\
\midrule

Signal Analysis  (\cite{2-138lu2019towards, 2-32zhou2022device, 2-79yang2023gait, 2-74zhuo2023position, 2-158abuhoureyah2024multi})
& Effective for noise removal, trend extraction, and background subtraction. 
& Often computationally intensive; may lose non-linear motion details. 
& Extracting motion-related components via signal decomposition (e.g., PCA, ICA, SSA). 
& Stable environments; offline analysis; scenarios requiring strong denoising. \\

\midrule
Handcrafted  (\cite{2-12meng2021wihgr, 2-41huang2021phaseanti, 2-102gao2022towards, 2-149zhang2022handgest, 2-105yu2023towards, 2-7zhang2022csi-based, 2-43peng2023rosefi, 2-110wang2024understanding, 2-17ding2024robust, 2-82peng2024likey, 2-156bu2018wi, 2-159zhang2023enhancing})
& High interpretability; robust to specific interference (e.g., CCI) and phase errors. 
& Sensitive to hardware variations and chipset-specific phase noise. 
& Designing invariant indicators (e.g., Phase Difference, IQ curvature) to bypass shifts. 
& Resource-constrained systems; applications demanding interpretability and robustness. \\

\midrule
Motion Indicators (\cite{2-163meneghello2022sharp, 1-48zhao2024one, 2-98yin2021towards, 2-89feng2022wi, 1-43niu2021understanding, 1-32Chen_LAGER, 2-115liwilife, 2-66van2023mini, 2-70van2022insights, 2-147wu2021witraj, 1-45zheng2019zero, 2-100zhang2023unsupervised, 2-65yang2023witransformer, 2-99shi2023location, 2-86bulugu2023gesture, 2-157zhang2020gaitid, 1-44xiao2021onefi, 2-117chi2024xfall, 2-124gao2021towards, 2-142shi2020towards, 2-46wang2022position, wu2016widir, wang2016human, 2-10qin2024direction, 2-23zhang2019widigr, 2-27zhang2021wi, 2-148zhang2019widetect, 2-63zhu2024wifi})
& High-level physical abstraction; achieves environment and location independence. 
& Requires high-resolution CSI or specific multi-antenna configurations. 
& Mapping signals to body-coordinate velocity/speed profiles (e.g., BVP, DFS). 
& Cross-environment deployment; user-independent sensing; mobility-centric tasks. \\

\midrule
Angle Indicators (\cite{2-13li2020location, 2-111han2023position, 2-76ren2022robust})
& Provides precise spatial geometry; robust for trajectory tracking and Re-ID. 
& Performance is sensitive to antenna array calibration and layout. 
& Utilizing spatial consistency (e.g., AoA, ADoA) to digitize human posture and movement. 
& Localization, tracking, and identification tasks with calibrated antenna arrays. \\

\bottomrule
\end{tabularx}
\end{table*}

\subsection{Signal Preprocessing Stage} \label{sec:signal_preprocessing}

Many works focus on the signal preprocessing stage to extract domain-invariant features to enhance the generalizability of Wi-Fi sensing. These approaches can be broadly categorized into four types: (1) using signal analysis methods, such as Principal Component Analysis (PCA), to extract action-related features from raw signals; (2) processing phase or amplitude information to generate advanced action indicators; (3) extracting motion-specific indicators like Doppler Frequency Shift (DFS) and Body Velocity Profile (BVP) to capture dynamic movements; (4) employing antenna array techniques to derive measurements such as Angle of Arrival (AoA) and Angular Difference of Arrival (ADoA) for sensing. In the following, we introduce four categories respectively.

\subsubsection{Signal Analysis Methods}\label{sec:signal-analysis-methods}

Signal analysis plays a crucial role in Wi-Fi sensing by extracting relevant features from raw signals, ensuring generalizability across various environments, locations, and orientations. WiHand~\cite{2-138lu2019towards} employs a Low Rank and Sparse Decomposition (LRSD) algorithm to separate gesture signals from background noise, enhancing its resilience to location variations. CLAR~\cite{2-32zhou2022device} utilizes Singular Spectrum Analysis (SSA)~\cite{golyandina2001analysis} to extract the trend components from Wi-Fi signals, enabling robust activity recognition. Gait-Enhance~\cite{2-79yang2023gait} introduces a CSI reconstruction algorithm that mitigates the negative impact on recognition accuracy caused by varying CSI patterns associated with walking in different directions. Zhuo et al.~~\cite{2-74zhuo2023position} apply a combination of PCA and Variational Mode Decomposition (VMD)~\cite{pca-vmd-dragomiretskiy2013variational} to reduce noise interference in non-ideal sleep positions, enabling position-free breath detection. Abuhoureyah et al.~\cite{2-158abuhoureyah2024multi} leverage Independent Component Analysis (ICA) and Continuous Wavelet Transform (CWT) to decompose CSI signals, improving multi-user human action recognition in complex environments.

\subsubsection{Advanced Handcrafted Indicators}\label{sec:advanced-wifi-indicators}

Many works have focused on exploring robust handcrafted features for domain-independent sensing. WiHGR~\cite{2-12meng2021wihgr} uses a Phase Difference Matrix to extract phase-related features. PhaseAnti~\cite{2-41huang2021phaseanti} introduces Nonlinear Phase Error Variation (NLPEV), which is independent of Cochannel Interference (CCI), to mitigate interference and enhance recognition accuracy. DPSense~\cite{2-102gao2022towards} proposes the Error of Dynamic Phase Index (EDP-index) to evaluate the sensing quality of signal segments, prioritizing high-quality segments to improve gesture recognition performance. HandGest~\cite{2-149zhang2022handgest} combines Dynamic Phase Vector (DPV) and Motion Rotation Variable (MRV) for in-the-air handwriting recognition. Similarly, Yu et al.~\cite{2-105yu2023towards} extract dynamic phase features for position-independent gesture recognition. Additionally, AF-ACT~\cite{2-7zhang2022csi-based} extracts semantic activity and temporal features to characterize activities across various locations. RoSeFi~\cite{2-43peng2023rosefi} identifies contextual association and waveform mirror symmetry in CSI data, introducing the Waveform Mirror Symmetry Factor (WMSF) to quantify symmetry between sit-stand postural transition (SPT) CSI data, aiding in sedentary behavior detection.

Some works also extract features in the IQ complex plane. For instance, Wang et al.~\cite{2-110wang2024understanding} analyze the curvature of signal curves over time in the IQ plane. WiGesFree\cite{2-17ding2024robust} introduces the Sample Point Distance (SPD), which uses changing CSI ratio patterns in the complex plane to capture gesture information accurately. LiKey~\cite{2-82peng2024likey} leverages rotational direction features in the CSI complex plane as an extraction of location-independent features. Other works transform Wi-Fi CSI data into images and use visual methods, such as pre-trained models and data augmentation, to improve generalization capabilities. Bu et al.~\cite{2-156bu2018wi} convert the amplitude of each CSI stream into a grayscale CSI image and fine-tune pre-trained networks like VGG16 and VGG19 for feature extraction. Zhang et al.~\cite{2-159zhang2023enhancing} convert one-dimensional CSI time series into two-dimensional recurrence plots (RP) and apply data augmentation techniques like horizontal flipping, cropping, and color distortion to improve model performance and robustness in action recognition.

\subsubsection{Motion Indicators}\label{sec:motion-indicators}

Many works compute domain-independent motion indicators in speed, velocity, direction, and trajectory, to enhance the generalizability of Wi-Fi sensing systems. Sharp~\cite{2-163meneghello2022sharp}, OneSense~\cite{1-48zhao2024one}, Yin et al.~\cite{2-98yin2021towards}, Wi-learner~\cite{2-89feng2022wi}, Niu et al.~\cite{1-43niu2021understanding}, and LAGER~\cite{1-32Chen_LAGER} focus on utilizing Doppler Frequency Shift (DFS). Mini-Batch Alignment~\cite{2-66van2023mini, 2-70van2022insights} incorporates both DFS and Gramian Angular Difference Field (GADF). WiLife~\cite{2-115liwilife} explores Doppler velocity as a key motion indicator. WiTraj~\cite{2-147wu2021witraj} identifies speed ambiguity in the Doppler-MUSIC algorithm's spectrum. To address this, it employs multiple receivers at different viewing angles and leverages the CSI quotient to eliminate this ambiguity, enabling robust trajectory reconstruction.

Widar 3.0~\cite{1-45zheng2019zero} introduces the Body-coordinate Velocity Profile (BVP), which captures the power distribution across different velocities of body parts involved in gesture movements. This approach enhances generalization to new locations, environments, and subjects while maintaining performance consistency. BVP and its variants have demonstrated remarkable generalization capabilities. For instance, Zhang et al.~\cite{2-100zhang2023unsupervised} and WiTransformer~\cite{2-65yang2023witransformer} adopt BVP, while Shi et al.~\cite{2-99shi2023location} and Bulugu~\cite{2-86bulugu2023gesture} extend it to 3D BVP. GaitID~\cite{2-157zhang2020gaitid} utilizes gait-specific BVP (gait-BVP), OneFi~\cite{1-44xiao2021onefi} employs velocity distribution, and XFall~\cite{2-117chi2024xfall} incorporates a speed distribution profile to enhance robustness. In addition to BVP, WiGesture~\cite{2-124gao2021towards} proposes the Motion Navigation Primitive (MNP) to improve gesture recognition accuracy. Shi et al.~\cite{2-142shi2020towards} convert CSI amplitude and relative amplitude into frequency-domain representations, such as STFT holograms, to characterize the moving speeds of different body parts. Wang et al.~\cite{2-46wang2022position} use phase changes of the reflection path to relate the speed vector and gesture trajectory in the body coordinate system, implicitly capturing the impact of position and orientation, which is estimated by a preamble gesture in each new environment.

The Fresnel Zone Model describes the signal propagation process with better interpretability, and is therefore widely used to enhance the generalization capability of Wi-Fi sensing. Widir~\cite{wu2016widir} utilizes it to capture walking direction. Wang et al.~\cite{wang2016human} apply the model to detect breast motion for respiration detection. WiNDR~\cite{2-10qin2024direction} leverages the model for direction-agnostic gesture recognition. Additionally, WiDIGR~\cite{2-23zhang2019widigr} and Wi-PIGR~\cite{2-27zhang2021wi} use the Fresnel Zone Model to eliminate direction information for gait recognition. Statistical EM methods have also been applied to Wi-Fi sensing. WiDetect~\cite{2-148zhang2019widetect} uses statistical EM theory to model motion detection, connecting the Autocorrelation Function~(ACF) of CSI to target motion, independent of environment, location, orientation, and subjects. Zhu et al.~\cite{2-63zhu2024wifi} extract the Amplified ACF spectrum from CSI, enabling motion features to be isolated, independent of environmental factors.

\subsubsection{Angle Indicators}\label{sec:aoa-indicators}

Angle indicators are also used to enhance the Wi-Fi sensing generalizability. Li et al.~\cite{2-13li2020location} use Angular Difference of Arrival for location-free CSI-based activity recognition. EasyTrack~\cite{2-111han2023position} utilizes AoA–AoD values, leveraging changes in the length and direction of the Wi-Fi signal path. Through incremental calculations and dynamic path extraction, it can reconstruct gesture trajectories in real-time without relying on the initial gesture position or a fixed device location. 3D-ID~\cite{2-76ren2022robust} uses 2D AoA estimation of signal reflections to enable Wi-Fi to visualize a person in the environment. This visualization is then digitized into a 3D body representation, which extracts both the static body shape and dynamic walking patterns for person re-identification.

Table~\ref{tab:preprocessing_comparison} compares the advantages and limitations of four signal preprocessing strategies, explains their underlying generalization mechanisms, and summarizes the scenarios in which each approach is best suited. 
Based on this comparison, the analysis of signal preprocessing provides several critical takeaways and lessons learned for building generalizable Wi-Fi sensing systems.

\begin{tcolorbox}[
    colback=hotPink!5!white,
    colframe=hotPink!75!black,
    title=\textbf{Key Takeaways and Lessons Learned: Signal Preprocessing},
    width=0.48\textwidth,
    boxrule=0.8pt,
    arc=2mm,
    left=2mm,
    right=2mm,
    top=1mm,
    bottom=1mm,
    breakable
]
\begin{itemize}
    \item \textbf{Signal Analysis for Denoising:} Algorithms like LRSD, SSA, and PCA are fundamental for extracting action-related components from background noise. These methods enhance resilience to location variations and enable robust monitoring (e.g., breath detection) even in non-ideal positions.
    \item \textbf{Handcrafted IQ \& Phase Features:} Advanced indicators derived from phase differences or the IQ complex plane capture fine-grained patterns. Notably, features like NLPEV provide independence CCI, which is crucial for real-world robustness.
    \item \textbf{Physical Invariance via Motion Indicators:} Transitioning to speed/velocity profiles, particularly BVP and the Fresnel Zone Model, represents a shift toward domain invariance. By mapping signals to speed or body-relative coordinates, systems decouple sensing from user orientation and environment.
    \item \textbf{Geometric Consistency with Angle Indicators:} Angle-based measurements, such as AoA and ADoA, provide spatial geometric paths consistent across domains. These enable real-time trajectory reconstruction and 3D body representation, supporting complex tasks like person re-identification.
    \item \textbf{Signal-to-Visual Feature Enrichment:} Converting 1D CSI time-series into 2D representations (e.g., Recurrence Plots, Grayscale images) bridges the gap between signal processing and computer vision, allowing models to benefit from powerful pre-trained visual architectures and data augmentation.
\end{itemize}
\end{tcolorbox}

\subsection{Feature Learning Stage} \label{sec:Feature Learning Stage}
The Feature Learning Stage is fundamental to improving the adaptability and robustness of Wi-Fi sensing systems across diverse domains. This stage includes several key techniques aimed at overcoming the challenges of domain variability, limited labeled data, and environmental changes. Methods such as domain alignment, metric learning, data augmentation, and pseudo-labeling work together to enhance model performance, ensuring that features learned from one domain can be successfully applied to others. These approaches through techniques like adversarial training, similarity metrics, data synthesis, and embedding refinement, enable Wi-Fi sensing models to generalize effectively in dynamic and heterogeneous real-world environments.

\vspace{5pt}

\subsubsection{\textbf{Domain Alignment}}\label{sec:domain-alignment}

Adversarial Domain Adaptation \cite{ganin2016domain, tzeng2017adversarial} aims to align the source and target domains, enabling Wi-Fi sensing to generalize across new individuals, environments, and other novel domains. 

\paragraph{Domain Alignment with Domain Discriminator}

\begin{figure}[t]
    \centering
    \includegraphics[width=1\linewidth]{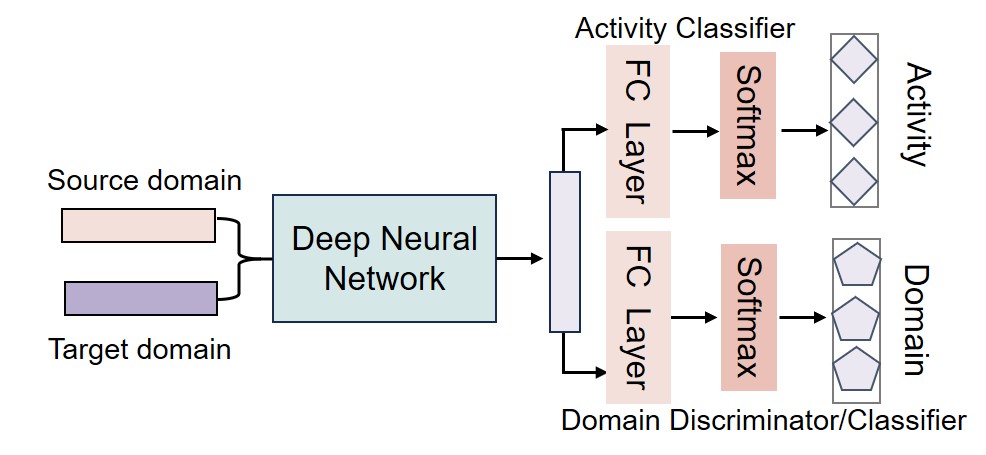}
    \caption{Domain alignment with domain discriminator or domain classifier. Both source and target data are mapped into a shared feature space. Through adversarial competition between an action classifier and a domain discriminator, the network learns to produce domain-agnostic representations that enhance cross-domain performance.}
    \label{fig:dann}
\end{figure}

Fig.~\ref{fig:dann} illustrates a representative domain-adversarial training framework for Wi-Fi sensing generalization. The model takes both source-domain and target-domain Wi-Fi data as input and employs a deep neural network to extract shared features. These features are then fed into two branches: an action classifier for activity recognition, and a domain discriminator that attempts to determine whether the input originates from the source or target domain. Through adversarial training, the feature extractor learns to produce domain-invariant representations that are disentangled along the domain dimension, thereby enabling more robust cross-domain activity recognition. This fundamental framework was employed in WiHARAN~\cite{1-17wang2021environment}, DA-HAR~\cite{2-69sheng2023har}, and Zhang et al.~\cite{1-5zhang2021privacy} to improve cross-environment activity recognition.

Several variants have been proposed to enhance this domain-adversarial learning framework further. For example, i-Sample~\cite{1-18zhou2024sample} adopts a two-stage training strategy. In the first stage, the model focuses on learning a domain-invariant feature extractor by jointly feeding source and target domain data. A domain discriminator is used to classify the domain origin of each sample, while a parallel branch performs activity classification. In the second stage, the feature extractor is frozen, and the classifier is fine-tuned using labeled source data. This staged approach effectively enhances cross-environment generalization performance. In another line of work, JADA~\cite{2-2zou2018joint} employs two separate encoders: one for the source domain and another for the target domain. Through adversarial training with a domain discriminator, the features extracted from both encoders are encouraged to become domain-invariant. After training, the source-domain encoder is frozen, and a classification head is trained using labeled source data. The final model, comprising the target-domain encoder and the trained classification head, is used for inference on target-domain samples, achieving robust cross-domain activity recognition. Other examples are CrossGR~\cite{1-9li2021crossgr} and Wi-Adaptor~\cite{2-30zhang2021wi}, which replace the traditional domain discriminator with an auxiliary action classifier. Unlike standard classifiers, this branch is trained to assign uniform probabilities across all action classes, effectively making actions indistinguishable in the feature space. This strategy forces the feature extractor to remove task-relevant cues from the domain-variant features, thereby encouraging it to learn more domain-invariant and task-discriminative representations. As a result, the main action classifier benefits from cleaner, more generalizable features during training.

\paragraph{Domain Alignment with Domain Classifier} 

Another line of work extends the domain-adversarial framework by replacing the domain discriminator with a domain classifier. While the standard domain discriminator performs binary classification to distinguish between source and target domains, the domain classifier generalizes this concept to multi-class classification, where each domain (e.g., different environments or deployment scenarios) is assigned a distinct label, also shown in Fig.~\ref{fig:dann}. For example, EI~\cite{jiang2018towards} introduces a domain classifier to explicitly recognize the environment in which the activities are recorded. Through adversarial training, the domain classifier is optimized to correctly classify each environment, while the feature extractor is trained to fool the domain classifier, thereby learning environment-invariant representations. Similarly, Khattak and Khan~\cite{2-72khattak2022cross} and WiCAR~\cite{1-11wang2019wicar} adopt this strategy to achieve location-invariant activity recognition and in-car environment-invariant activity recognition, respectively.

Beyond environments and locations, user identity can also serve as a domain label. Systems such as GESFI~\cite{2-168zhang2024objective}, ADA~\cite{2-55zinys2021domain}, and DATTA~\cite{2-155strohmayer2024datta} use the user ID as the domain label to enable user-independent gesture recognition. This idea can be extended further: domain labels include user identity, body orientation, physical position, and environment. Building on this fine-grained decomposition, models like WiSR~\cite{1-29liu2023wisr}, Yin et al.~\cite{2-98yin2021towards}, INDG-Fi~\cite{2-48yang2024domain}, WiCross~\cite{2-78qin2023cross}, and Berlo et al.~\cite{2-68berlo2023use} demonstrate the feasibility of learning representations that generalize across multiple domain factors simultaneously. 

It is worth noting that this domain-classifier-based adversarial framework is not limited to activity recognition. By replacing the action classifier with other task-specific heads, this architecture has been successfully applied to a variety of tasks, including: pose estimation across subjects (e.g., DINN~\cite{1-31ZhouSubject-independent}), gait recognition across actions (e.g., freeGait~\cite{2-45yan2024freegait}), indoor localization across environments (e.g., DAFI~\cite{1-7li2021dafi}), and user identification across locations (e.g., Shi et al.~\cite{2-142shi2020towards}). These extensions further underscore the versatility and effectiveness of domain-classifier-based adversarial training in improving generalization across a wide range of Wi-Fi sensing applications. In addition, Fidora~\cite{1-3chen2022fidora} replaces the domain classification branch with a data reconstruction branch. In this framework, both source and target domain samples are first encoded by a shared feature extractor, and then passed through a decoder to reconstruct the original input data. The reconstruction loss serves as an indirect objective for domain alignment, encouraging the feature extractor to retain domain-invariant information that is useful for reconstruction.

\paragraph{Domain Alignment with Similarity Computing}

\begin{figure}[t]
    \centering
    \includegraphics[width=1\linewidth]{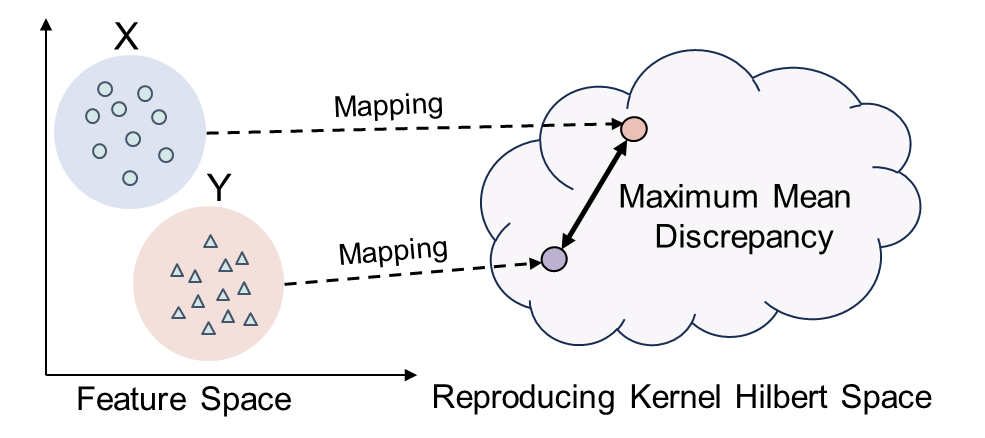}
    \caption{Maximum Mean Discrepancy~\cite{gretton2006kernel} quantifies the similarity between two probability distributions by projecting data into a Reproducing Kernel Hilbert Space (RKHS) and computing the squared distance between their mean embeddings, encouraging the model to learn domain-invariant representations.}
    \label{fig:mmd}
\end{figure}

Another widely adopted strategy for domain alignment is to explicitly minimize the distribution discrepancy between source and target domains. One representative approach is Maximum Mean Discrepancy (MMD)~\cite{gretton2006kernel}, which quantifies the difference between two probability distributions by projecting data into a Reproducing Kernel Hilbert Space (RKHS) and computing the squared distance between their mean embeddings, encouraging the model to learn domain-invariant representations, as shown in Fig.~\ref{fig:mmd}. In the context of Wi-Fi sensing, MMD has been successfully integrated into model objectives to mitigate domain shifts and enhance cross-domain generalization. For example, AdaPose~\cite{1-22zhou2024adapose} and BSWCLoc~\cite{2-96rao2024novel} utilize MMD-based alignment for cross-environment human pose estimation and indoor localization, respectively.

Beyond standalone MMD alignment, some works combine MMD with domain classifiers to further enhance generalization. For instance, Kang et al.~\cite{1-12kang2021context} introduce a dual-branch architecture, where MMD is applied between the source and target domains at the output of the activity classifier, while a parallel domain classifier is trained to align features across domains. This design enables robust activity recognition across variations in user identity, environment, location, and body orientation. Similarly, WiAi-ID~\cite{2-109liang2023wiai} adopts MMD to align source and target features, while employing a domain classifier to discriminate appearance characteristics, achieving appearance-independent passive person identification.

In addition to MMD, several alternative metrics have been adopted to quantify distributional similarity between source and target domains. For example, Wi-CHARS~\cite{2-107zhang2023device} applies Earth Mover’s Distance (EMD) to align features for cross-environment activity recognition. Mehryar~\cite{2-119mehryar2023domain} utilizes the Kullback–Leibler (KL) divergence to reduce domain shift, while AdapLoc~\cite{2-151zhou2020adaptive} employs Euclidean distance for indoor localization across domains. PIAS~\cite{2-77xiao2024pattern} further explores Contrastive Domain Discrepancy (CDD)~\cite{kang2019contrastive} to align representations in a contrastive manner. WiCAU~\cite{1-34WiCAU} uses Wasserstein distance to measure and quantify distributional similarity between source and target domains.

In some cases, in addition to knowing the domain labels (e.g., environment), the class labels (e.g., action types) of both source and target domain samples are also available during training. This enables a finer-grained alignment approach, where domain-level similarity measures can be replaced by class-level distribution alignment, allowing for more precise cross-domain representation learning. A number of studies have explored this strategy under different names and formulations. For example, local MMD~\cite{2-15hassan2024adversarial,1-37P3ID}, mini-batch alignment~\cite{2-66van2023mini,2-70van2022insights}, Multiple Kernel MMD~\cite{2-44zhao2024crossfi}, and sub-domain alignment~\cite{1-41li2021subdomain} all share the common idea of aligning corresponding action classes across domains, instead of aligning the global distributions. The similarity metrics used for sub-domain alignment vary across works. For instance, P$^3$ID utilizes Euclidean distance to measure the closeness between class features from different environments. WiSDA~\cite{2-1jiao2024wisda}, FDAS~\cite{1-30gong2024privacy}, and CDFi~\cite{1-24sheng2024cdfi} adopt cosine similarity to measure the closeness between class-conditional features from different domains. Notably, CDFi further incorporates Jensen–Shannon (JS) divergence~\cite{remus2012domain} to perform an additional global domain alignment, complementing the sub-domain matching process.

In cases where only domain-level labels (e.g., environment, location) are available, but class labels of the target domain samples are unknown, some approaches generate pseudo-labels for target domain data during training to enable the above fine-grained domain alignment. For example, LAGER~\cite{1-32Chen_LAGER} and Zhan \& Wu~\cite{2-67zhan2024indoor} adopt this strategy by first predicting class labels for the unlabeled target domain samples. These pseudo-labeled data are then used to perform sub-domain alignment. Both methods use Euclidean distance to measure the similarity between class-conditional feature distributions. LAGER achieves domain generalization in cross-user, cross-person, cross-location, and cross-orientation activity recognition, while Zhan \& Wu's approach targets cross-scene indoor localization.

When only multiple source domains are available during training and no target domain data is accessible, models can still improve generalization by aligning the source domains themselves. For example, AirFi~\cite{1-10wang2022airfi} performs MMD-based alignment across multiple source domains, learning domain-invariant features that generalize to unseen target environments for activity recognition. Similarly, WiLISensing~\cite{2-136ding2020device} adopts an MMD alignment strategy across source domains to train the feature extractor. During deployment, the feature extractor is frozen, and only a small number of labeled samples from the new environment are used to train the final fully-connected layers. This approach effectively enables cross-environment activity recognition with minimal adaptation effort.

\paragraph{Domain Alignment with Generative Adversarial Networks}

\begin{figure}[ht]
        \centering
        \includegraphics[width=1\linewidth]{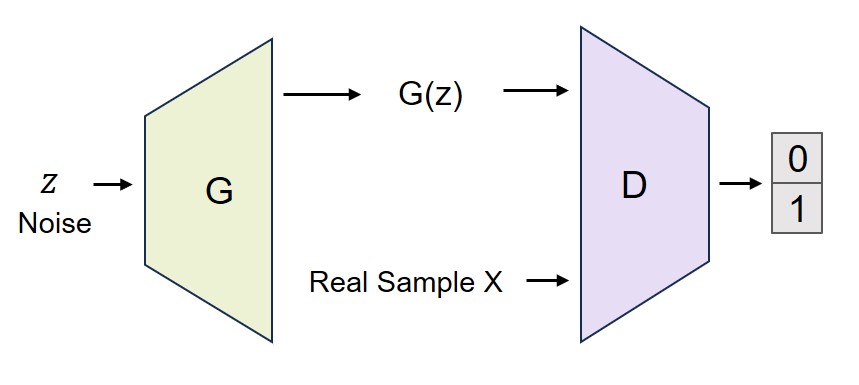}
    \caption{Generative Adversarial Networks. \textbf{Generator ($G$):} The network responsible for creating synthetic data. Its primary objective is to learn the underlying distribution of the real data and produce samples that are indistinguishable from the ground truth to ``fool" the discriminator. \textbf{Discriminator ($D$):} The network acting as a binary classifier. Its role is to evaluate both real data ($x$) and synthetic data ($G(z)$), outputting a probability that a given sample is ``real" (from the actual dataset) rather than ``fake" (produced by the generator). It provides the gradient signal that guides the generator to improve its realism.} 
    \label{fig:gan}
\end{figure}

 Another line of research explores Generative Adversarial Networks (GANs)~\cite{goodfellow2014generative}  to perform domain alignment by transforming data distributions or learning shared latent representations. GAN-based alignment builds on the adversarial learning mechanism of GANs, shown in Fig.~\ref{fig:gan}, which involves a generator (G) and a discriminator (D) engaged in a minimax game. The generator aims to produce data indistinguishable from real samples, while the discriminator attempts to differentiate between real and generated data. Through iterative optimization, the generator learns to synthesize data that closely follows the expected distribution. For example, MetaGanFi~\cite{2-3zhang2022metaganfi} proposes a three-stage pipeline: (1) In the first stage, a CycleGAN-based framework~\cite{zhu2017unpaired} is used to convert Wi-Fi data from multiple source domains into a unified Uni-domain representation. This is achieved through adversarial training, where the generator transforms source-domain data to resemble a common distribution, and a discriminator is trained to identify the originating domain. The generator is optimized to fool the discriminator, thereby removing domain-specific cues. (2) In the second stage, the generated Uni-domain data are used to train an activity classifier, ensuring that the classifier learns from domain-invariant representations. (3) Finally, during inference, target-domain data are passed through the trained generator and classifier to perform activity recognition for unseen users. 
 Similarly, Zhang et al.~\cite{2-61zhang2022cross} and Wi-Cro~\cite{2-52mao2024wi} and WiTeacher~\cite{2-39xiao2023mean} employ CycleGAN and StyleGAN~\cite{karras2019style}, respectively, but instead of mapping source domains to a unified representation, they transform source domain data to match the target domain distribution, thereby enabling cross-domain sign language recognition and activity recognition.

\paragraph{Domain Alignment with Multi-task Learning}

Multi-task learning improves generalization by jointly training a model on multiple related tasks or domains, either through training multiple expert networks or by sharing parameters across tasks. This approach encourages the learning of representations that are both domain-invariant and task-relevant. 

For example, CrossSense~\cite{1-50zhang2018crosssense} trains 10 expert models. The experts are trained offline, and at runtime, the appropriate expert for a given input is automatically chosen. This expert selection mechanism allows the system to adaptively handle different input conditions, improving robustness across domains. Similarly, Sugimoto et al.~\cite{2-59sugimoto2023towards} propose a system that employs a multi-task learning approach based on an encoder-decoder network architecture, where each decoder corresponds to a specific target environment. This structure ensures that the encoder learns a more general representation that can accommodate the diversity between different environments. CSI-MTGN~\cite{1-28zhang2023location} also adopts multi-task learning to learn shared parameters for the target task. Instead of using separate decoders or experts, CSI-MTGN builds a unified model where parameters are shared across multiple tasks. This shared representation facilitates knowledge transfer between tasks and domains, enhancing the model's ability to generalize from limited data. By integrating shared learning modules and task-specific heads, CSI-MTGN effectively balances generalization and specialization, making it well-suited for location-aware sensing tasks with high domain variability.

\paragraph{Domain Alignment with Cross-modal Embedding}

Cross-modal alignment leverages the rich semantic information inherent in modalities such as text, images, and radar signals, which are often better understood by large-scale pre-trained models (e.g., language models or vision-language models) than raw Wi-Fi CSI. By mapping Wi-Fi data into these semantically meaningful spaces, models can benefit from external knowledge and enhanced generalization. Consequently, several works explore cross-modal embedding techniques to bridge CSI with modalities such as text, vision, and mmWave radar, enabling more robust domain adaptation.

Wi-Fringe~\cite{2-29islam2020wi} converts action class labels into BERT embeddings~\cite{devlin2019bert} and uses them as targets to guide the learning of Wi-Fi representations. This enables the system to bridge the semantic gap between textual activity descriptions and Wi-Fi signal patterns. Such alignment with language models facilitates better generalization to unseen activities, particularly in open-set scenarios where the label space at test time differs from training. XRF55~\cite{wang2024xrf55} also employs BERT embeddings as additional supervision to enhance the semantic alignment of Wi-Fi feature learning. Wi-Chat~\cite{zhang2025wichat} goes one step further by using natural language to describe Wi-Fi signal patterns corresponding to behaviors such as walking, falling, no-event, and breathing. These descriptions are provided to a large language model (GPT-4o), enabling it to understand and interpret Wi-Fi sensing data. This approach enables zero-shot action recognition and respiratory monitoring, demonstrating a novel integration of wireless sensing with LLMs.

In addition to text-based alignment, radar signals also offer fine-grained motion information that complements Wi-Fi data. WiFi2Radar~\cite{1-33Wifi2Radar} proposes a cross-modal translation framework that converts CSI spectrograms into mmWave radar Doppler spectrograms using a U-Net~\cite{ronneberger2015u} trained with synchronized radar data. This allows the model to inherit radar’s high-resolution motion sensitivity while retaining the deployment convenience of Wi-Fi infrastructure.

\paragraph{Summary and Design Insights for Domain Alignment} 
The above domain alignment strategies reveal a fundamental trade-off between alignment granularity, supervision availability, and training stability. 
Adversarial alignment with domain discriminators provides a flexible and widely adopted solution for unsupervised domain adaptation, particularly when only source-domain labels are available. 
Extending the discriminator to multi-class domain classifiers further enables disentanglement across multiple domain factors (e.g., user, environment, orientation), making it well-suited for complex real-world deployments with explicitly labeled domain attributes. 
However, adversarial training may introduce optimization instability and potential over-alignment, motivating the exploration of more stable alternatives.

Similarity-based approaches, such as MMD, Wasserstein distance, and KL divergence, offer a theoretically grounded and optimization-friendly mechanism for distribution matching. 
They are particularly effective under moderate domain shifts and limited computational budgets, though their effectiveness may diminish when distribution gaps become large. 
When finer-grained supervision is available, class-level or sub-domain alignment provides more precise cross-domain matching, reducing negative transfer by aligning semantically corresponding categories instead of global distributions. 
Nevertheless, such approaches rely on accurate class labels or high-quality pseudo-labels, which may not always be obtainable.

GAN-based alignment extends domain adaptation by explicitly transforming data distributions, making it advantageous for large domain discrepancies or cross-modality translation. 
However, its higher computational complexity and adversarial instability limit scalability in resource-constrained sensing systems. 
Multi-task learning alignment, in contrast, improves generalization by leveraging diversity across tasks or domains without necessarily requiring target-domain data, making it attractive for domain generalization settings. 
Finally, cross-modal embedding represents a promising direction, as it connects Wi-Fi sensing representations with semantically rich modalities (e.g., text or radar), enabling zero-shot generalization and integration with foundation models.

In practice, the selection of domain alignment strategy should be guided by (i) the availability of domain and class labels, (ii) the magnitude and nature of domain shifts, and (iii) computational constraints. It is expected to move toward hybrid frameworks that combine stable similarity metrics, fine-grained alignment, and cross-modal semantic supervision to enable scalable and robust generalization of Wi-Fi sensing.

\vspace{5pt}


\subsubsection{\textbf{Component Disentangle}}\label{sec:component-disentangle}

Wi-Fi signals inherently contain entangled information about the environment, human presence, and motion. Several works aim to disentangle these components to enable cross-domain generalization.

For example, Elujide et al.~\cite{1-42elujide2022location} leverage unsupervised adversarial invariance~\cite{jaiswal2018unsupervised_uai} to disentangle gesture-related features from location-related features in the latent space, thereby enabling gesture recognition in unseen locations.
UH-Sense~\cite{2-103chen2024task} employs a GAN-based framework to denoise the data affected by UAV jitter. By filtering out motion-induced distortions, the model achieves robust human localization based on Wi-Fi signals collected from UAV platforms.
Person-in-WiFi~\cite{2-154wang2019person} assumes that Wi-Fi signals are composed of both environment-related and motion-related components. It uses a GAN to transform the signal representation into an environment-invariant latent space—one in which a domain discriminator cannot predict the environment index—thereby improving generalization for cross-environment human pose estimation.
Wi-Piga~\cite{2-80hao2021wi} assumes that Wi-Fi signals contain identity-related and action-related components. It utilizes a GAN to generate features in which the identity information is no longer distinguishable. These identity-invariant features are then used for yoga pose recognition, enabling cross-user generalization.
CrossID~\cite{2-18wu2021device} leverages conditional instance normalization~\cite{dumoulin2016learned} to learn affine parameters that capture identity attributes. During inference, these parameters can be adapted to new users, enabling gesture recognition for previously unseen individuals.

\vspace{5pt}

\subsubsection{\textbf{Metric Learning for Zero/Few-shot Learning}}\label{sec:metric-learning}

\begin{figure}[t]
    \centering
    \includegraphics[width=1\linewidth]{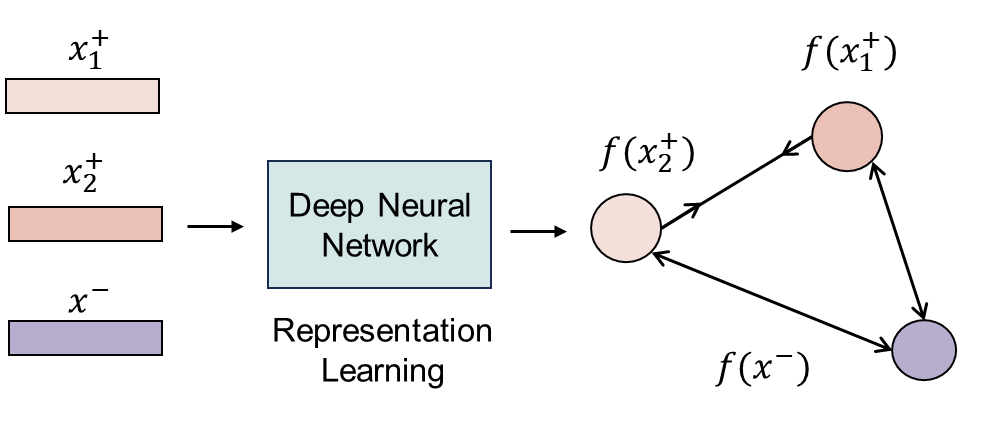}
    \caption{Triplet loss aims to structure the feature space such that samples from the same class are mapped close together ($x_1^+, x_2^+$), while samples from different classes are pushed farther apart ($x^+, x^-$).}
    \label{fig:triplet-loss}
\end{figure}

Metric learning is another important strategy for Wi-Fi sensing generalization, which aims to structure the feature space such that samples from the same class are mapped close together ($x_1^+, x_2^+$), while samples from different classes are pushed farther apart ($x^+, x^-$). This is typically achieved during training by optimizing pairwise or triplet losses over sample embeddings, as shown in Fig.~\ref{fig:triplet-loss}. After training, the resulting feature space forms well-separated clusters corresponding to different semantic classes.

At inference time, unlike traditional classification approaches that directly predict class labels, metric learning relies on a support set, which contains representative examples (templates) for each class. For example, in cross-environment activity recognition, the support set may consist of one labeled sample per activity from the target environment. To classify a query sample, its feature is extracted via the trained feature encoder and compared against the features of all support set samples using a similarity metric such as cosine or Euclidean distance. The class label of the closest support sample is then assigned to the query. This enables few-shot inference in the target domain with only a handful of labeled examples. Furthermore, if the support set is constructed using only training data without any target-domain samples, the model can perform zero-shot inference, which makes it particularly suitable for domain generalization scenarios.

\paragraph{Metric Learning with Triplet Loss}

Several works have leveraged triplet loss, as illustrated in Fig.~\ref{fig:triplet-loss}, for cross-domain few-shot Wi-Fi sensing. For example, LT-WiOB~\cite{2-28zhou2022enabling} and CATFSID~\cite{2-53wei2024catfsid} adopt the classic triplet loss strategy to train feature extractors that pull embeddings of intra-class samples closer while pushing those of inter-class samples further apart. During inference, a support set is constructed using a few labeled samples per class from the target domain. The test sample is then classified by computing distances to the support set embeddings and assigning the label of the nearest neighbor, thereby enabling cross-environment activity recognition and user identification. Similarly, TransferSense~\cite{1-15bu2022transfersense} employs triplet loss for cross-domain gait recognition, where the support set consists of a few samples from previously unseen users in a new environment, enabling few-shot identification based on walking patterns.

While most approaches use the standard Euclidean distance to compare embeddings, recent methods explore alternative similarity measures to improve robustness. UniFi~\cite{2-49liu2024unifi} introduces a mutual information-based loss~\cite{belghazi2018mine} to enhance discrimination between positive and negative pairs, facilitating cross-scene, cross-location, and cross-orientation activity recognition. WiOpen~\cite{2-54zhang2024wiopen} adopts a metric learning strategy based on neighbor component analysis (NCA)~\cite{wu2018improving}, which minimizes intra-class distances while maximizing inter-class separability, thereby improving few-shot classification on unseen gesture classes.

Beyond handling domain and user shifts, triplet-loss-based models have also been applied to cross-modality recognition. For instance, SFTSeg~\cite{2-37xiao2022self} trains a model on the IMU-based HandGesture dataset~\cite{bulling2014tutorial} and transfers it to the Wi-Fi-based WiFiAction dataset~\cite{xiao2020deepseg}, achieving few-shot recognition across sensing modalities.

\paragraph{Learning with Contrastive Learning}

Contrastive learning also relies on constructing positive and negative sample pairs, and often incorporates data augmentation to expand the training set. Unlike triplet loss, where each training iteration involves a single positive and negative pair, contrastive learning considers all pairwise relationships within a mini-batch. Specifically, for a mini-batch of $n$ samples, an $n \times n$ label matrix $X = {x_{i,j}}$ is constructed, where each element $x_{i,j}$ indicates whether sample $i$ and sample $j$ belong to the same class (1) or not (0). The diagonal elements are always 1, as each sample matches itself. The loss is typically computed as a row-wise or column-wise cross-entropy over this similarity matrix, enabling the model to learn class-discriminative embeddings across all pairs in the batch. During inference, a query sample is compared with support samples in the feature space, and the label of the most similar support sample is assigned as the prediction result.

Several recent works have demonstrated the effectiveness of contrastive learning for few-shot and cross-domain Wi-Fi sensing tasks.
DualConFi\cite{2-167xu2022dual} applies a contrastive learning framework to enable robust activity recognition across different environments.
CFLH~\cite{2-166wang2024csi} enhances the diversity of positive sample pairs by applying random scaling augmentations to Wi-Fi signals, and then trains the model using contrastive loss to enable few-shot recognition of unseen activities in new environments.
Xiao et al.~\cite{2-56xiao2024diffusion} further integrate diffusion-based generative modeling (e.g., Denoising Diffusion Probabilistic Models~\cite{ho2020denoising}) to synthesize additional positive samples. These synthetic samples are incorporated into a contrastive training pipeline, which improves the model’s ability to recognize novel activities in unseen environments using limited labeled data.

\paragraph{Metric Learning with Siamese Neural Networks}

\begin{figure}[t]
    \centering
    \includegraphics[width=1\linewidth]{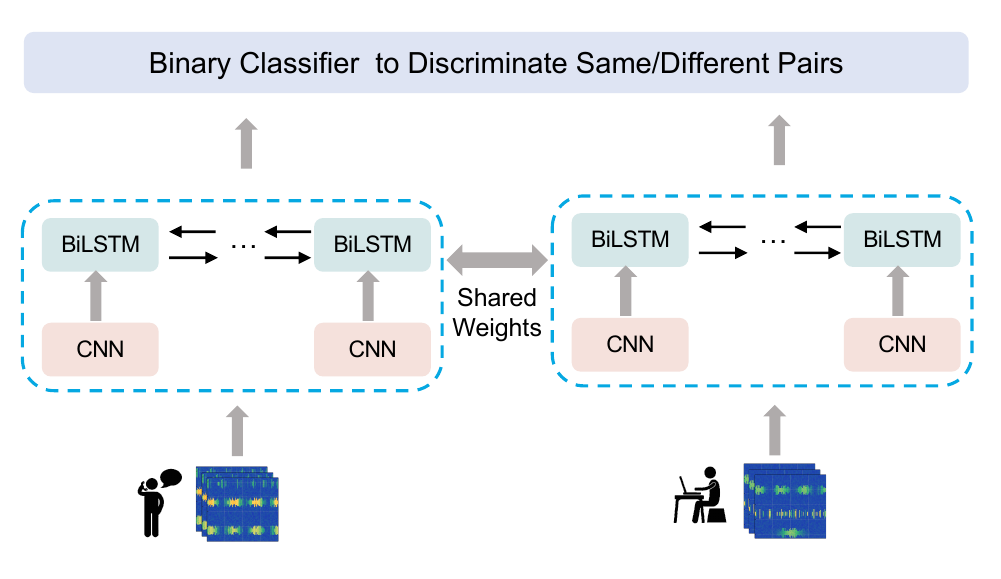}
    \caption{Siamese Neural Networks. (1) The network takes a pair of inputs (e.g., two CSI segments) and feeds them into two identical sub-networks. Both branches share the same parameters (weights), mapping the inputs into a common low-dimensional feature space. (2) The model computes a similarity metric (usually Euclidean distance or Cosine similarity) between the two generated feature vectors. (3) Based on the distance, the system determines whether the two inputs belong to the same class (small distance) or different classes (large distance), often using a contrastive loss function.} 
    \label{fig:SiameseNet}
\end{figure}

Siamese neural networks~\cite{koch2015siamese} offer a classical yet effective architecture for few-shot learning and domain generalization. Unlike contrastive learning, which refers to a training paradigm defined by similarity-based objectives (e.g., contrastive or triplet loss), Siamese networks represent a weight-sharing architecture that is commonly used to implement such metric learning objectives.
In this framework, as shown in Fig.~\ref{fig:SiameseNet}, a pair of samples is fed into two identical networks that share weights. Each network extracts feature embeddings from its input, and the resulting embeddings are concatenated and passed through a binary classifier. The classifier is trained to predict whether the two inputs belong to the same class: if they do (e.g., two samples of the same activity), the target label is 1; otherwise, it is 0. During inference, a support set is constructed from a few labeled samples per class in the target domain. To classify a query sample, it is paired with each support sample and processed by the Siamese network. The network outputs a confidence score indicating the likelihood of the two samples belonging to the same class. The class label of the support sample with the highest confidence score is then assigned to the query sample. 

This pipeline has been widely adopted in Wi-Fi sensing tasks for few-shot cross-domain generalization. For instance, 3D-ID~\cite{2-76ren2022robust} utilizes this pipeline for few-shot user identification in unseen environments. MaP-SGAN~\cite{2-50liang2024map} applies it for gait recognition by enhancing the diversity of training samples via generative models. CrossFi~\cite{2-44zhao2024crossfi} extends the Siamese framework to handle both cross-domain and new-class scenarios, enabling gesture recognition and user identification in previously unseen domains.

\paragraph{Metric Learning with Relation Network}

\begin{figure}[t]
    \centering
    \includegraphics[width=1\linewidth]{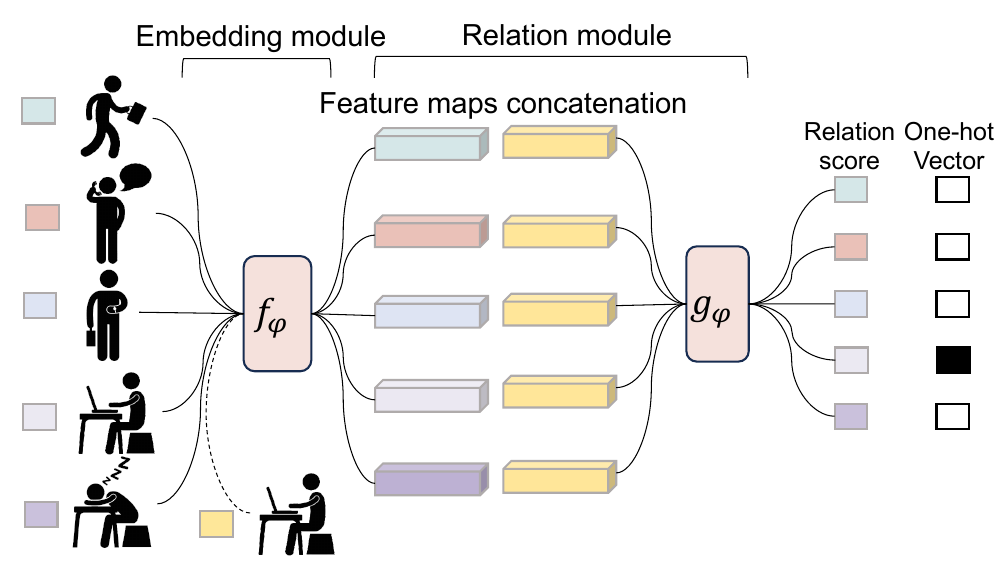}
    \caption{Relation Network. (1) Both the support set and the query sample are passed through an embedding backbone to extract their respective feature maps. (2)The query feature map is paired and concatenated with each feature map in the support set, creating combined feature representations. (3) These concatenated pairs are fed into a Relation Module (usually a non-linear CNN or MLP), which acts as a learnable metric to calculate a relation score (between 0 and 1) for each pair. (4) The query is assigned to the class with the highest relation score, representing the strongest similarity determined by the deep neural network rather than a fixed distance metric.} 
    \label{fig:RelationNet}
\end{figure}

Relation networks~\cite{sung2018learning} provide a powerful framework for few-shot learning by explicitly modeling the similarity between samples through deep relational reasoning. As illustrated in Fig.~\ref{fig:RelationNet}, training involves two components: a support set and a query set. The support set contains one labeled sample per class, while the query set contains a sample whose class is to be predicted. Both support and query samples are passed through a shared feature extractor. The resulting feature vectors are then concatenated and passed to a relation module (typically a neural network), which outputs a similarity score indicating how likely the two samples belong to the same class. The network is optimized using a multi-class cross-entropy loss, where the ground-truth label is encoded in a one-hot format: for each query sample, only the score corresponding to the matching support class is set to 1, while all others are 0. During inference, a test support set and a query sample are fed into the model. The network outputs relation scores between the query and each support sample. The class corresponding to the support sample with the highest relation score is then selected as the prediction result.

Wang et al.~\cite{2-146wang2020learning} follow the above pipeline to enable few-shot activity recognition in unseen environments. WiGR~\cite{2-57hu2021wigr} adopts relation networks and trains the model on 100 action classes. To evaluate generalization to novel classes, the model is tested on previously unseen activities, using only three support samples per new class. LESS~\cite{2-132zhang2023learning} also uses relation networks ito achieve few-shot cross-environment fingerprint-based localization. WiGesID~\cite{2-8zhang2022wi} further evaluates this framework to simultaneously support cross-environment action recognition and user identification. Besides,  WiGesID also supports few-shot classification of both new actions and unseen individuals, showcasing the flexibility of relation networks in multi-task and cross-domain scenarios.

\paragraph{Metric Learning with Matching Network}

\begin{figure}[t]
    \centering
    \includegraphics[width=1\linewidth]{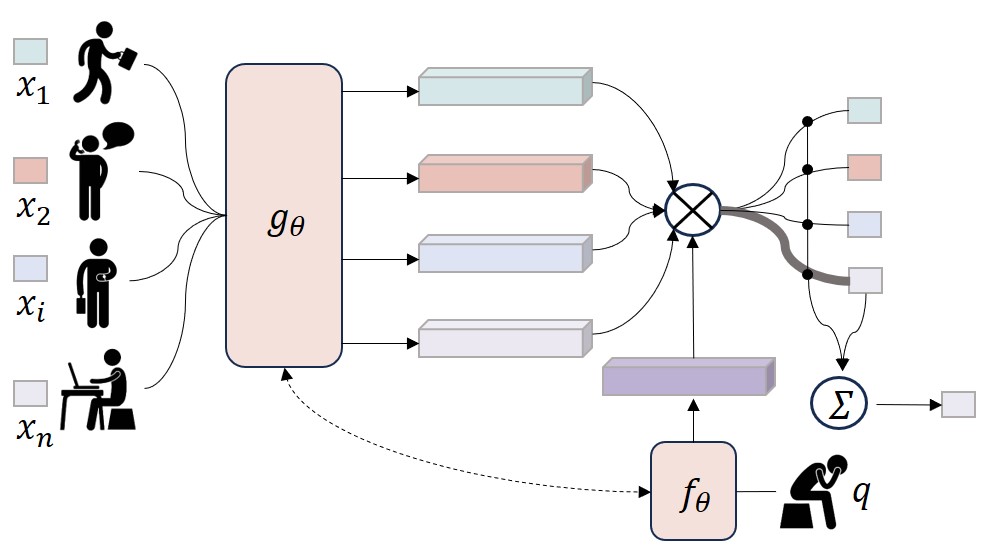}
    \caption{Matching Network. (1) Both the small support set and the query sample are mapped into a shared low-dimensional feature space using a neural network.
(2) The model computes the cosine similarity (or distance metric) between the query embedding and each embedding in the support set.
(3) These similarity scores are processed through a softmax function to generate attention weights, indicating which support examples most closely resemble the query.
(4) The final label is predicted by a weighted sum of the support set labels, allowing the model to recognize new classes without any parameter updates.} 
    \label{fig:MatchingNet}
\end{figure}

Matching networks~\cite{vinyals2016matching} provide an effective framework for few-shot learning by leveraging attention mechanisms and episodic training to mimic test-time inference. As illustrated in Fig.~\ref{fig:MatchingNet}, during training, each episode includes a support set $S=\{(x_i,y_i)|i=1,2,...,n\}$, where $x_i$ is a sample, $y_i$ is  its corresponding label, and $q$ is a query sample. The support set consists of a few labeled samples from each class, while the query sample is to be classified. 

The predicted label of the query sample is computed as a weighted sum of the support set labels, where the weights are derived from the similarity between the query and each support sample:
\begin{equation}\label{eq:matching-network}
\begin{aligned}
    a(q,x_i) &= {e^{c(f(q),g(x_i))}}/{\sum_{j=1}^{n}e^{c(f(q),g(x_j))}} \\
        \hat{y} &= \sum_{i=1}^{n}a(q,x_i)y_i 
\end{aligned}
\end{equation}
Here, $c(f(q),g(x_i))$ computes the cosine similarity between the feature of the query sample and that of a support sample; The similarity scores are passed through a softmax function to produce the weights $a(q,x_i)$, which reflect the relative similarity between $q$ and each $x_i$. The final prediction $ \hat{y}$ is obtained by weighting the support labels accordingly. As shown in the Equation.~\ref{eq:matching-network}, samples more similar to the query contribute more significantly to the prediction.

Building on matching networks~\cite{vinyals2016matching}, AFEE-MatNet~\cite{2-128shi2022environment} and HAR-MN-EF~\cite{1-19shi2020towards} enhance domain generalization by training on data from four distinct environments and evaluating few-shot activity recognition in three unseen environments. MatNet-eCSI~\cite{2-21shi2020environment}, WiLiMetaSensing~\cite{2-6ding2021wifi-based}, and Ding et al.~\cite{2-122ding2021improving} also apply matching networks to achieve cross-environment activity recognition, where each class in the support set contains one single labeled instance during inference.

\paragraph{Metric Learning with Prototypical Network}

\begin{figure}[t]
    \centering
    \includegraphics[width=1\linewidth]{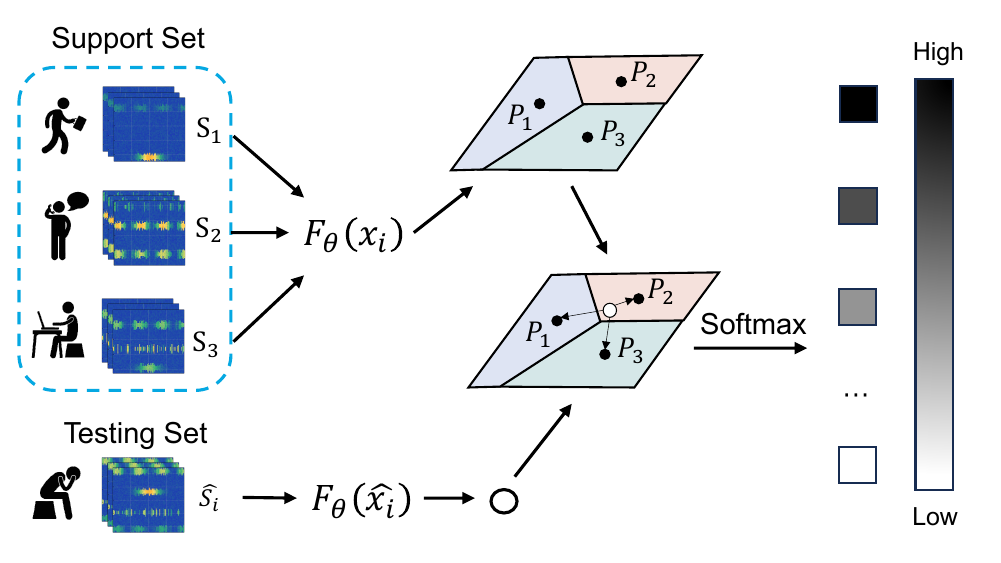}
    \caption{Prototypical Network. (1) For each class in the support set, the model computes a prototype (mean vector) by averaging the feature embeddings of all samples belonging to that class. (2) The query sample is mapped into the same feature space using the shared embedding backbone. (3) The model calculates the Euclidean distance between the query embedding and each class prototype. (4) The query is classified based on a softmax distribution over the negative distances; it is assigned to the class whose prototype is nearest in the embedding space.} 
    \label{fig:PrototypeNet}
\end{figure}

Prototypical networks~\cite{finn2017model} provide a simple yet effective approach for few-shot learning by representing each class with a prototype — the mean vector of embedded support samples. As shown in Fig.~\ref{fig:PrototypeNet}, training follows an episodic paradigm. In each episode, $K$ classes are randomly sampled, with $N$ labeled samples per class forming the support set. Each sample is passed through a shared feature encoder to extract embeddings. For each class $i$, its prototype $c_i$ is computed as the mean of the support embeddings:
\begin{equation}\label{eq:mean-prototype}
c_i = \frac{1}{N} \sum_{j=1}^{N} F_{\theta}(x_{i}^{j})
\end{equation}
where $F_{\theta}(x_{i}^{j})$ denotes the embedding of the $j$-th sample from class $i$. To classify a query sample $q$, its embedding is compared to all class prototypes using a distance metric (typically Euclidean distance):
\begin{equation}\label{eq:prototypical-network}
    p(y=i|q) = {e^{-d(F_{\theta}(q),c_i)}}/{\sum_{j=1}^{K}e^{-d(F_{\theta}(q),c_j))}}
\end{equation}
The resulting softmax probabilities over distances $d$ serve as similarity scores between the query and each class prototype. The class with the highest score is the predicted label.

Several Wi-Fi sensing studies have adopted prototypical networks for few-shot cross-domain Wi-Fi sensing. For example,  ReWiS~\cite{2-26bahadori2022rewis} applies prototypical networks to compute class centroids using support samples and performs classification by measuring the Euclidean distance between the query and each class prototype, enabling cross-environment few-shot activity recognition. Similarly, Ding et al.~\cite{2-123ding2021device} use Euclidean distance to facilitate few-shot activity recognition across different locations within the same environment. For user identification tasks, WiONE~\cite{2-126gu2021wione} computes prototype representations of each user’s identity and classifies test samples based on the squared Euclidean distance to these prototypes, enabling cross-environment one-shot user authentication. CAUTION~\cite{2-130wang2022caution} adopts a similar Euclidean-distance-based framework for few-shot user authentication across scenes.

In addition to Euclidean distance, cosine similarity has also been used to compute distances between query samples and class prototypes. For instance, FewCS~\cite{2-4yang2023few}, Zhou et al.~\cite{1-27zhao2024functional}, and Zhang et al.~\cite{1-6zhang2021wifi} utilize cosine similarity to measure similarity, achieving robust few-shot activity recognition across domains such as environment, user, location, and orientation.

\paragraph{Summary and Design Insights for Metric Learning} 
Metric learning approaches share a common objective of structuring the embedding space to enable instance-level comparison rather than fixed classifier-based prediction, making them naturally suited for zero-shot and few-shot Wi-Fi sensing scenarios. 
Unlike domain alignment methods that primarily focus on reducing distribution discrepancy, metric learning emphasizes relational structure within the feature space, allowing the model to generalize to new domains or classes through support-set-based inference.

Among different paradigms, triplet-loss and contrastive-learning-based methods directly optimize pairwise relationships and often produce highly discriminative embeddings. 
They are flexible and compatible with various backbone architectures, but their performance depends on effective sample mining strategies and sufficient diversity within mini-batches. 
Siamese networks provide a straightforward architectural implementation of similarity learning, offering interpretability and ease of deployment, though they may suffer from scalability issues when the number of support comparisons grows. 

Relation networks extend metric learning by introducing a learnable non-linear similarity function, enabling more expressive relational reasoning beyond fixed distance metrics. 
This improves adaptability to complex intra-class variations but increases computational overhead and model complexity. 
Matching networks and prototypical networks, in contrast, adopt episodic training to mimic test-time inference. 
Prototypical networks are particularly attractive due to their simplicity, computational efficiency, and strong performance under limited support samples, while matching networks provide a more flexible attention-based weighting mechanism at the cost of increased inference complexity.

From a design perspective, the choice of metric learning strategy should consider (i) the expected number of support samples per class, (ii) whether novel classes or only novel domains are encountered, (iii) computational constraints at inference time, and (iv) the degree of intra-class variability. 
Prototypical networks are well-suited for practical few-shot deployment with limited support data and strict efficiency requirements. 
Contrastive and triplet-based methods offer stronger embedding discrimination when large-scale pretraining is feasible. 
Relation-based models are preferable when complex relational reasoning is required, such as simultaneous cross-domain and new-class generalization.

\begin{figure*}[t]
    \centering
    \includegraphics[width=1\textwidth]{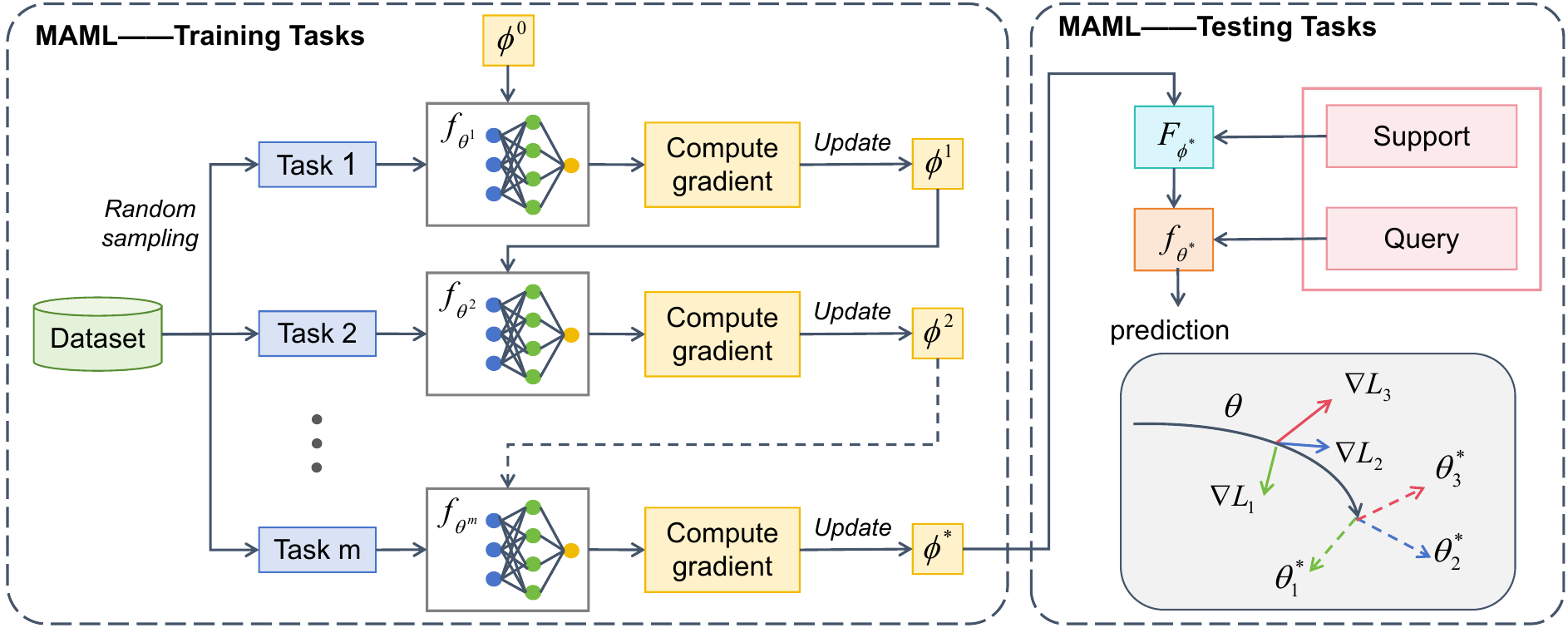}
    \caption{Model-Agnostic Meta-Learning (MAML)~\cite{finn2017model} includes a meta-training stage and a meta-testing stage. The meta-training stage is designed to find a good initialization of model parameters such that the model can quickly adapt to new tasks with only a few gradient steps and a small amount of labeled data in the meta-testing stage.}
    \label{fig:maml}
\end{figure*}

\vspace{5pt}

\subsubsection{\textbf{Meta Learning for Few-shot Learning}}\label{sec:meta-learning}

Meta learning aims to enable models to ``learn how to learn'', and represents a distinct paradigm from metric learning approaches described in Section~\ref{sec:metric-learning}. 
While both are designed for few-shot learning, metric learning focuses on constructing an embedding space where intra-class samples are pulled closer and inter-class samples are pushed apart, thereby allowing direct inference based on similarity measures. In contrast, meta learning emphasizes learning an optimal initialization or training strategy that can rapidly adapt to new tasks with only a few labeled samples. Model-Agnostic Meta-Learning (MAML)~\cite{finn2017model} is one of the most representative and widely adopted approaches.

As shown in Fig.~\ref{fig:maml}, MAML employs a two-level optimization framework. In each meta-training stage, a batch of tasks is sampled from $p(\mathcal{T})$. For each task $\mathcal{T}_i$, the model first performs a few gradient updates using the task-specific support set, leading to adapted parameters $\theta_i$:
\begin{equation}\label{eq:Inner-loop}
    \theta_i = \phi - \alpha \nabla_\phi \mathcal{L}_{\mathcal{T}_i}^{\text{support}}(\phi)
\end{equation}
where $\phi$ is the meta-model's parameter, $\alpha$ is the inner-loop learning rate, and $\mathcal{L}_{\mathcal{T}_i}^{\text{support}}$ is the task-specific loss.

Then, the performance of $\theta_i$ is evaluated on the corresponding query set of each task, and the meta-objective is to optimize the model parameters $\phi$ so that they perform well after the inner-loop update:
\begin{equation}\label{eq:Outer-loop}
    \phi \leftarrow \phi - \beta \nabla_\phi \sum_{\mathcal{T}_i \sim p(\mathcal{T})} \mathcal{L}_{\mathcal{T}_i}^{\text{query}}(\theta_i)
\end{equation}
where $\beta$ is the meta (outer-loop) learning rate. This outer-loop update accumulates the gradients across multiple tasks, ensuring that the learned initialization $\phi$ is broadly effective across different domains.

During meta-testing, the model receives a new task (e.g., a new environment in Wi-Fi sensing) and is fine-tuned using just a few labeled support set samples. Thanks to the meta-learned initialization, the model can quickly converge to an effective solution with only a few gradient steps, achieving robust performance under cross-domain few-shot settings.

Huang et al.~\cite{2-47huang2022few} and TOSS~\cite{1-13zhou2022target} employ the original MAML~\cite{finn2017model} framework to perform few-shot activity recognition across different environments. MetaLoc~\cite{2-125gao2023metaloc} leverages MAML for cross-scene indoor localization, demonstrating the generality of meta-learning across sensing modalities.

Beyond vanilla MAML, several studies propose MAML-like adaptations to better suit specific tasks in Wi-Fi sensing. MetaFormer~\cite{1-47sheng2024metaformer} introduces a MAML-inspired training strategy combined with Transformer backbones, achieving one-shot cross-user and cross-environment activity recognition. OneSense~\cite{1-48zhao2024one} and Wi-learner~\cite{2-89feng2022wi} adopt similar MAML-like frameworks to handle domain shifts caused by varying environments, locations, and orientations. RF-Net~\cite{1-2ding2020rf} also applies a MAML-style strategy for one-shot activity recognition in unseen environments.

In addition, researchers have explored more recent meta-learning paradigms to improve few-shot cross-domain performance. ML-WiGR~\cite{2-75gao2022ml} incorporates Reptile~\cite{nichol2018first}, a first-order approximation to MAML, enabling efficient meta-training without second-order gradients. MetaGanFi~\cite{2-3zhang2022metaganfi} adopts the BOIL~\cite{oh2020boil} method, which focuses meta-learning on the feature extractor rather than the task-specific head. Furthermore, Zhang et al.~\cite{2-131zhang2020human} utilize Learning to Learn~\cite{andrychowicz2016learning}, a meta-optimizer-based approach, to support cross-domain activity recognition. From a networking perspective, meta-learning typically incurs limited communication overhead during deployment, as model adaptation is performed locally using a small number of samples. However, the meta-training phase is often computationally intensive and may introduce higher offline training latency, especially when learning across a large number of tasks or domains.

\vspace{5pt}
\subsubsection{\textbf{Data Augmentation and Data Synthesis}}\label{sec:data-augmentation}

Data scarcity and limited diversity are key challenges that hinder Wi-Fi sensing generalizability. To address these issues, a growing number of studies leverage data augmentation and synthetic data generation. Notably, data generation is often used in conjunction with other methods, such as domain adaptation, metric learning, or meta-learning, serving as a plug-in to enhance overall sensing robustness.

Generative models like GANs, VAEs, and diffusion models are commonly employed to synthesize Wi-Fi signal data. For example:
CrossGR~\cite{1-9li2021crossgr}, Wang et al.~\cite{2-146wang2020learning},  SIDA~\cite{2-51zhang2023sida}, and WiARGAN~\cite{2-64huang4160593low} use GANs~\cite{goodfellow2014generative} to generate synthetic CSI data and increase data diversity.
AFSL-HAR~\cite{2-24wang2021robust} and LESS~\cite{2-132zhang2023learning} adopt Wasserstein GANs~\cite{gulrajani2017improved} to create more stable and realistic samples.
Wi-Dist~\cite{2-71zhang2024toward} and CsiGAN~\cite{2-22xiao2019csigan} utilize CycleGAN~\cite{zhu2017unpaired} for domain translation and data augmentation.
Fidora~\cite{1-3chen2022fidora} and Fido~\cite{2-135chen2020fido} employ Variational Autoencoders (VAEs) to synthesize location fingerprints.
freeLoc~\cite{1-36FreeLoc} and freeGait~\cite{2-45yan2024freegait} use Adversarial Autoencoders (AAEs)~\cite{makhzani2015adversarial} to generate location and gait-related signals.
Xiao et al.~\cite{2-56xiao2024diffusion} apply a diffusion model~\cite{ho2020denoising} to produce a large number of realistic samples.

Besides generative models, various non-generative strategies are adopted to expand the training set and improve generalization: Zhou et al.~\cite{1-49zhou2022towards} uses an autoencoder to synthesize action samples with one sample per action in the target domain. 
CSI-Net~\cite{wang2018csi} augments data by combining same-class samples to introduce intra-class variability. 
BullyDetect~\cite{lan2024bullydetect} applies time-series transformations such as window wrapping and window slicing to increase data volume.
P$^3$ID~\cite{1-37P3ID} uses magnitude warping and time permutation to augment training data.
OpenFi~\cite{2-33OpenFi} generates virtual samples with magnitude warping and Cutout~\cite{devries2017improved} to train and recognize unseen actions and identities.
AirFi~\cite{1-10wang2022airfi} adds arbitrary Gaussian noise to create variations in the training data.
RFBoost~\cite{2-60hou2024rfboost} proposes frequency- and time-domain augmentations to enhance signal robustness.
WiSGP~\cite{1-21liu2023generalizing} leverages domain gradient-based augmentation to generate samples with enhanced domain diversity.
WiAG~\cite{2-129virmani2017position} generates gesture variants through linear and nonlinear signal transformations.

When limited target domain data is available, augmentation strategies can be used to synthesize domain-aligned samples:
Wi-Cro~\cite{2-52mao2024wi} uses CycleGAN~\cite{zhu2017unpaired} to generate synthetic samples mimicking the target domain distribution, mitigating domain shift.
DiRA~\cite{wang2025generative2} develops a robust data augmentation scheme using conditioned diffusion models~\cite{dhariwal2021diffusion} to expand both the quantity and quality of wireless sensing data, effectively enhancing the accuracy and robustness of sensing models where data is insufficient or unevenly distributed.
SFTSeg~\cite{2-37xiao2022self} employs a sliding window strategy to segment Wi-Fi signals in target domain and takes segmented signals to augment training data.
OneFi~\cite{1-44xiao2021onefi} simulates gesture rotations to generate direction-variant samples for cross-orientation recognition.
OneSense~\cite{1-48zhao2024one} generates new samples by composing gesture primitives based on signal propagation priors, e.g., combining the hand-drawing ``L" gesture and the hand-drawing ``I" gesture to simulate the hand-drawing ``U" gesture, thereby enabling recognition of unseen gestures.
Wang et al.~\cite{2-81wang2024feature} propose a feature decoupling approach that employs two separate classifiers to disentangle identity and gesture features. The model extracts identity-specific and gesture-specific representations independently, and then combines the gesture feature with a target user's identity representation to generate synthetic gesture samples for that specific user for gesture recognition.

\vspace{5pt}

\subsubsection{\textbf{Pseudo Labeling}}\label{sec:pseudo-labeling}

Pseudo-labeling, particularly the assignment of virtual labels to unlabeled target domain samples, is an effective strategy to increase training data volume, enhance model generalization to unseen domains, and reduce manual annotation costs. This approach typically involves an initial feature extractor or classifier trained on source domain data, which is then used to generate labels for the target domain samples.

CLAR~\cite{2-32zhou2022device} first trains multiple classifiers on labeled source data and uses them to predict class probabilities for each target domain sample. These predictions are aggregated—typically via weighted averaging—to produce a pseudo-label. The resulting pseudo-labeled target samples are combined with labeled source data to train a unified model, thereby enabling cross-location activity recognition. LAGER~\cite{1-32Chen_LAGER} further refines the pseudo-labeling process by recognizing the unreliability of early pseudo-labels. It introduces a two-stage pseudo-labeling framework: after initial label assignment, the model computes the centroid of each class in the feature space, and then reassigns labels based on the distance between each target sample and the respective class centroids. This centroid-based refinement improves pseudo-label accuracy and ultimately enhances cross-domain action recognition performance.

Other works, such as Zhang et al.\cite{1-14zhang2023unsupervised}, DA-HAR\cite{2-69sheng2023har}, and Zhang et al.~\cite{2-100zhang2023unsupervised}, adopt a similar strategy. They train a classifier on the source domain and use it to assign pseudo-labels to target domain samples. However, only samples with high prediction confidence (e.g., exceeding thresholds such as 0.92 or 0.8) are selected for pseudo-labeling. These high-confidence pseudo-labeled samples are then used, together with the source domain data, to retrain the classifier. The updated model generates new pseudo-labels in an iterative process, gradually improving performance on the target domain.

\begin{table*}[htbp]
\centering
\caption{Comparison of Pros, Cons, and Best-fit Scenarios of Feature Learning Techniques in Wi-Fi Sensing.}
\label{tab:feature_learning_comparison}
\small
\begin{tabularx}{\textwidth}{p{5cm}XXXX}
\toprule
\textbf{Category} 
& \textbf{Pros (Advantages)} 
& \textbf{Cons (Limitations)} 
& \textbf{Generalization Mechanism} 
& \textbf{Best-fit Scenarios} \\
\midrule

Domain Alignment (\cite{1-17wang2021environment, 2-69sheng2023har, 1-5zhang2021privacy, 1-18zhou2024sample, 2-2zou2018joint, 1-9li2021crossgr, 2-30zhang2021wi, jiang2018towards, 2-72khattak2022cross, 1-11wang2019wicar, 2-168zhang2024objective, 2-55zinys2021domain, 2-155strohmayer2024datta, 1-29liu2023wisr, 2-98yin2021towards, 2-48yang2024domain, 2-78qin2023cross, 2-68berlo2023use, 1-31ZhouSubject-independent, 2-45yan2024freegait, 1-7li2021dafi, 2-142shi2020towards, 1-3chen2022fidora, 1-22zhou2024adapose, 2-96rao2024novel, 1-12kang2021context, 2-109liang2023wiai, 2-107zhang2023device, 2-119mehryar2023domain, 2-151zhou2020adaptive, 2-77xiao2024pattern, kang2019contrastive, 1-34WiCAU, 2-1jiao2024wisda, 1-30gong2024privacy, 1-24sheng2024cdfi, 1-32Chen_LAGER, 2-67zhan2024indoor, 1-10wang2022airfi, 2-136ding2020device, 2-3zhang2022metaganfi, 2-61zhang2022cross, 2-52mao2024wi, 2-39xiao2023mean, 1-50zhang2018crosssense, 2-59sugimoto2023towards, 1-28zhang2023location, 2-29islam2020wi, wang2024xrf55})
& Mitigates domain and hardware shifts; statistical metrics provide theoretical grounding. 
& Risk of suppressing discriminative motion features during aggressive alignment. 
& Enforces distribution-level invariance by aligning feature statistics across domains (e.g., adversarial learning, MMD). 
& Large domain shifts (e.g., cross-room or cross-device) with labeled source and unlabeled target data. \\

\midrule
Disentanglement (\cite{1-42elujide2022location, 2-103chen2024task, 2-154wang2019person, 2-80hao2021wi, 2-18wu2021device})
& Strong physical interpretability; separates motion features from environment or identity noise. 
& Requires complex architectures (e.g., GANs or autoencoders) and careful hyperparameter tuning. 
& Factorizes latent representations into independent components to isolate task-relevant signals. 
& Scenarios with multiple interference factors (e.g., identical gestures by different users or locations). \\

\midrule
Metric Learning (\cite{2-28zhou2022enabling, 2-53wei2024catfsid, 1-15bu2022transfersense, 2-49liu2024unifi, 2-54zhang2024wiopen, wu2018improving, 2-37xiao2022self, 2-167xu2022dual, 2-166wang2024csi, 2-56xiao2024diffusion, 2-76ren2022robust, 2-50liang2024map, 2-44zhao2024crossfi, 2-146wang2020learning, 2-57hu2021wigr, 2-132zhang2023learning, 2-8zhang2022wi, 2-128shi2022environment, 1-19shi2020towards, 2-21shi2020environment, 1-27zhao2024functional, 2-6ding2021wifi-based, 2-26bahadori2022rewis, 2-123ding2021device, 2-126gu2021wione, 2-130wang2022caution, 2-4yang2023few, 1-6zhang2021wifi})
& Native few-shot and zero-shot support; scalable to new classes via template updates. 
& Highly dependent on support set quality; risk of class overlap in the embedding space. 
& Structures the embedding space by minimizing intra-class distance and maximizing inter-class separability. 
& Scenarios with extreme scarcity of target labels or dynamic needs for novel class recognition. \\

\midrule
Meta Learning (\cite{2-47huang2022few, 1-13zhou2022target, 2-125gao2023metaloc, 1-47sheng2024metaformer, 1-48zhao2024one, 2-89feng2022wi, 1-2ding2020rf, 2-75gao2022ml, 2-3zhang2022metaganfi, 2-131zhang2020human})
& Rapid adaptation to new tasks; learns an initialization for fast convergence. 
& High computational cost during offline meta-training (e.g., second-order gradients). 
& Optimizes model parameters for fast task-level adaptation through episodic training across domains. 
& Edge deployment requiring immediate high accuracy after minimal user calibration. \\

\midrule
Data Synthesis (\cite{1-9li2021crossgr, 2-146wang2020learning, 2-51zhang2023sida, 2-64huang4160593low, 2-24wang2021robust, 2-132zhang2023learning, 2-71zhang2024toward, 2-22xiao2019csigan, 1-3chen2022fidora, 2-135chen2020fido, 1-36FreeLoc, 2-45yan2024freegait, 2-56xiao2024diffusion, 1-49zhou2022towards, wang2018csi, lan2024bullydetect, 1-37P3ID, 2-33OpenFi, devries2017improved, 1-10wang2022airfi, 2-60hou2024rfboost, 1-21liu2023generalizing, 2-129virmani2017position, 2-52mao2024wi, wang2025generative2, 2-37xiao2022self, 1-44xiao2021onefi, 1-48zhao2024one, 2-81wang2024feature})
& Directly mitigates data scarcity; generative models improve coverage of rare cases. 
& Potential sim-to-real gap between synthetic samples and real-world signal data. 
& Expands training distributions by synthesizing diverse or unseen signal patterns. 
& Plug-in augmentation for tasks with insufficient, imbalanced, or hard-to-collect labeled data. \\

\midrule
Pseudo Labeling (\cite{2-32zhou2022device,1-32Chen_LAGER,1-14zhang2023unsupervised,2-69sheng2023har,2-100zhang2023unsupervised}) 
& Leverages abundant unlabeled target data via self-training. 
& Risk of error accumulation from early incorrect pseudo-labels. 
& Iteratively refines decision boundaries using confident predictions as surrogate labels. 
& Semi-supervised sensing tasks with long-term accumulation of unlabeled data. \\

\bottomrule
\end{tabularx}
\end{table*}

Table~\ref{tab:feature_learning_comparison} compares the advantages and limitations of representative feature learning strategies in Wi-Fi sensing, explains their underlying generalization mechanisms, and summarizes the scenarios in which each approach is best suited. 
The analysis of the feature learning stage highlights several strategic insights, as highlighted in the Key Takeaways and Lessons learned below, for developing robust and cross-domain representations, following the technical progression outlined in Section \ref{sec:Feature Learning Stage}. 

\begin{tcolorbox}[
    colback=hotPink!5!white,
    colframe=hotPink!75!black,
    title=\textbf{Key Takeaways and Lessons Learned: Feature Learning},
    width=0.48\textwidth,
    boxrule=0.8pt,
    arc=2mm,
    left=2mm,
    right=2mm,
    top=1mm,
    bottom=1mm,
    breakable
]

\begin{itemize}
    \item \textbf{Domain Alignment for Distribution Matching:} Minimizing domain divergence via adversarial discriminators or statistical metrics (e.g., MMD, EMD) is fundamental. A key lesson is using cross-modal embeddings (e.g., BERT, GPT-4o) as ``semantic anchors" to bridge the gap between raw CSI and high-level human activities.
    \item \textbf{Component Disentanglement for Interpretability:} Explicitly separating environment, identity, and motion components using GANs or specialized encoders ensures that the model ignores environmental noise. This disentangled latent space is critical for handling multi-factor variations in real-world settings.
    \item \textbf{Metric Learning for Few-shot Scalability:} Structuring the feature space using triplet loss, contrastive learning, or prototypical networks allows for clustering similar actions. The major takeaway is the high scalability of similarity-based matching, which enables zero/few-shot recognition of novel classes without retraining the backbone.
    \item \textbf{Meta-Learning for Rapid Adaptation:} Paradigms like MAML and Reptile focus on ``learning how to learn" by optimizing for a superior parameter initialization. This ensures that the sensing system can converge to a site-specific optimum with minimal gradient steps (e.g., one-shot or few-shot).
    \item \textbf{Data Synthesis as a Robustness Plug-in:} Beyond simple augmentations, generative models (GANs, VAEs, and Diffusion) are essential for simulating target-domain conditions. These modules mitigate the Sim-to-Real gap by expanding training diversity with physically plausible synthetic signals.
    \item \textbf{Pseudo Labeling for Iterative Refinement:} Leveraging abundant unlabeled target data through virtual labels reduces annotation costs. The iterative refinement of these labels (e.g., via confidence thresholds or centroid-based reassignment) is vital for long-term performance improvement in unfamiliar domains.
\end{itemize}
\end{tcolorbox}

\begin{figure*}[t]
    \centering
    \includegraphics[width=1\textwidth]{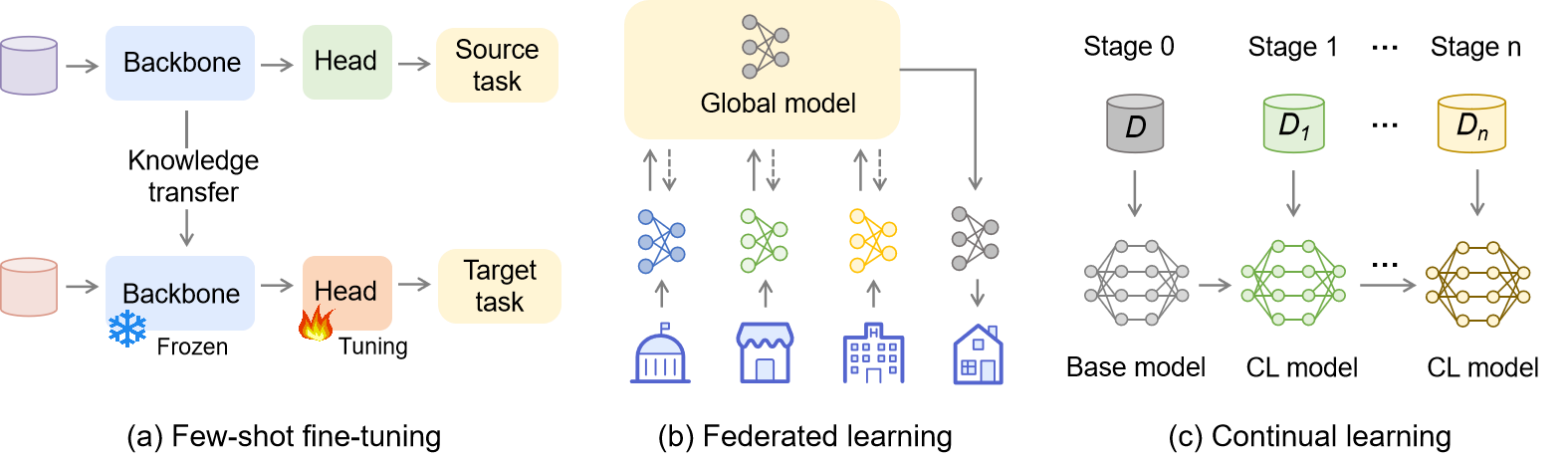}
    \caption{Wi-Fi sensing generalization methods in the deployment stage. (a) Few-shot fine-tuning. In this approach, a Wi-Fi sensing network is first pre-trained on a large-scale dataset to learn general feature extraction capabilities. During deployment, the parameters of the backbone are kept frozen, while only the task head is fine-tuned using data from the target scenario. This strategy enables rapid deployment in new environments with minimal computational overhead.
(b) Federated Learning. This method involves aggregating data from diverse scenarios for centralized training on the server side. By exposing the model to a sufficiently wide range of environments during the training phase, it learns to capture robust, cross-domain features. Consequently, the model can be deployed into target scenarios, achieving seamless zero-shot generalization.
(c) Continual Learning. To handle the continuous influx of new sensing tasks and environmental data, the model evolves through sequential training stages as new data arrives, allowing it to incrementally acquire knowledge from novel scenarios. This approach ensures the model remains adaptive to dynamic environmental changes while mitigating the risk of catastrophic forgetting.} 
    \label{fig:deployment-stage}
\end{figure*}

\subsection{Deployment Stage} \label{sec:deployment}


\begin{table*}[htbp]
\centering
\caption{Comparison of Model Deployment Strategies in Wi-Fi Sensing.}
\label{tab:deployment_comparison}
\small
\begin{tabularx}{\textwidth}{X XX X X}
\toprule
\textbf{Approach} 
& \textbf{Pros (Advantages)} 
& \textbf{Cons (Limitations)} 
& \textbf{Generalization Mechanism} 
& \textbf{Best-fit Scenarios} \\
\midrule

Transfer Learning (\cite{wang2024xrf55, lan2024bullydetect, 2-136ding2020device, 2-19chen2021semi, 2-161brinke2019scaling, 2-95fang2020witransfer, 1-25yin2022fewsense, 2-38zheng2023resmon, 2-112bi2024roger, 2-127hou2022sample, 2-36hou2022dasecount})
& Fast deployment by leveraging pre-trained weights. \newline Resource efficient with minimal fine-tuning on edge devices. 
& Requires labeled target samples. \newline Performance degrades under large domain gaps. 
& Adapts shared representations by fine-tuning higher layers to capture environment- or user-specific patterns. 
& Rapid adaptation to a specific new environment, user, or novel sensing task. \\

\midrule
Federated Learning (\cite{2-5hernandez2021wifederated, 2-84geng2023personalized, 1-20li2023towards, 2-9qi2023resource, 2-20zhang2023cloud, 1-5zhang2021privacy, 1-35AdaWifi, 2-83zhang2023variance})
& Preserves data privacy as raw CSI remains local. \newline Benefits from collaborative learning across clients. 
& Communication overhead from frequent model synchronization. \newline Latency in large-scale deployments. 
& Learns globally robust models by aggregating heterogeneous local updates from diverse environments. 
& Privacy-sensitive applications such as home security or multi-user collaborative sensing. \\

\midrule
Continual Learning (\cite{2-4yang2023few, 2-134bai2019widrive, 2-144soltanaghaei2020robust, 2-94zhai2021rise, 2-170fu2024ccscontinuouslearningcustomized, zhang2025carec, 2-42zhang2023csi})
& Incrementally acquires new knowledge with local updates. \newline Communication efficient for long-term deployment. 
& Risk of catastrophic forgetting. \newline Potential inference or update latency on COTS devices. 
& Maintains generalization over time by balancing knowledge retention and adaptation to new data distributions. 
& Long-term deployments where environments, furniture layouts, or user behaviors evolve. \\

\bottomrule
\end{tabularx}
\end{table*}

Sensing models ultimately need to be deployed, and many studies focus on enhancing cross-environment generalization during deployment or enabling models to adapt to environmental changes and meet new task requirements through continual learning after deployment.

\subsubsection{Transfer Learning with Fine-tuning}\label{sec:transfer-learning-with-fine-tuning}

The typical steps of transfer learning involve first training a Wi-Fi sensing model on source domain data, followed by updating the model using target domain samples, as shown in Fig.~\ref{fig:deployment-stage}(a). This update can be applied to the task head, feature extractor, or the entire model. For example, XRF55~\cite{wang2024xrf55}, BullyDedect~\cite{lan2024bullydetect}, and WiLISensing~\cite{2-136ding2020device} train models with source domain data and fine-tune the task head using several samples from the target domain, while freezing other parts of the model, enabling few-shot cross-domain activity recognition. DADA-AD~\cite{2-19chen2021semi} fine-tune the deep layers of the model. Besides, Brinke \& Meratnia~\cite{2-161brinke2019scaling} experimentally compare the performance of fine-tuning the task head versus the entire network, finding that fine-tuning the whole network with 20\% of target domain data yields the best results. In addition, in some cases, the target domain classification task differs from the source domain, such as in the number of action categories. To address this, WiTransfer~\cite{2-95fang2020witransfer} replaces the classification head and fine-tunes the new head using target domain data. FewSense~\cite{1-25yin2022fewsense}, on the other hand, discards the classification head and fine-tunes the feature extractor using a cosine similarity loss. This loss pulls features of the same class closer while pushing features of different classes apart. During inference, it employs distance matching for classification, enabling the recognition of novel classes.

Another type of transfer learning consists of two stages: the meta-training stage and the meta-testing stage. In the meta-training stage, a rich labeled dataset from the source domain is used to train an embedding model or feature extractor. In the meta-testing stage, the embedding model is frozen, and a small number of samples from the target domain are input for feature extraction. These features are then used to fine-tune a task head, allowing the model to adapt to the target domain with limited data. Several typical works employ this framework, such as ResMon~\cite{2-38zheng2023resmon}, which uses BayesCNN as the embedding model for domain-adaptive respiration state monitoring. RoGER~\cite{2-112bi2024roger} leverages CNN as the feature extractor to achieve domain-robust gesture recognition.  Hou et al.~\cite{2-127hou2022sample} and DASECount~\cite{2-36hou2022dasecount} both use CNN networks as feature extractors to achieve cross-domain crowd counting.

\subsubsection{Federated Learning}\label{sec:federated-learning}

FDAS~\cite{1-30gong2024privacy}, WiFederated~\cite{2-5hernandez2021wifederated}, pFedBKD~\cite{2-84geng2023personalized}, CARING~\cite{1-20li2023towards}, and Qi et al.~\cite{2-9qi2023resource} utilize federated learning frameworks to achieve efficient model deployment. As shown in Fig.~\ref{fig:deployment-stage}(b), the core components include: (1) numerous clients with data collected from diverse environments, (2) a central server that trains a global model with data from diverse environments, which is expected to possess cross-environment generalization capabilities, and (3) the distribution of the global model to client devices, where it is fine-tuned using a small amount of local data. By leveraging these federated learning principles, Wi-Fi sensing models can be effectively deployed in unseen client environments. 

Some works adopt frameworks similar to federated learning but use different terminology. For instance, Co-WiSensing~\cite{2-20zhang2023cloud} proposes a cloud-edge collaborative framework. In this approach, a cloud server trains a cloud Wi-Fi sensing model using data from various edge devices. The higher layers of the trained model are pruned, and the lower layers are distributed to the edge devices. At the edge, the pruned model is extended with a task-specific head, which is fine-tuned using the edge user's local data to generate a Wi-Fi sensing model tailored for the edge environment. Zhang et al.~\cite{1-5zhang2021privacy} also employs a central server and adversarial learning to extract shared attributes across different environments. The activity-oriented attributes are then transferred to the local server, enabling cross-environment Wi-Fi human activity recognition. AdaWiFi~\cite{1-35AdaWifi} proposes a collective sensing framework, where a central device collects data using multiple sensors and trains multiple encoders collaboratively, which are then aggregated into a unified model. In the end-user environment, the model is collaboratively tuned with a small amount of labeled data. Zhang et al.~\cite{2-83zhang2023variance} propose a local–global modeling method for indoor localization. The approach involves clustering Wi-Fi fingerprints of a large environment into multiple sub-areas, training local localization models for each sub-area, and then aggregating these local models into a global model. The authors demonstrate that this method achieves superior cross-environment generalization.

While federated learning effectively preserves data privacy by keeping raw CSI data on local devices, it introduces non-negligible communication overhead due to frequent model parameter exchanges between clients and the central server. The resulting communication latency can become a bottleneck, particularly in large-scale deployments with heterogeneous network conditions and limited uplink bandwidth.

\subsubsection{Continual Learning}\label{sec:continuous-learning}

After model deployment, it is desirable to continuously update the model to adapt to new environments or tasks, as shown in Fig.~\ref{fig:deployment-stage}(c). Some works achieve this with online learning~\cite{shalev2012online}, a machine learning paradigm where the model learns and updates incrementally as new data becomes available. For instance, FewCS~\cite{2-4yang2023few} collects a small amount of online data to fine-tune model parameters in new environments, enabling few-shot cross-environment human action recognition. Similarly, WiDrive~\cite{2-134bai2019widrive} utilizes online model adaptation and EM algorithm~\cite{dempster1977maximum} to update the model when in-car driver activities are misclassified, improving recognition for diverse vehicles and drivers. Unlike WiDrive's automatic error detection, the User-in-the-loop approach allows users to report misclassifications or annotate a small number of samples, which are then used to update the model. M-WiFi~\cite{2-144soltanaghaei2020robust} specifically asks users to annotate critical time periods to fine-tune the trained model. To minimize the effort required from users, RISE~\cite{2-94zhai2021rise} combines probability and statistical assessments with anomaly detection to identify samples that are likely to be misclassified, reducing the need for users to report every error.

Class incremental learning and domain incremental learning are also applied to enable Wi-Fi human sensing models to adapt to new activities or environments. For example, CCS~\cite{2-170fu2024ccscontinuouslearningcustomized} uses a data replay-based incremental learning approach~\cite{zhao2020maintaining}, where the model is updated to sense new human activity categories using representative samples from previous stages and samples from the current stage. CAREC~\cite{zhang2025carec} is a class-incremental learning framework for Wi-Fi-based indoor action recognition, effectively mitigating catastrophic forgetting~\cite{kirkpatrick2017overcoming} through dynamic model expansion and compression, and achieves high accuracy with an 80\% reduction in parameters. Similarly, CSI-ARIL~\cite{2-42zhang2023csi} treats the environment as a domain, updating the model with representative samples from both old and current environments. CCS, CAREC, and CSI-ARIL ensure that the model does not suffer from catastrophic forgetting when adapting to new classes or domains, preserving its ability to sense old tasks or domains.

Continual learning primarily relies on local model updates and thus incurs minimal communication overhead during operation. However, frequent online adaptation may introduce additional inference or update latency on resource-constrained devices, highlighting the trade-off between real-time responsiveness and long-term adaptability.

Table~\ref{tab:deployment_comparison} compares the advantages and limitations of representative model deployment strategies in Wi-Fi sensing, explains their underlying generalization mechanisms, and summarizes the scenarios in which each approach is best suited. 
The analysis of the model deployment stage highlights several strategic insights, as highlighted in the Key Takeaways and Lessons learned below, for developing generalizable Wi-Fi sensing systems.

\begin{tcolorbox}[
    colback=hotPink!5!white,
    colframe=hotPink!75!black,
    title=\textbf{Key Takeaways and Lessons Learned: Model Deployment},
    width=0.48\textwidth,
    boxrule=0.8pt,
    arc=2mm,
    left=2mm,
    right=2mm,
    top=1mm,
    bottom=1mm,
    breakable
]
\begin{itemize}
    \item \textbf{Hierarchical Efficiency in Transfer Learning:} While fine-tuning the entire network yields superior accuracy, freezing the backbone and updating only the task head is more viable for resource-constrained edge devices. This strategy balances global representation with site-specific adaptation.
    \item \textbf{Privacy-Preserving Collaborative Generalization:} Federated Learning is a powerful engine for building robust global models without moving raw CSI data. A major lesson is managing communication overhead during parameter exchange and addressing the performance drops caused by shifted data distributions across heterogeneous clients.
   \item \textbf{Evolution through Continual Learning:} To handle environmental drifts (e.g., furniture moving), models must incrementally acquire knowledge. The key challenge is mitigating catastrophic forgetting; strategies like replay and model expansion help preserve performance on original tasks.
    \item \textbf{Minimizing User Burden in Online Adaptation:} Incorporating ``User-in-the-loop" mechanisms allows for continuous refinement. The key takeaway is using anomaly detection and high-confidence sampling to minimize the labeling effort required from users during the adaptation phase.
    \item \textbf{Systems-level Latency and Resource Trade-offs:} Beyond accuracy, real-world deployment on COTS chipsets requires balancing model update frequency with inference latency. For high-responsiveness applications, local updates must be optimized to prevent system bottlenecks.
\end{itemize}
\end{tcolorbox}

To further explain why approaches in the four stages of the Wi-Fi sensing pipeline can enhance generalizability, we analyze them through the lens of \textit{Domain Adaptation Theory}~\cite{ben2010theory}. The generalization error on a target domain $\epsilon_T(h)$ is bounded by the source error $\epsilon_S(h)$, the domain divergence $d_{\mathcal{H}\Delta\mathcal{H}}(P_S, P_T)$, and the complexity of the ideal joint hypothesis $\lambda$: $$\epsilon_T(h) \leq \epsilon_S(h) + \frac{1}{2} d_{\mathcal{H}\Delta\mathcal{H}}(P_S, P_T) + \lambda$$ 

\begin{tcolorbox}[
    colback=blue!5!white,
    colframe=blue!75!black,
    title=\textbf{Theoretical Analysis into Generalizability},
    width=0.48\textwidth,
    boxrule=0.8pt,
    arc=2mm,
    left=2mm,
    right=2mm,
    top=1mm,
    bottom=1mm,
    breakable
]

\begin{itemize}
    \item \textbf{Experimental Setup (Distribution Approximation):} By scaling up datasets and utilizing distributed hardware, these approaches ensure the empirical training distribution provides a denser coverage of the physical signal space. This not only minimizes $\epsilon_S(h)$ but also constrains the inherent bias $\lambda$ between source and target environments.
    
    \item \textbf{Signal Preprocessing (Nuisance Variable Elimination):} Physics-based indicators (e.g., BVP, DFS) act as \textit{canonical coordinate transformations} that project raw CSI into a domain-invariant physical space. By decoupling activity features from environmental \textit{nuisance variables}, these methods significantly reduce the domain divergence $d_{\mathcal{H}\Delta\mathcal{H}}$ at the input level.
    
    \item \textbf{Feature Learning (Information Bottleneck):} Techniques like domain alignment and disentanglement operate on the \textit{Information Bottleneck principle}~\cite{tishby2015deep}. They maximize the mutual information $I(Z; Y)$ for task labels while minimizing $I(Z; D)$ for domain-specific patterns, thereby aligning feature manifolds in a latent subspace.
    
    \item \textbf{Model Deployment (Bayesian Adaptation):} Strategies such as few-shot fine-tuning represent a \textit{Bayesian posterior update}. They leverage a robust prior $P(\theta)$ learned across diverse source domains to rapidly converge to a site-specific optimum $P(\theta|D_{target})$ with minimal target data.
\end{itemize}

\end{tcolorbox}

\begin{tcolorbox}[
    colback=green!5!white,     
    colframe=green!45!black,   
    title=\textbf{Highlight},
    width=0.48\textwidth,
    boxrule=0.8pt,
    arc=2mm,
    left=2mm,
    right=2mm,
    top=1mm,
    bottom=1mm,
    breakable
] 

While this survey focuses on Wi-Fi sensing, many of the taxonomy and generalization challenges discussed also apply to millimeter-wave radar–based human sensing~\cite{zhang2023survey}, which has gained attention with the rise of commodity radars. Although mmWave radar uses different signal representations, such as range–Doppler and range–angle maps instead of Wi-Fi CSI, it faces similar challenges involving device heterogeneity, human body diversity, and environmental variability. Variations in radar hardware and antenna arrays parallel device heterogeneity in Wi-Fi; differences in body shape, motion, and orientation affect radar reflections similarly to CSI-based sensing; and environmental factors like clutter, multipath, and deployment geometry likewise influence robustness. Therefore, generalization-oriented strategies, e.g., multi-view sensing, large-scale in-the-wild datasets, domain adaptation, and continual learning, can be extended to mmWave radar sensing despite differences in signal modality.

\end{tcolorbox}

\section{Dataset}\label{sec:dataset}

Recent years have witnessed a clear shift in Wi-Fi sensing research from lab-scale prototypes toward tools and datasets that support modern wireless standards and more complex, real-world sensing configurations. Early studies often relied on limited-bandwidth devices and small-scale, controlled environments, which constrained the evaluation of generalization performance. In contrast, emerging tools and benchmarks increasingly embrace higher bandwidths, denser subcarrier representations, and diverse deployment scenarios, enabling more realistic and robust system-level research.

Several open-source tools are available for extracting Wi-Fi Channel State Information and Beamforming Feedback Information (BFI), as summarized in Table~\ref{tab:wifi-tools}. Among these, Linux CSI Tool~\cite{halperin2011tool} was the earliest open-source CSI extraction tool released in 2011, while PicoScenes~\cite{jiang2021eliminating,li2024reshaping} represents a new generation of sensing tools that align with modern wireless standards. Specifically, PicoScenes supports advanced IEEE 802.11 standards up to Wi-Fi~7, offering bandwidths up to 320 MHz and up to 1024 subcarriers, which significantly expands the sensing resolution in both time and frequency domains and enables sensing under more complex multipath and heterogeneous deployment conditions. In addition, due to the limited availability of networks and devices supporting open CSI extraction, BFM-Tool~\cite{yi2024bfmsense} facilitates BFI extraction, enabling Wi-Fi sensing across a broader range of commercial devices.

\begin{table}[!htbp]
\centering
\caption{Wi-Fi Channel State Information and Beamforming Feedback Information extraction tools.}
\setlength{\tabcolsep}{1pt}
\begin{tabular}{lccccc}
\toprule
\textbf{Name} & \makecell{ \textbf{max.} \\\textbf{MIMO}} & \makecell{\textbf{802.11} \\ \textbf{support}} & \makecell{\textbf{max.} \\ \textbf{\# subcarriers}} & \makecell{\textbf{max.} \\ \textbf{bandwidth}} & \textbf{Wi-Fi}\\
\midrule
Intel CSITool~\cite{halperin2011tool} & 2$\times$2 & n & 60 & 40 MHz & CSI \\
Atheros CSITool~\cite{xie2015precise} & - & n & 114 & 40 MHz & CSI \\
Nexmon CSITool~\cite{gringoli2019free} & 4$\times$4 & n/ac & 256 & 80 MHz & CSI \\
Wi-ESP~\cite{atif2020wi} & 2$\times$2 & b/g/n/ac & 64 & 40 MHz & CSI \\
ZTE CSITool~\cite{wang2025wi} & 3$\times$2 & n/ac/ax & 512 & 160 MHz & CSI \\
PicoScenes~\cite{jiang2021eliminating,li2024reshaping} & 4$\times$4 & a/g/n/ac/ax/be & 1024 & 320 MHz& CSI \\
BFM-Tool~\cite{yi2024bfmsense} & 4$\times$4 & ac/ax & 512 & 160 MHz & BFI \\
\bottomrule
\end{tabular}
\label{tab:wifi-tools}
\end{table}

\begin{figure*}[t]
    \centering
    \includegraphics[width=1\linewidth]{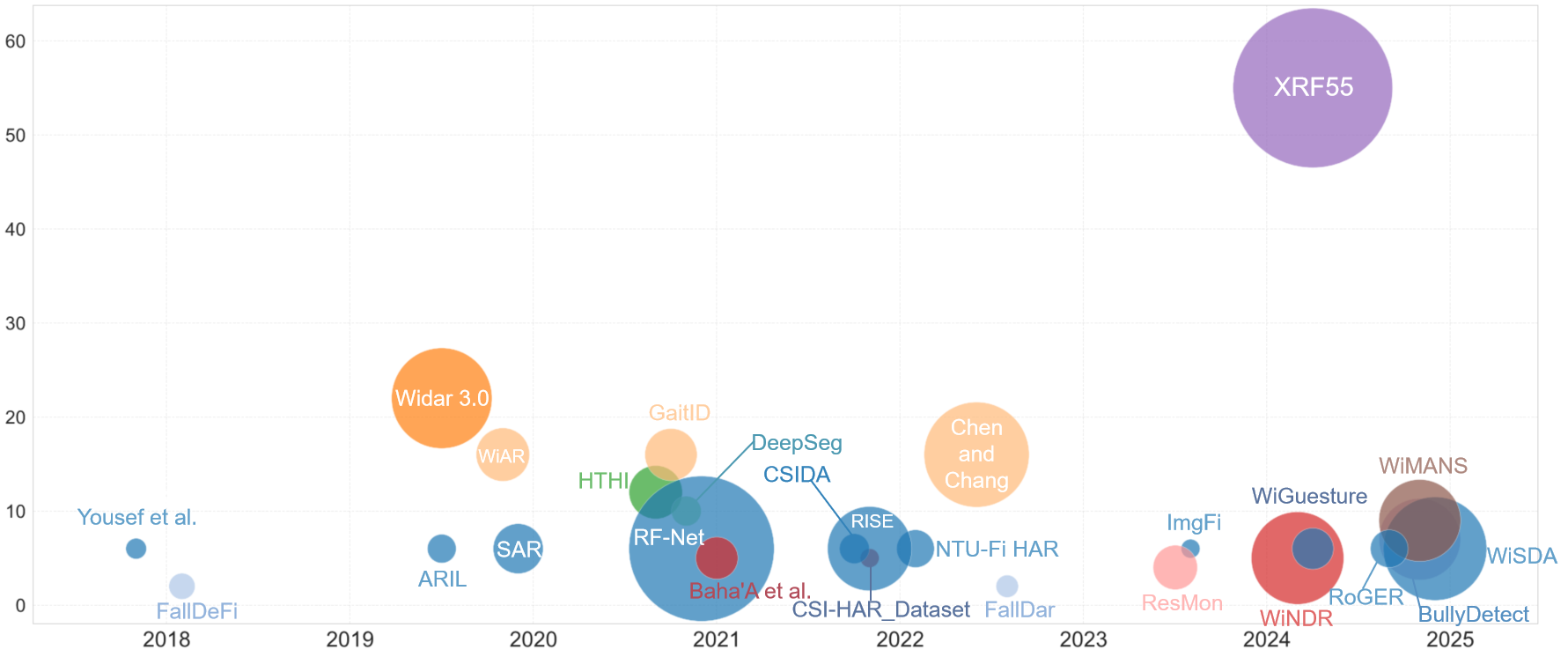}
    \caption{Bubble chart of public Wi-Fi sensing datasets, where the bubble size indicates the number of samples and the vertical axis represents the number of classes. For visual clarity, datasets with extreme values, such as SignFi~\cite{ma2018signfi} (296 classes) and CSI-Bench~\cite{zhu2025csi} (over 1,000,000 samples), are excluded from the plot. Besides, datasets like WiFiTAD~\cite{liu2025wifi}, MM-Fi~\cite{yang2024mm}, and XRFv2~\cite{lan2025xrf}, which collect continuous sequences over time without clear per-sample annotations, are also not included due to the ambiguity in estimating their sample counts.}
    \label{fig:dataset-over-the-years}
\end{figure*}

Beyond advances in data acquisition tools, recent Wi-Fi sensing datasets have also shifted from small-scale, lab-controlled collections toward large-scale, in-the-wild benchmarks. Over the years, many researchers have generously released public datasets, significantly improving the efficiency of algorithm evaluation by removing the need to collect data from scratch. Among the various sensing tasks, activity recognition is currently the most well-represented in public datasets, as summarized in Table~\ref{tab:dataset}. To provide a more intuitive overview of activity recognition datasets, Fig.~\ref{fig:dataset-over-the-years} presents a bubble chart that visualizes the number of activity classes, release date, and dataset volume. 

In addition, datasets are also available for tasks such as gait recognition, pose estimation, indoor localization, and temporal action localization.
These datasets often include multiple domain attributes, such as environment, participant identity, and action category, making them especially suitable for evaluating generalization and cross-domain performance. In this section, we briefly introduce the domain properties of each dataset to help researchers efficiently identify the most appropriate datasets for evaluating their methods.

Furthermore, to address the scarcity of data, promote industry-academia collaboration, and foster a thriving sensing ecosystem, we have initiated and built the Sensing Dataset Platform (SDP:~\textcolor{mypink}{\url{https://www.sdp8.org/}}). As the most comprehensive open repository for wireless sensing datasets to date, SDP hosts a wide range of publicly available datasets contributed by both industrial and academic communities. The platform also encourages researchers to share their own data, aiming to foster open collaboration, accelerate innovation, and support reproducible research in the field of wireless sensing.

\begin{table*}[!ht]
    \centering
    \scriptsize
    \setlength{\tabcolsep}{1pt}
    \caption{Wi-Fi Datasets. The main results column mainly denotes the in-domain results from the original publications.} 
    \begin{tabularx}{{\textwidth}}{llp{5.3cm}lll}
    \toprule
    \textbf{ID} & \textbf{Dataset} & \textbf{Main Domain Information} & \textbf{Device Setting}& \textbf{Highlight} & {Main Results} \\
    \midrule
    \multicolumn{6}{c}{\textbf{Action Recognition}} \\ 
    1 & XRF55~\cite{wang2024xrf55} & 4 environments, 39 participants, 55 actions & 1Tx, 3Rx, 20MHz, 5GHz & Multimodal, 55 actions  & 87.26\%\\
    2 & BullyDetect~\cite{lan2024bullydetect} & 8 environments, 20 paired participants, 7 actions & 1Tx, 1Rx, 20MHz, 5GHz & Bullying actions & 90.4\% \\
    3 & WiSDA~\cite{2-1jiao2024wisda} & 1 environments, 3 participants, 6 actions & 1Tx, 1Rx, 20MHz & Widar3.0-like & 99.5\%\\
    4 & WiNDR~\cite{2-10qin2024direction} & 1 environments, 3 participants, 5 actions, 24 orientation & 1Tx, 1Rx, 20MHz, 5GHz & Full 360° coverage & 78\%-92\%\\
    5 & WiMANS~\cite{huang2024wimans} & 3 environments, 6 participants, 9 actions & 1Tx, 1Rx, 20MHz, 2.4/5GHz& Multi-person& 89.1\%-96.6\%\\
    6 & WiGuesture~\cite{zhao2024finding} & 1 environments, 8 participants, 6 actions & 1Tx, 1Rx, 100Hz, 2.4GHz & ESP32S3 & 92.57\% \\
    7 & RoGER~\cite{2-112bi2024roger} & 2 environments, 4 participants, 6 actions, 4 orientation & 1Tx, 2Rx, 20MHz, 2.437GHz & AR9580& 94.84\%-99.71\%\\
    8 & ImgFi~\cite{zhang2023imgfi} & 1 environments, 5 participants, 6 actions & 1Tx, 1Rx, 20MHz & Converted to images& 99.8\% \\
    9 & ResMon~\cite{2-38zheng2023resmon} & 2 environments, 3 areas, 6 participants, 4 actions & 1Tx, 1Rx, 20/40MHz, 2.4/5GHz & Cross-band test& 83.80\%-89.67\%\\
    10 & Meneghello et al.~\cite{meneghello2023csi} & 6 environments, 4 participants, 7 actions & 1Tx, 1Rx, 80MHz, 5.21GHz & 80MHz& not reported \\
    11 & Demrozi et al.~\cite{demrozi2023dataset} & 2 environments, 6 participants, 6 actions & 2Tx, 2Rx, 2.4GHz & Nexmon firmware& 99.3\%\\
    12 & MM-Fi~\cite{yang2024mm} & 4 environments, 40 participants, 27actions & 1TX, 1RX, 20MHz, 5GHz& Multimodal& not reported \\
    13 & NTU-Fi HAR~\cite{yang2022efficientfi} & 1 environments, 20 participants, 6 actions & 1Tx, 1Rx, 40MHz, 5GHz & TP-Link N750 & 98.6\% \\
    14 & FallDar~\cite{yang2022rethinking} & 4 environments, 6 locations, 6 participants, fall \& normal & 1Tx, 1Rx, 1000Hz, 5GHz & 6 months&   5.7\% FAR \\
    15 & ReWis~\cite{2-26bahadori2022rewis} & 3 environments, 2 participants, 4 actions & 1Tx, 3Rx, 20/80MHz, 5GHz & 80MHz& 78.25\%-99.82\% \\
    16 & OPERAnet~\cite{bocus2022operanet} & 2 environments, 6 participants, 6 actions & 1Tx, 2Rx,  5GHz & Multimodal& 71\%-100\%\\
    17 & SHARP~\cite{2-163meneghello2022sharp} & 3 environments, 4 locations, 3 participants, 7 actions & 1Tx, 1Rx, 80MB, 5GHz & 80MHz& 95.99\%-99.79\%\\
    18 & CSI-HAR-Dataset~\cite{moshiri2021csi} & 1 environments, 3 participants, 7 actions & 1Tx, 1Rx, 20MHz, 5GHz & 802.11ac& 95.5\% \\
    19 & CSIDA~\cite{1-6zhang2021wifi} & 2 environments, 5 locations, 5 participants, 6 actions & 1Tx, 1Rx, 40MHz, 5GHz & 114 subcarriers  & 90.10\%$\pm$1.03\%\\
    20 & RISE~\cite{2-94zhai2021rise} & 2 environments, 6 participants, 6 actions & 1Tx, 1Rx & multi-rf devices&  93.9\%-100\%\\
    21 & HTHI~\cite{alazrai2020dataset} & 1 environment, 40 paired, 12 actions & 2Tx, 3Rx, 20MHz, 2.4GHz & Human-human actions& note reported \\
    22 & DeepSeg~\cite{xiao2020deepseg} & 1 environment, 5 participants, 10 actions & 1Tx, 3Rx &  fine-/coarse-grained actions& 94\% \\
    23 & RF-Net~\cite{1-2ding2020rf} & 6 environments, 11 participants, 6 actions & 2 Tx-Rx pairs, 20MHz & 120 environment & $\approx$ 80\%\\
    24 & Baha’A et al.~\cite{baha2020dataset} & 3 environments, 30 participants, 12 actions & 1Tx, 3Rx, 20MHz, 2.4GHz & LOS/NLOS& not reported\\
    25 & Widar 3.0~\cite{1-45zheng2019zero} & 3 environments, 5 locations, 5 orientations, 16 participants, 16 actions & 1Tx, 6Rx, 20MHz, 5.825GHz & Rich domains& 92.7\%-92.9\%\\
    26 & ARIL~\cite{wang2019joint} & 16 locations, 1 participants, 6 actions & 1Tx, 1Rx, 20MHz, 2.4GHz & USRP data, clean phase& 89.57\% \\
    27 & WiAR~\cite{guo2019wiar} & 3 environments, 10 participants, 16 actions & 1Tx, 1Rx, 20MHz, 5GHz & RSSI and CSI & 90.62\%-96.25\%\\
    28 & SAR~\cite{brinke2019dataset,2-161brinke2019scaling} & 1 environment, 9 participants, 6actions & 2-3Tx, 3Rx, 20MHz, 2.4GHz & 6 days& 60\%-100\%\\
    29 & FallDeFi~\cite{palipana2018falldefi} & 5 environments, 3 participants, 9 actions & 2Tx, 2Rx, 20MHz, 5.2GHz & Fall detection & 80.10\%-88.90\% \\
    30 & SignFi~\cite{ma2018signfi} & 2 environments, 5 participants, 276 gestures & 3Tx, 1Rx, 20MHz, 5GHz& 276 ASL Gestures& 94.81\%-98.91\% \\
    31 & CrossSense~\cite{1-50zhang2018crosssense} & 3 environments, 15 locations, 100 participants, 40 actions & 1Tx, 1Rx, 20MHz, 5GHz & 100 participants & 90\%\\
    32 & Yousef et al.~\cite{yousefi2017survey} & 1 environments, 6 participants, 6 actions & 1Tx, 1Rx, 20MHz, 5GHz & Fall& 81\%-97\% \\ 

33 & OctoNet~\cite{yuanoctonet2025} & 3 environments, 41 participants, 62 actions & 1Tx, 4Rx, 40MHz, 5.18GHz & Multimodal, 62 actions& 93.3\%$\pm$2.3\% \\ 

    \midrule
    \multicolumn{6}{c}{\textbf{Gait Recognition}} \\  
    34 & NTU-Fi HumanID~\cite{2-130wang2022caution} & 2 environments, 20 participants, walking & 1Tx, 1Rx, 40MHz, 5GHz & TP-Link N750 & 86.29\&-98.34\% \\ 
    20 & RISE~\cite{2-94zhai2021rise} & 3 environments, 15 participants & 1Tx, 1Rx &  multi-rf devices& 97.6\%\\
    35 & GaitID~\cite{2-157zhang2020gaitid} & 2 environments, 11 participants, 8 directions & 1Tx, 6Rx& 8 directions& 61\%-100\% \\
    31 & CrossSense~\cite{1-50zhang2018crosssense} & 3 environments, 100 participants & 1Tx, 1Rx, 20MHz, 5GHz & 100 participants & 80\% \\

    \midrule
    \multicolumn{6}{c}{\textbf{Pose Estimation}} \\ 
    33 & OctoNet~\cite{yuanoctonet2025} & 3 environments, 41 participants, 62 actions & 1Tx, 4Rx, 40MHz, 5.18GHz & Multimodal & 147.3$\pm$4.7mm MPJPE \\ 
    36 & XRFv2~\cite{lan2025xrf} & 3 environments, 16 participants, 30 actions & 1Tx, 3Rx, 20MHz, 5GHz & Multimodal, continuous actions& not reported \\
    37 & Person-in-WiFi 3D~\cite{yan2024person} & 3 environments, 7 participants, 8actions &1Tx, 3Rx, 20MHz, 5GHz & Multi-person&  107.2mm MPJPE\\
    1 & XRF55~\cite{wang2024xrf55} & 4 environments, 39 participants, 55 actions & 1Tx, 3Rx, 20MHz, 5GHz & Multimodal, 55 actions & not reported\\
    12 & MM-Fi~\cite{yang2024mm} & 4 environments, 40 participants, 27 actions & 20MHz, 5GHz& Multimodal& 197.1mm MPJPE \\
   
    \midrule
    \multicolumn{6}{c}{\textbf{Indoor Localization}} \\ 
    38 & MetaLoc~\cite{2-125gao2023metaloc} & 2 environments, 180 locations & 3Tx, 1Rx, 20MHz, 5GHz &  RSSI and CSI & 2.07$\pm$1.11m error \\
    39 & Chen and Chang~\cite{2-121chen2022few} & 2 environments, 16 locations & 2Tx, 2Rx & CSI fingerprints & 49.47\%-90.44\%\\
    26 & ARIL~\cite{wang2019joint} & 1 environments, 16 locations, 1 participants, 6 actions & 1Tx, 1Rx, 20MHz, 2.4GHz & USRP data, clean phase& 95.68\% \\

    \midrule
    \multicolumn{6}{c}{\textbf{Crowd Counting}}  \\ 
    40 & SDP~\cite{tony-sdp-2024} &  4 scenarios & multi-APs, 80/160MHz, 5GHz& 0-2 people& not reported \\
    41 & DASECount~\cite{2-36hou2022dasecount} &  2 environments, 2 scenarios (NLOS/LOS), 3 motions & 1Tx, 1Rx, 40MHz, 2.4GHz& 0-8 people& 93.29\%-99.17\%\\
    10 & Meneghello et al.~\cite{meneghello2023csi} & 1 environments, 10 participants & 1Tx, 1Rx, 80MHz, 5.21GHz & 1-10 people& not reported \\
    \midrule
    \multicolumn{5}{c}{\textbf{Temporal Action Localization, Action Summarization}} & \\ 
    36 & XRFv2~\cite{lan2025xrf} & 3 environments, 16 participants, 30 actions & 1Tx, 3Rx, 20MHz, 5GHz & Multimodal, continuous actions& 78.74 mAP\\
    41 & WiFiTAD~\cite{liu2025wifi} & 1 environment, 3 participants, 7 actions & 1Tx, 1Rx,  5GHz& Continuous actions & 74.5 mAP\\

     \midrule
    \multicolumn{6}{c}{\textbf{Multiple Sub-Datasets}} \\ 
    43 & CSI-Bench~\cite{zhu2025csi} & 26 environments, 35 participants & 16 device configuration & In-the-wild data, 461 hours& 94.88\% fall detection, etc.\\
    10 & Meneghello et al.~\cite{meneghello2023csi} & 7 environments, 13 participants & 1Tx, 1Rx, 80MHz, 5.21GHz & Three tasks, 80MHz & not reported \\
    44 & CSI-Net~\cite{wang2018csi} & 1 environments, 5 positions, 30 participants & 1Tx, 1Rx, 20MHz, 5GHz & Four tasks, Biometrics& 96.67\% fall detection, etc. \\
    \bottomrule
    \end{tabularx}
    \label{tab:dataset}
\end{table*}

     \textbf{(1) XRF55} \cite{wang2024xrf55} dataset was collected in four indoor environments. In each environment, users performed 55 types of actions, categorized into five categories: human-object interactions, human-human interactions, fitness, body motions, and human-computer interactions, within a 3.1m $\times$ 3.1m area. The first scene involved 30 participants, while the other three scenes involved 3 participants each. Each participant performed each action 20 times, with 42,900 samples in total, lasting 59.58 hours. The data collection setup included one transmitter with a single antenna working, and three receivers each with three antennas, arranged in a quadrilateral layout. The sampling rate was 200 Hz. Notably, XRF55 dataset also includes synchronized modalities from millimeter-wave radar, RFID, and RGB+D+IR, supporting multimodal learning and cross-modal tasks. Additionally, XRF55 also releases 2D human pose annotations for the pose estimation task.

     \textbf{(2) BullyDetect} \cite{lan2024bullydetect} dataset was collected in 8 different environments: a classroom, corridors, playgrounds, rooftops, a meeting room, a corner of the campus, an Office, and an abandoned building. Five volunteers were paired in 10 pairs, each acting as either an attacker or a victim during the experiment. The participants performed 7 different bullying actions: pushing, kicking, slapping, grabbing, punching, kneeing, and hitting with a stick. Each action was performed 5 seconds and 10 times, resulting in 11,200 samples. The experimental setup included a transmitter with a single antenna and one 3-antenna receiver, spaced 3 meters apart, with a sampling rate of 1000 Hz. 

      \textbf{(3) WiSDA}~\cite{2-1jiao2024wisda} dataset was collected in an office setting, equipped with a desk and chair, within a 2m$\times$2m sensing area. Three participants performed six activity categories: Push, Sweep, Clap, Slide, Draw-Z, and Draw-N. Each person collected 100 samples on 30 subcarriers and converted them to images, resulting in 3000 samples per action. The data was captured using a single pair of antennas, with one acting as the transmitter and the other as the receiver within the sensing area, similar to the setup used in the Widar 3.0 dataset. The bandwidth is 20 MHz.

       \textbf{(4) WiNDR}~\cite{2-10qin2024direction} dataset was collected in an indoor office room measuring 4.8m$\times$3.1m. Three volunteers incluing two males and one female, aged between 25 and 30 years old and weighing between 50 to 65 kg, were asked to perform five distinct hand gestures: drawing a circle (start/accept), crossing hands (stop), clapping (switch context/menu), raising hands (increase/turn up), and lowering hands (decrease/turn down). The volunteers were instructed to perform these gestures in 24 different directions, marked by gray lines on the floor, with each direction spaced 15 degrees apart. These gestures were performed 40 times in each of the 24 directions, sampled every 15 degrees around the subjects. The collection of gesture data was facilitated by two computers, the antennas were extended with 3.5m cables and positioned 1.2m above the ground using tripods.The sampling rate was 200Hz.

     \textbf{(5) WiMANS}~\cite{huang2024wimans} dataset was collected in three environments including a classroom, a meeting room and an empty room. Each environment was equipped with a transmitter, a receiver and a monitor camera used to capture synchronized videos. The activities were performed by five volunteers including an adult, a middle-aged person, and an elderly person. Nine activities were included in the dataset: nothing, walking, rotation, jumping, waving, lying down, picking up, sitting down and standing up. Each environment has five different locations. The dataset provides 11286 samples (over 9.4 hours) of dual-band WiFi CSI and synchronized videos.

     \textbf{(6) WiGuesture}~\cite{zhao2024finding} dataset was collected in a conference room. Eight volunteers participated in the study. The participants performed six distinct gestures: moving left-right, forward-backward, up-down, circling clockwise, clapping, and waving. Each gesture was recorded continuously for one minute. The data collection setup consisted of a six-antenna home Wi-Fi router acting as the transmitter and a single-antenna ESP32S3 microcontroller serving as the receiver. The transmitter and receiver were placed 1.5 meters apart, and data was captured at a sampling rate of 100 Hz.  

    \textbf{(7) RoGER}~\cite{2-112bi2024roger}  dataset was collected in a meeting room at two distinct locations. Location 1: A single volunteer performed six gestures, i.e., Push \& Pull, Draw Zigzag, Clap, Sweep, Draw Circle, Slide, the same as Widar 3.0~\cite{1-45zheng2019zero}, in four different directions. Each gesture was repeated 30 times, resulting in 480 samples. Location 2: Four volunteers (including the participant from Location 1) performed the same six gestures in four directions. Each volunteer repeated each gesture 30 times, generating 1,920 samples. Each sample lasts for 10 seconds. The setup included a two-antenna transmitter and a three-antenna receiver, operating at 2.437 GHz with a sampling rate of 100 Hz.

     \textbf{(8) ImgFi}~\cite{zhang2023imgfi} dataset was collected in a laboratory, where 5 volunteers performed 6 actions: bending, drinking, nodding, squatting, drawing the letter O, and drawing the letter B. Each action was repeated 20 times. Data collection was conducted using one transmitter and one receiver positioned 4 meters apart, with the volunteers performing actions in the center. The sampling rate was 100Hz. 

    \textbf{(9) ResMon}~\cite{2-38zheng2023resmon} dataset was collected in two environments: a laboratory and a meeting room, with the laboratory further collected at two areas. Six volunteers performed 4 actions: stable breath, cough, sneeze, and yawn. The experimental setup included a two-antenna transmitter and a three-antenna receiver, operating at 2.437 GHz with a 20MHz bandwidth and sampling rate of 100 Hz. ResMon also includes samples collected at 5 GHz with a 40 MHz bandwidth for cross-band evaluation.

\textbf{(10) Meneghello et al.}~\cite{meneghello2023csi}  present a Wi-Fi sensing dataset collected using the 802.11ac protocol with an 80 MHz bandwidth centered at 5,210 MHz. It supports action recognition (AR), person identification (PI), and people counting (PC) tasks. For AR, four participants performed up to seven activities (e.g., walking, running, jumping, sitting still, standing still, sitting down/standing up, arm exercises) in six environments: including a bedroom, living room, kitchen, university laboratory, university office, and semi-anechoic chamber. For PI, 10 individuals were recorded moving freely in a meeting room (7 × 7.5 × 3.5 m). For PC, 1 to 10 people moved simultaneously in the meeting room. The dataset spans over 13 hours of Channel State Information recordings, collected using the Nexmon tool with Netgear X4S AC2600, Asus RT-AC86U, and TP-Link AD7200 routers.

\textbf{(11) Demrozi et al.}~\cite{demrozi2023dataset}~dataset was collected in two office environments. The first office, measuring 12m$\times$6m$\times$3m, is divided into two blocks of workspaces, each equipped with a Raspberry Pi device and an access point (AP). The second office is smaller, 6m$\times$4m$\times$2.75m, and also contains two Raspberry Pi devices and two APs. The Raspberry Pis and APs were paired to collect Channel State Information. Six participants, including two females and four males, performed five types of activities: entering the office, walking, standing, sitting, and leaving the office. The devices were arranged such that two wireless routers were placed 5 meters apart at a height of 140 cm, creating two separate 2.4 GHz WiFi networks. The total size of the dataset is approximately 70 GB.

\textbf{(12) MM-Fi}~\cite{yang2024mm}  dataset comprises data collected in two rooms, 4 environmental settings, with 10 participants in each environment. The participants performed 14 daily activities, such as raising the left arm, waving the left arm, and waving the right arm, as well as 13 rehabilitation exercises, such as left front lunge and right front lunge. The setup included a transmitter with one antenna and a receiver with three antennas, arranged with a spacing of 3.75 meters between the receiver antennas. Participants performed actions while positioned 3 meters away from the receiver, facing the receiver. Each action was performed for 30 seconds. The devices operated at 5 GHz with a sampling rate of 1000 Hz. In addition to Wi-Fi sensing, MM-Fi includes data from LiDAR, mmWave radar, and camera modalities, supporting multimodal learning tasks.   

 \textbf{(13) NTU-Fi HAR}~\cite{yang2022efficientfi}  was collected in a laboratory setting with 20 participants performing six types of actions, including running, walking, falling down, boxing, circling arms, and cleaning the floor. 
    Each participant performed each action 20 times, resulting in a total of 2,400 samples. The setup consisted of a transmitter with one antenna and a receiver with three antennas. The equipment operated at 5 GHz with a sampling rate of 500 Hz.

\textbf{(14) FallDar}~\cite{yang2022rethinking} dataset, developed for fall detection, was collected in two environments: a home environment and an office environment. In the home environment, five participants performed fall actions at three locations—dining room, living room, and balcony—resulting in 115 fall samples. The transmitter and receiver were placed in two settings: line-of-sight and non-line-of-sight. In the office environment, the transmitter was fixed in position, while the receiver was placed at 16 locations sequentially. Two participants contributed to the dataset. One performed fall actions at locations 1 and 3, while the other performed falls at locations 1 and 4. Additionally, one participant walked freely in the office to collect normal activity data. This resulted in 228 fall samples and 504 normal activity samples. In both environments, the transmitter had one antenna, and the receiver had three antennas. Data was collected at a sampling rate of 1,000 Hz with a center frequency of 5.825 GHz. 

\textbf{(15) ReWiS}~\cite{2-26bahadori2022rewis} dataset was collected in three environments: an office, a meeting room, and a classroom. Two participants performed four actions---empty room, jumping, walking, and standing---10 times in each environment, with each session lasting 180 seconds and a minimum interval of 2 hours between sessions. A Netgear R7800 Wi-Fi router with a Qualcomm Atheros chipset operated in AP mode, while an off-the-shelf laptop functioned as the client. Additionally, three Asus RT-AC86U Wi-Fi routers, each equipped with four antennas, recorded Wi-Fi CSI at a rate of 100 Hz over 20 MHz (52 subcarriers) and 80 MHz (242 subcarriers) bandwidths at 5 GHz.

 \textbf{(16) OPERAnet}~\cite{bocus2022operanet} dataset was collected in two rooms and includes data for seven activities: walk, sit, stand from a chair, lie down, stand up from the floor, rotate the upper-half body, and steady state, totaling 417 minutes of activity data. Additionally, the dataset includes 18 minutes of background data and 27 minutes of crowd-counting data. The data collection setup consisted of one transmitter with three antennas and two receivers, each with three antennas, arranged in a right-angled triangular configuration. The second receiver was positioned at the right-angle vertex of the triangle. The sampling rate for Wi-Fi signals was 1600 Hz. OPERAnet dataset also includes synchronized data from two Kinect cameras, two UWB systems (4 nodes in system 1, 5 nodes in system 2), and a Passive WiFi Radar system constructed using a USRP-2945. This multimodal setup supports research in both multi-modal learning and cross-modal tasks. 

 \textbf{(17) SHARP}~\cite{2-163meneghello2022sharp}  dataset was collected in three environments: a bedroom, a living room, and a laboratory, featuring four activities—sit, walk, run, and jump—along with an empty room state. In the bedroom, volunteers P1 and P2 performed at two locations; in the living room, volunteer p1 performed at one location; and in the laboratory, volunteer P3 performed. These activities were conducted at different times over ten months. Each activity lasted 120 seconds per session, resulting in a total of 120 minutes of data. The setup included a single-antenna transmitter and receiver, along with a 4-antenna Asus RT-AC86U router as a monitoring device for collecting Wi-Fi data at a sampling rate of 173 Hz.

    \textbf{(18) CSI-HAR-Dataset}~\cite{moshiri2021csi} dataset was collected in a single indoor environment. The dataset was collected by 3 volunteers of different ages (adult, middle-aged, and elderly). Each volunteer performed seven categories of actions: walking, running, falling, lying down, sitting down, standing up, and bending. Each action was repeated 20 times by each participant, resulting in 420 samples. The experimental setup involved a Raspberry Pi 4 as the receiver and a Tp-link Archer C20 wireless router as the transmitter, operating on a 5 GHz WiFi network with a 20 MHz bandwidth on channel 36 (IEEE 802.11ac standard). The transmitter and receiver were positioned 3 meters apart, both 1 meter above the ground to ensure an unobstructed signal path. The sampling rate was 200 Hz, with 4000 samples collected over 20 seconds per activity, where the activity occurred in the middle of this period (approximately 3–6 seconds).

     \textbf{(19) CSIDA},  related in WiGr~\cite{1-6zhang2021wifi},  was collected in an office environment with three distinct locations and a classroom environment with two locations. Five volunteers performed six gestures: pull left, pull right, lift up, press down, draw a circle, and draw zigzag. Each gesture was repeated 10 times, with each repetition lasting 1.8 seconds. The device setup consisted of a single-antenna transmitter and a three-antenna receiver, separated by a distance of 2.6 meters. The system operated in the 5 GHz mode with a 40 MHz channel bandwidth, capturing 114 subcarriers with a sampling rate of 1000 Hz.

\textbf{(20) RISE}~\cite{2-94zhai2021rise} dataset includes data for gesture recognition and gait recognition, collected in two environments: an office and a controlled setting with a radio frequency anechoic chamber to minimize multipath effects. Gesture Recognition: Six volunteers performed six gestures—push and pull, draw a circle, throw, slide, sweep, and draw zigzag—across ten locations (five per environment). Each gesture was repeated 30 times per location, resulting in 10,800 samples (6 volunteers ×6   gestures ×10 locations ×30 repetitions). Gait Recognition: Fifteen volunteers walked under five different configurations in the controlled environment, generating 1,150 samples across ten activity types.

     \textbf{(21) HTHI}~\cite{alazrai2020dataset} dataset was collected in a furnished room. 66 volunteers participated in the study, forming 40 unique pairs to perform 12 types of human-to-human interaction activities, such as handshaking, high fives, and hugging. Each pair performed each activity 10 times, resulting in 4,800 samples. The data collection setup consisted of a commercial off-the-shelf Sagemcom 2704 router with two antennas acting as the transmitter and a desktop computer with three antennas serving as the receiver. The transmitter and receiver were positioned 4.3 meters apart. The sampling rate for the data collection was not reported.

    \textbf{(22) DeepSeg} ~\cite{xiao2020deepseg} dataset was collected in a meeting room with data from 5 volunteers performing 10 activities. These actions include 5 fine-grained movements such as swinging the hand, raising the hand, making a pushing motion, tracing a circular pattern (O), and tracing a cross pattern (X), as well as 5 coarse-grained activities, namely boxing, picking up an object, running, squatting down, and walking. Each participant repeated each action 30 times, resulting in 1,500 samples. The dataset collection setup included a single-antenna transmitter and a three-antenna receiver placed 2 meters apart, with a sampling rate of 50 Hz.

    \textbf{(23) RF-Net}~\cite{1-2ding2020rf} dataset was collected in six different rooms with 11 volunteers performing six types of actions: wiping, walking, moving, rotating, sitting, and standing up. The Wi-Fi setup included two transmitter-receiver pairs, operating at a sampling rate of 100 Hz.   

dataset was collected in six different rooms with 11 volunteers performing six types of actions: wiping, walking, moving, rotating, sitting, and standing up. The Wi-Fi setup included two transmitter-receiver pairs, operating at a bandwidth of 20 MHz and a sampling rate of 100 Hz. Each participant performed each action 20 times in 100 environments, resulting in 12,000 samples.

    \textbf{(24) Baha’A et al.}~\cite{baha2020dataset} data collected in three different environments: two with line-of-sight settings and one with a non-line-of-sight  setting. Thirty volunteers performed five different actions: falling from a sitting position, falling from a standing position, walking, sitting down and standing up, and picking up a pen from the ground. Each action was performed 20 times. The experimental setup included a single-antenna transmitter and a three-antenna receiver, operating at a frequency of 2.4 GHz with a sampling rate of 320 Hz.

     \textbf{(25) Widar3.0} \cite{1-45zheng2019zero} dataset was collected in three environments: a classroom, a spacious hall, and an office room. In each environment, volunteers performed actions within a 2m $\times$ 2m area, with five designated locations and five orientations at each location. The dataset includes two types of hand gestures. The first type consists of general hand movements, including pushing and pulling, sweeping, clapping, sliding, drawing circles, and drawing zigzags. The second type involves drawing the numbers 0$\sim$9. The first type of action was performed by 16 participants, resulting in 12,000 samples (16$\times$5$\times$5$\times$6$\times$5 instances), while the second type was demonstrated by 2 participants, with 5,000 samples (2$\times$5$\times$5$\times$10$\times$10 instances). Widar3.0 was collected using a transmitter with one antenna and at least three receivers, each with three antennas. The dataset was captured on channel 165 at 5.825 GHz with a sampling rate of 1,000 Hz.

     \textbf{(26) ARIL} \cite{wang2019joint}  dataset was collected in a 3.2m $\times$ 3.2m area within a laboratory, with 16 positions. The actions include 6 hand gestures for potential human-computer interaction applications: hand up, hand down, hand left, hand right, hand circle, and hand cross. One volunteer performed each gesture 15 times at each position. The dataset contains a total of 1,394 valid samples. Data was collected using two USRPx210 devices, each with a single antenna, as transmitter and receiver. The sampling rate was approximately 60 Hz.

     \textbf{(27) WiAR} \cite{guo2019wiar} dataset was collected in three indoor environments: one empty room, one meeting room, and one office, as well as two outdoor environments, including a playground and a corridor. Ten volunteers performed three categories of actions: upper body activities, lower body activities, and whole-body activities, with 16 different actions. Each action was performed 30 times. The experimental setup included a transmitter with a single antenna and one receiver with three antennas, spaced 4 meters apart. The participants performed the actions at the center of the setup.

  \textbf{(28) SAR} ~\cite{brinke2019dataset,2-161brinke2019scaling} dataset was collected in the living room of an apartment. A total of 9 volunteers participated in experiments over 6 days. During the first 3 days, 3 different volunteers participated each day, while 2 volunteers participated repeatedly over the last 3 days. During the first 3 days, participants performed 6 actions: Clap, Walk, Wave, Jump, Sit, and Fall. In the last 3 days, participants performed the same actions except for Jump, resulting in a total of 5 actions. Each action was performed continuously for 5 seconds and repeated 50 times. The data collection setup consisted of a transmitter and a receiver, each equipped with three antennas, spaced approximately 2.5 meters apart. The sampling rate was 20 Hz.

     \textbf{(29) FallDeFi}~\cite{palipana2018falldefi}, designed for fall detection, was collected in five environments: a kitchen, bedroom, corridor, laboratory, and bathroom. Three participants performed fall actions and other activities such as walking, jumping, sitting down, and standing up, with each action lasting for 10 seconds and 1153 samples in total. Each environment featured one transmitter and one receiver, with varying placement distances between 4 and 9 meters depending on the setting. To introduce environmental variations, data was collected on different days, and the transmitter's position was adjusted during the experiments. All experiments were conducted at 5.2 GHz with a sampling rate of 1000 Hz and a bandwidth of 20 MHz.

     \textbf{(30) SignFi} \cite{ma2018signfi} dataset was collected in two environments: a laboratory and a home. The lab, measuring 13m $\times$ 12m, the distance between the AP and the STA was 230 centimeters. While the home environment, with dimensions of 4.11m $\times$ 3.86m, offered a shorter distance of 130 centimeters between the AP and STA. 5,520 instances were recorded in the lab, while the home environment provided 2,760 instances. The data involved 276 different sign language gestures, encompassing movements of the head, arm, hand, and fingers. The gestures were performed by 5 different users, and the users were instructed to perform each sign gesture repeatedly, with 20 instances for the lab environment and 10 instances for the home environment, resulting in a total of 8280 gesture instances. The experiments were conducted at 5 GHz with a bandwidth of 20 MHz.

  \textbf{(31) CrossSense}~\cite{1-50zhang2018crosssense}  dataset was collected in three scenarios: a hall entrance, a narrow corridor, and a room, with the participation of 100 volunteers. For gait recognition, each volunteer walked 20 times in the same direction in each scenario, resulting in a total of 6,000 samples. For action recognition, five predefined positions were set in each scenario. At each position, each volunteer performed 40 different actions (e.g., pull, kick, throw), repeating each action 10 times. This resulted in a total of 600,000 samples. The experimental setup included one transmitter and one receiver, both equipped with three antennas. The devices operated at 5 GHz with a sampling rate of 1,000 Hz. 

\textbf{(32) Yousef et.al. 2017}~\cite{yousefi2017survey}dataset was collected in an indoor office area where the WiFi transmitter and receiver were positioned 3m apart in a line-of-sight  condition. Six volunteers performed six different activities: lying down, falling down, walking, running, sitting down, and standing up. Each activity was performed 20 times, yielding 720 samples, and each activity was recorded for 20 seconds. Additionally, video recordings were made during the experiments to help label and annotate the activities in the dataset. The sampling rate was 1000 Hz.

\textbf{(33) OctoNet}~\cite{yuanoctonet2025} is collected in three indoor environments (office, laboratory, and living room) within a 4.1m $\times$ 4.1m area. It includes 41 participants performing 62 daily activities. Participants repeatedly performed each assigned activity following standardized instructions, including both structured routines and freestyle movements. For Wi-Fi sensing, one transmitter with a single antenna operates at 5.18 GHz with 40 MHz bandwidth. Four dual-antenna receivers are arranged in a rectangular layout, forming eight Wi-Fi links (1 Tx $\times$ 4 Rx). The CSI sampling rate is 75.62 Hz, yielding approximately 27.35 million synchronized frames. OctoNet further integrates 12 synchronized modalities and provides high-fidelity 3D skeletal keypoints captured by a motion-capture system, enabling research on multimodal learning, cross-modal tasks, and 3D human pose estimation.

\textbf{(34) NTU-Fi HumanID}~\cite{2-130wang2022caution} dataset was collected in a laboratory and a cubic office. Twenty volunteers, including 12 males and 8 females aged between 20 and 28, participated in the experiments, with 15 designated as legal users and the remaining as illegal intruders. Volunteers walked individually through the testing area, and their gait was captured using WiFi signals. Two TP-Link N750 routers were used: one as a transmitter with a single antenna, operating in 802.11n AP mode at 5 GHz with a 40 MHz bandwidth, and the other as a receiver with three antennas.

 \textbf{(35) GaitID}~\cite{2-157zhang2020gaitid} dataset is designed for gait recognition tasks. Data was collected in two scenarios: a hall and a discussion room. Volunteers walked along four predefined tracks, each with two walking directions (starting from both ends of each track). In the hall, 10 volunteers participated, with each walking 50 times in each direction. In the discussion room, 3 volunteers repeated each track 25 times in each direction. A total of 4,600 samples were collected. The data collection setup included one transmitter and six receivers, forming six Wi-Fi links. The sampling rate, bandwidth, and center frequency were not reported.

 \textbf{(36) XRFv2}~\cite{lan2025xrf} dataset is designed for continuous action localization and action summarization. It was collected in three distinct indoor settings—a study room, a dining room, and a bedroom. In each environment, 16 volunteers performed continuous actions spanning 30 categories. The dataset comprises 853 continuous action sequences, each lasting approximately 80 seconds, making it suitable for continuous action detection, localization, and summarization tasks. The data collection setup utilized one transmitter and three receivers with a sampling rate of 200 Hz. In addition to Wi-Fi measurements, participants wore six IMU sensors placed on an earphone, glasses, two phone pockets, and two wrist-worn watches. Furthermore, XRFv2 provides pose annotations, supporting multimodal fusion and cross-modal research.

 \textbf{(37) Person-in-WiFi 3D}~\cite{yan2024person} dataset was collected in three environments: an office, a classroom, and a corridor, each within a 4m $\times$ 3.5m area. Seven participants freely performed eight actions: reaching out, raising hands, bending over, stretching, sitting down, lifting legs, standing, and walking. In addition to scenarios with a single participant, there were also cases where 2, 3, or 4 participants simultaneously performed actions within the scene. Each clip was recorded for 40 seconds, resulting in a total of 456 40-second recordings. The data collection setup included one transmitter with a single antenna and three receivers, each with three antennas, arranged in a quadrilateral layout. The sampling rate was 200 Hz.

\textbf{(38) MetaLoc}~\cite{2-125gao2023metaloc} dataset was designed for indoor localization and collected in a 12 m × 5 m hall and a 10 m × 8 m laboratory. In each environment, Wi-Fi fingerprints (RSS and CSI) were recorded at 90 grid points. The setup included three transmitters: an ASUS RT-AC86U, a TP-Link TL-WR885N, and a TP-Link TL-WR886N, and a Nexus-5 smartphone as the receiver. The system operated at 5 GHz with a bandwidth of 20 MHz.

 \textbf{(39) Chen and Chang}~\cite{2-121chen2022few} released a dataset used for WiFi fingerprinting localization tasks. Data was collected in two different environments: a conference room and a cubicle office. In the conference room, data was collected at 16 locations, with 600 samples recorded at each location. In the cubicle office, data was collected at 18 locations, with 500 samples recorded at each location. The data acquisition system in the conference room consisted of one fixed-location transmitter-receiver pair, each equipped with two antennas, with a distance of 6 meters between them. In the cubicle office, the data acquisition system consisted of two fixed-location transmitter-receiver pairs, each equipped with two antennas. The number of volunteers participating in data collection, sampling rate, bandwidth, and center frequency were not reported.

\textbf{(40) The SDP dataset}~\cite{tony-sdp-2024} is an industrial-grade Wi-Fi sensing dataset tailored for person detection and counting tasks, featuring realistic deployments in both home and office scenarios. In the home setting, two scenarios are included, each consisting of three rooms—one parlor equipped with an access point (AP) and two bedrooms with replay devices. Each device is equipped with two antennas and operates at 5GHz using the 802.11ax protocol, with a sampling rate of 20Hz and bandwidths of 80MHz or 160MHz. Up to two people may be present and moving in each room, and a sweeping robot is introduced as an additional source of interference. The office setting includes two scenarios: the first features four rooms, each with an AP and four-antenna devices, operating on the 5GHz band with 802.11ac and 20MHz bandwidth; the second involves a corridor with three APs placed at the left, center, and right, and four adjacent rooms without APs. This scenario captures up to two occupants per room and includes varying sampling rates—5Hz, 20Hz, 30Hz, 50Hz, and 100Hz—to explore the effect of temporal resolution.

\textbf{(41) DASECount} ~\cite{2-36hou2022dasecount} dataset, designed for crowd counting, was collected in two environments: an office and a lecture hall, each with both line-of-sight and non-line-of-sight configurations, resulting in four different environmental settings. Volunteers, ranging from 0 to 8 people, performed three types of motion: static (seated with free actions like eating or typing), dynamic (random walking), and mixed (unrestricted activities including walking and sitting). Each sampling sequence was recorded for 5 minutes. The experimental setup included a two-antenna transmitter and a three-antenna receiver, operating at 2.4 GHz with a bandwidth of 40 MHz and a sampling rate of 100 Hz.

\textbf{(42) WiFiTAD}~\cite{liu2025wifi} is a dataset specifically designed for temporal action detection. It was collected in a single office with size of $7m \times 12m \times 2.5m$, where participants continuously performed a sequence of seven actions between the transmitter and receiver, including walking, running, jumping, waving, falling, sitting, and standing. The dataset comprises a total of 553 action sequences, each with an average duration of 85 seconds. Data were collected from 3 volunteers using a laptop equipped with an Intel 5300 NIC, configured with one transmitter and one receiver, each having a single antenna, with a sampling rate of 100Hz.

 \textbf{(43) CSI-Bench}~\cite{zhu2025csi} is a comprehensive, in-the-wild Wi-Fi sensing benchmark that integrates multiple classification-oriented sub-datasets to support a wide range of sensing tasks. These include fall detection (6 environments, 17 participants, 6,700 samples), breathing detection (3 environments, 3 users, 100,000 samples), motion source recognition (10 environments involving 13 humans, 20 pets, a robot, and a fan), room-level localization (6 environments, 8 users), user identification (6 users), activity recognition (5 activity classes), and proximity recognition (4 distance levels). A key strength of CSI-Bench lies in its emphasis on realism: data are collected using commercial Wi-Fi edge devices across 26 diverse indoor environments with 35 real users. Spanning over 461 hours of effective recordings, CSI-Bench captures rich signal variations under natural, uncontrolled conditions, making it a valuable resource for evaluating the generalization, robustness, and scalability of Wi-Fi sensing algorithms.

 \textbf{(44) CSI-Net}~\cite{wang2018csi} is a multi-task Wi-Fi sensing dataset collected in a single indoor environment ($5m\times6m$) using one transmitter and one receiver, each with three antennas, and a sampling rate of 100Hz. It includes four sub-tasks: (1\&2) User Identification and Biometrics Estimation, involving 30 participants who remained stationary for 100 seconds while their CSI signals were recorded at a distance of 1.6m; additional biometric attributes such as body fat and muscle rate were also provided; (3) Sign Language Recognition, in which one participant performed 10 American Sign Language digits (0–9), each lasting about 60 seconds, with the transmitter-receiver distance set to 0.6m; and (4) Fall Detection, where the same participant simulated falls at five different locations in the room, with each fall motion lasting 30 seconds and a 3.0m separation between transmitter and receiver.

\section{Challenges and Future Directions}\label{sec:future}

\subsection{Data, Data, Data}

The scaling law, which involves training large-scale models with large-scale datasets, has proven its success since the advent of AlexNet~\cite{krizhevsky2012imagenet}, achieving remarkable breakthroughs in fields such as computer vision (CV), natural language processing (NLP), robotics, and science. Large-scale datasets like ImageNet~\cite{deng2009imagenet}, COCO~\cite{lin2014microsoft}, and Kinetics~\cite{kay2017kinetics}
have significantly accelerated advancements in these areas. In contrast, the field of Wi-Fi human sensing has not progressed as rapidly, despite the widespread adoption of deep learning methods. A key reason lies in the limited scale and diversity of available Wi-Fi sensing datasets.


Recent initiatives are actively narrowing the data gap; for instance, CSI-Bench~\cite{zhu2025csi} achieves a milestone in realism with 461 hours of recordings, XRF55~\cite{wang2024xrf55} significantly expands the sample count to 42,900, and OctoNet~\cite{yuanoctonet2025} comprises 12 heterogeneous modalities. Despite these advancements in data duration and environmental complexity, Wi-Fi sensing datasets still lack the vast scale of CV benchmarks like ImageNet or Kinetics. The gap remains evident not only in raw sample counts but also in the breadth of action categories and participant demographics, emphasizing the need for continued expansion in data diversity.

Collecting and annotating Wi-Fi sensing data is inherently time-consuming and labor-intensive, creating a data-scarcity bottleneck for deep learning. To address this, synthetic data generation has emerged as a promising solution. By leveraging high-fidelity 3D modeling and computational electromagnetics, researchers can simulate diverse human-environment interactions. For example, by placing dynamic 3D human models into various virtual environments and employing ray-tracing algorithms to emulate signal propagation, realistic Wi-Fi CSI data can be synthesized. This approach provides the massive datasets required to train large-scale foundation models, potentially revolutionizing the scalability of Wi-Fi sensing. However, several open challenges, such as the domain gap between simulated and physical signals, remain to be addressed.


 
\begin{enumerate}
    \item How to model the propagation of signals through people and environments? Can methods like ray-tracing and Fresnel zone models effectively simulate Wi-Fi signals' interactions with objects and people? How to enable modern approaches, such as diffusion models~\cite{ho2020denoising}, Variational Autoencoders (VAEs)~\cite{kingma2013auto}, Neural Radiance Fields (NeRF)~\cite{mildenhall2021nerf}, or 3D Gaussian Splatting~\cite{kerbl20233d} learn probabilistic mappings of signal propagation patterns from observed data?

    \item How to construct diverse environments for data generation? Should we rely on manual modeling of environments, even though it is resource-intensive? Or should we leverage open-source tools like Unity, Unreal Engine, or Infinigen Indoors~\cite{raistrick2024infinigen}, which provide scalable solutions for generating dynamic, realistic environments tailored to Wi-Fi sensing scenarios?

    \item How to synthesize diverse human body models?  Is it more effective to use game engines like Unity or Unreal Engine to create diverse human models? Alternatively, could tools such as SMPL~\cite{loper2015smpl} or MANO~\cite{romero2022embodied} be utilized to extract human body/hand models from large-scale video datasets for synthetic data generation?

    \item How to bridge the gap between synthetic and real data? Could Generative Adversarial Networks~\cite{goodfellow2014generative} help enhance the realism of synthetic features or adapt synthetic data to better match real-world characteristics? What role could domain adaptation and domain-invariant feature learning play in aligning synthetic and real data to improve model performance?

\end{enumerate}

Addressing these challenges offers a pathway to creating realistic, diverse, and scalable synthetic Wi-Fi sensing datasets. By bridging the data gap, the Wi-Fi sensing field could achieve progress comparable to the rapid advancements seen in video, image, and text recognition, ultimately unlocking new frontiers for this promising domain. We consider a strategic roadmap for future research in synthetic data generation:

\begin{itemize}
    \item Short-term Goals (1–2 years): The primary focus should be on enhancing signal realism. Leveraging advanced generative models, such as Diffusion Models or Generative Adversarial Networks, researchers can simulate more authentic multi-path effects and hardware-induced noise. The prioritized challenge lies in establishing a precise ``signal-to-physical-entity" mapping, ensuring that synthesized CSI samples are indistinguishable from real-world data in both the frequency and time domains.

    \item Long-term Goals (3–5 years): The ultimate objective is to realize large-scale, multi-environment closed-loop simulation systems. This involves developing sophisticated simulation engines capable of automatically generating complex 3D indoor scenes and simulating dynamic human-environment interactions. This leap from ``single-sample synthesis" to ``full-scene simulation" will be critical. The core challenge remains overcoming the ``Sim-to-Real" domain gap, ensuring that models trained entirely on simulated data can be seamlessly deployed in physical hardware environments.
\end{itemize}

\subsection{Wi-Fi Sensing Foundation Model Pre-Training}\label{sec:wifi-foundation-model}

Foundation models in vision and language, such as CLIP~\cite{radford2021clip}, GPT~\cite{brown2020gpt3}, and BERT~\cite{devlin2019bert}, have demonstrated remarkable success in adapting to diverse downstream tasks. Similarly, Wi-Fi sensing encompasses a wide range of applications, including human activity recognition, indoor localization, presence detection, and pose estimation. These varied tasks make the development of a Wi-Fi sensing foundation model both promising and necessary. However, training such a model raises several critical open questions that must be addressed:

\begin{enumerate}
    \item Should we utilize traditional architectures like CNNs and RNNs, or opt for Transformers~\cite{waswani2017attention}, which have demonstrated strong performance across various domains? Alternatively, emerging architectures such as Mamba~\cite{gu2023mamba}, known for their smaller parameter sizes, reduced memory consumption, and faster inference speeds, may also be viable options. Should we further tailor the architecture design to cater to different computational platforms, optimizing for specific hardware constraints and efficiency requirements?

    \item How can unlabeled real-world data be incorporated effectively? While large-scale synthetic labeled data and small amounts of labeled real-world data may be used for training, how can we leverage large-scale unlabeled real-world data in the pre-training of the foundation model? Should this real-world data be transmitted to a central server for processing, or should edge computing be adopted to process data locally? Additionally, how can we coordinate with a large number of real-world users to ensure efficient data collection and processing while respecting privacy and computational constraints?

    \item What self-supervised learning strategy is most suitable? Self-supervised learning can effectively leverage unlabeled data, but how should we design the proxy tasks? Should we employ contrastive learning strategies that bring positive sample pairs closer while pushing negative pairs apart? Or would a masking-then-reconstruction approach, as used in models like BERT~\cite{devlin2019bert} or MAE~\cite{he2022masked}, be more appropriate for Wi-Fi sensing? Additionally, are there domain-specific proxy tasks that could better capture the unique spatiotemporal characteristics of Wi-Fi signals?

    \item Wi-Fi sensing is significantly influenced by environmental factors, human body factors, and device configurations. Incorporating such information could enable the foundation model to reason more effectively about environmental changes and adapt to new scenarios. Should this integration be achieved through explicit embeddings of environmental parameters, human body factors, and device configurations, or should the model be designed to infer these characteristics implicitly from the data? 
    
\end{enumerate}

Addressing these questions will be critical for developing a robust and generalizable Wi-Fi sensing foundation model that can serve as a backbone for various downstream applications. For the pre-training of large-scale Wi-Fi sensing models, we outline the following roadmap:
\begin{itemize}
    \item Short-term Goals (1–2 years): The primary objective is to construct task-agnostic universal feature extractors. By leveraging existing medium-scale datasets, researchers can employ self-supervised learning strategies, such as Masked Signal Modeling, to learn fundamental feature representations that are robust against device heterogeneity.

    \item Long-term Goals (3–5 years): The ultimate aim is to train perception foundation models with billions of parameters. By integrating ultra-large-scale unlabeled real-world data with synthetic datasets, the model will achieve powerful zero-shot generalization capabilities. The key challenges involve addressing the massive computational costs associated with processing large-scale CSI data, as well as coordinating globally distributed devices for pre-training, while strictly adhering to user privacy protocols.
\end{itemize}

\subsection{When Wi-Fi Sensing Meets Large Multimodal Models}\label{sec:wifi-meets-llms}

Beyond training Wi-Fi foundation models from scratch, as discussed in Section~\ref{sec:wifi-foundation-model}, another promising research direction is leveraging existing large multimodal models by fine-tuning them to process Wi-Fi sensing data. Large multimodal models, such as those trained on diverse combinations of images, text, and audio, have demonstrated impressive adaptability across various domains. By aligning Wi-Fi sensing data with these models, researchers may unlock new possibilities for enhanced understanding and application of Wi-Fi sensing technologies. However, integrating Wi-Fi data into such models raises several open questions and challenges:

\begin{enumerate}
    \item What fine-tuning strategy is most effective? Fine-tuning large multimodal models requires balancing computational efficiency with task-specific performance improvements. Techniques like LoRA~\cite{hu2022lora}, adapters~\cite{rebuffi2017learning}, or prompt-based fine-tuning~\cite{liu2021pretrain} are popular in the NLP and vision domains. Should one of these strategies be adopted for Wi-Fi sensing, or is there a need for a novel approach tailored to the unique characteristics of Wi-Fi data? Moreover, how can we efficiently fine-tune large models with limited computational resources while ensuring high adaptability?


    \item How can Wi-Fi data be aligned with the modalities handled by large multimodal models (LMMs)? Wi-Fi sensing data typically exists in time-series formats, which differ significantly from the visual or textual tokens processed by LMMs. Building on proven strategies such as converting CSI into spectrograms or image-like representations~\cite{zhang2023imgfi,2-1jiao2024wisda}, and inspired by cross-modal translation works like WiFi2Radar~\cite{1-33Wifi2Radar}, it is evident that aligning Wi-Fi with more `visual' modalities is a viable path for leveraging LMMs. However, while these transformation techniques have shown success in specific tasks, they have yet to become a mainstream standard for universal LMM integration. This raises a critical research question: should we continue to adapt Wi-Fi data to fit visual architectures, or is it more effective to retain its native time-series format and develop specific alignment techniques (e.g., cross-modal contrastive learning) tailored for LMMs? Furthermore, identifying which pre-processing or feature extraction methods are essential to bridge the semantic gap between Wi-Fi signals and existing modalities remains a key bottleneck for achieving optimal performance in multimodal sensing.

    \item How should domain-specific knowledge be incorporated into the model? Wi-Fi sensing encompasses a wealth of domain-specific knowledge, including signal propagation principles, communication protocols, e.g. IEEE 802.11 bf, and established research findings. Should this information be used to fine-tune models before fine-tuning on Wi-Fi signals, or would integrating Retrieval-Augmented Generation (RAG)~\cite{lewis2020retrieval} as an external database allow the model to access Wi-Fi knowledge on demand? How can this domain knowledge improve interpretability, efficiency, and overall performance?

    \item What role does multimodal fusion play in Wi-Fi sensing? In scenarios where a multimodal sensing setup is available, such as Wi-Fi transceivers co-deployed with cameras or IMUs, multimodal fusion plays a vital role in enhancing performance, robustness, and sensing coverage. Large multimodal models excel at integrating these diverse data sources by aligning them in a shared semantic space together with techniques like ALBEF~\cite{li2021align}, Cross-Attention Mechanism~\cite{chen2024x}, and Mixture of Experts~\cite{yun2024flex}. However, achieving seamless fusion requires advanced techniques to address modality-specific noise, temporal synchronization, and the high computational overhead associated with processing multi-source data.

\end{enumerate}

Addressing these challenges will advance the integration of Wi-Fi sensing with large multimodal models and push the boundaries of their application across various tasks and environments. This direction holds great potential for accelerating progress in Wi-Fi sensing by leveraging the capabilities of state-of-the-art AI models. To facilitate the integration of Wi-Fi sensing with LMMs, we envision the following roadmap:

\begin{itemize}
    \item Short-term Goals (1–2 years): The immediate priority is to establish efficient semantic alignment mechanisms between Wi-Fi signals and other modalities like vision or text. Drawing inspiration from cross-modal translation frameworks such as WiFi2Radar, Wi-Fi signals can be transformed into LMM-compatible tokens or spectrograms. Research should prioritize resolving the spatio-temporal misalignment between Wi-Fi time-series data and visual imagery.

    \item Long-term Goals (3–5 years): The long-term objective is to develop native sensing agents with inherent wireless perception capabilities. These agents will enable LMMs to perform direct logical reasoning based on Wi-Fi signals, facilitating complex, real-world tasks. The fundamental challenge lies in bridging the semantic gap between the low-level physical properties of wireless signals and the high-level reasoning logic of large models.
\end{itemize}

\subsection{System Deployment}\label{sec:system-deployment} 

As Wi-Fi sensing technologies advance and large multimodal models are integrated into Wi-Fi sensing applications, an essential next step involves addressing system deployment. Typically, this deployment could follow a hierarchical `Cloud-to-Edge' pipeline: initially, a Wi-Fi sensing model is pre-trained on extensive data in the cloud to learn generalizable representations. This model is then compressed and transmitted to resource-constrained edge devices (e.g., routers), where it performs lightweight on-device adaptation to align with the unique multipath profiles and user behaviors of the target environment. However, several critical questions arise when considering how to effectively and efficiently deploy such systems in real-world settings:

\begin{enumerate}

\item Efficient Inference: Deploying large models on COTS chips (e.g., ESP32) requires handling Wi-Fi's frequency-domain sparsity. The core research question is: How to design compression techniques such as structure-aware pruning~\cite{han2015learning}, quantization~\cite{jacob2018quantization}, and knowledge distillation~\cite{hinton2015distilling} that identify and retain only those subcarriers sensitive to human motion, thereby minimizing computational overhead while preserving sensing granularity?

\item On-device training and fine-tuning. Wi-Fi sensing is uniquely susceptible to environmental shifts; even minor furniture changes alter multipath profiles. Centralized updates often fail to address these site-specific fingerprints. Therefore, a more effective strategy is environment-aware edge adaptation, deploying a foundation model centrally while performing on-demand fine-tuning with techniques like on-device training~\cite{lin2022device}. This allows rapid calibration to local environments and user patterns without massive data backhaul.

\item Sensing-Communication Coexistence. Continuous CSI streaming significantly consumes bandwidth, affecting concurrent communication tasks. The critical challenge is optimizing resource allocation. Specifically, during network congestion, how can the system adaptively downsample CSI rates or perform temporal-spectral compression to maintain real-time sensing (e.g., fall detection) without degrading communication Quality of Service? Future strategies must focus on ISAC-aware scheduling, ensuring sensing pilots and data packets are synchronized based on task urgency.

\item Adversarial Resilience. Sensing models are vulnerable to physical adversarial attacks that manipulate system output~\cite{liu2022physical}. How can we protect systems against channel interference and signal alteration? What approaches, such as robust adversarial training or error correction protocols, can be adopted to mitigate such risks? Additionally, recent advances in Generative AI may offer new solutions for physical-layer security, as seen in DFSS~\cite{wang2025generative}.

\item In consumer scenarios, requiring users to manually collect and annotate data is often infeasible. Practical deployment therefore depends on minimizing this burden through two main strategies. First, implicit feedback loops can leverage user corrections, e.g.,  manually switching a light after a false detection, as supervision signals. Second, cross-modal supervision can use auxiliary sensors, such as smartphone IMUs, to automatically generate labels during the initial adaptation phase, eliminating the need for manual annotation. Despite these strategies, significant challenges remain: feedback- or sensor-derived labels may be noisy or incomplete, privacy concerns can restrict the use of auxiliary sensing modalities, and long-term deployments must contend with concept drift caused by changes in user behavior or the environment. These issues underscore the need for labeling-efficient, privacy-preserving, and user-friendly data collection mechanisms, which continue to be an open research direction for deployable Wi-Fi sensing systems.

\end{enumerate}

Addressing these questions will help define how Wi-Fi sensing services can be deployed and scaled efficiently in real-world applications, ultimately determining the practicality and success of Wi-Fi sensing systems across diverse industries and environments. Regarding the practical deployment and long-term evolution of Wi-Fi sensing, we outline the following roadmap:

\begin{itemize}
    \item Short-term Goals (1–2 years): The focus is on establishing lightweight adaptive deployment mechanisms and standardized sensing interfaces. Research should prioritize low-power, calibration-free adaptive algorithms that enable systems to rapidly adjust to new environments via weak supervision signals (e.g., user-driven corrections). Simultaneously, promoting the early adoption of standards such as IEEE 802.11bf is crucial to unifying CSI extraction interfaces across hardware vendors, thereby lowering the deployment barriers caused by device heterogeneity.

    \item Long-term Goals (3–5 years): The ultimate objective is to realize a large-scale, privacy-preserving ``Sensing-as-a-Service" ecosystem. This involves integrating Wi-Fi sensing into the fundamental infrastructure of smart cities.
\end{itemize}

\section{Conclusion}\label{sec:conclusion}

In this survey, we systematically reviewed and categorized Wi-Fi sensing generalization studies published between 2015 and 2025, following the end-to-end pipeline of Wi-Fi sensing systems—spanning device deployment, signal preprocessing, feature learning, and model deployment. We summarized key techniques proposed in each stage to address the challenges of domain variability. In addition, we provided a comprehensive overview of existing publicly available datasets, highlighting their domain diversity and applicability to generalization research.

By integrating methodological insights with data resources, this survey serves as a practical handbook for researchers aiming to advance the generalizability of Wi-Fi sensing systems. Beyond the current landscape, we also discussed future research directions, including synthetic dataset generation, pretraining large-scale perception models, integration with multimodal foundation models, and deployment-aware continual learning. 

To further promote the development of Wi-Fi sensing, we have initiated the Sensing Dataset Platform (SDP: \textcolor{mypink}{\url{http://www.sdp8.org}}
), an open dataset and model sharing hub designed to foster collaboration between academia and industry, enabling researchers worldwide to contribute and access high-quality resources. Given the rapid pace in this field, we will continuously maintain and update relevant resources at \textcolor{mypink}{\url{https://github.com/aiotgroup/awesome-wireless-sensing-generalization}}.

\ifCLASSOPTIONcaptionsoff
  \newpage
\fi



%


\bibliographystyle{IEEEtran}
\bibliography{references-simple}

\vspace{-20pt}

\begin{IEEEbiography}[{\includegraphics[width=1in,height=1.25in,clip,keepaspectratio]{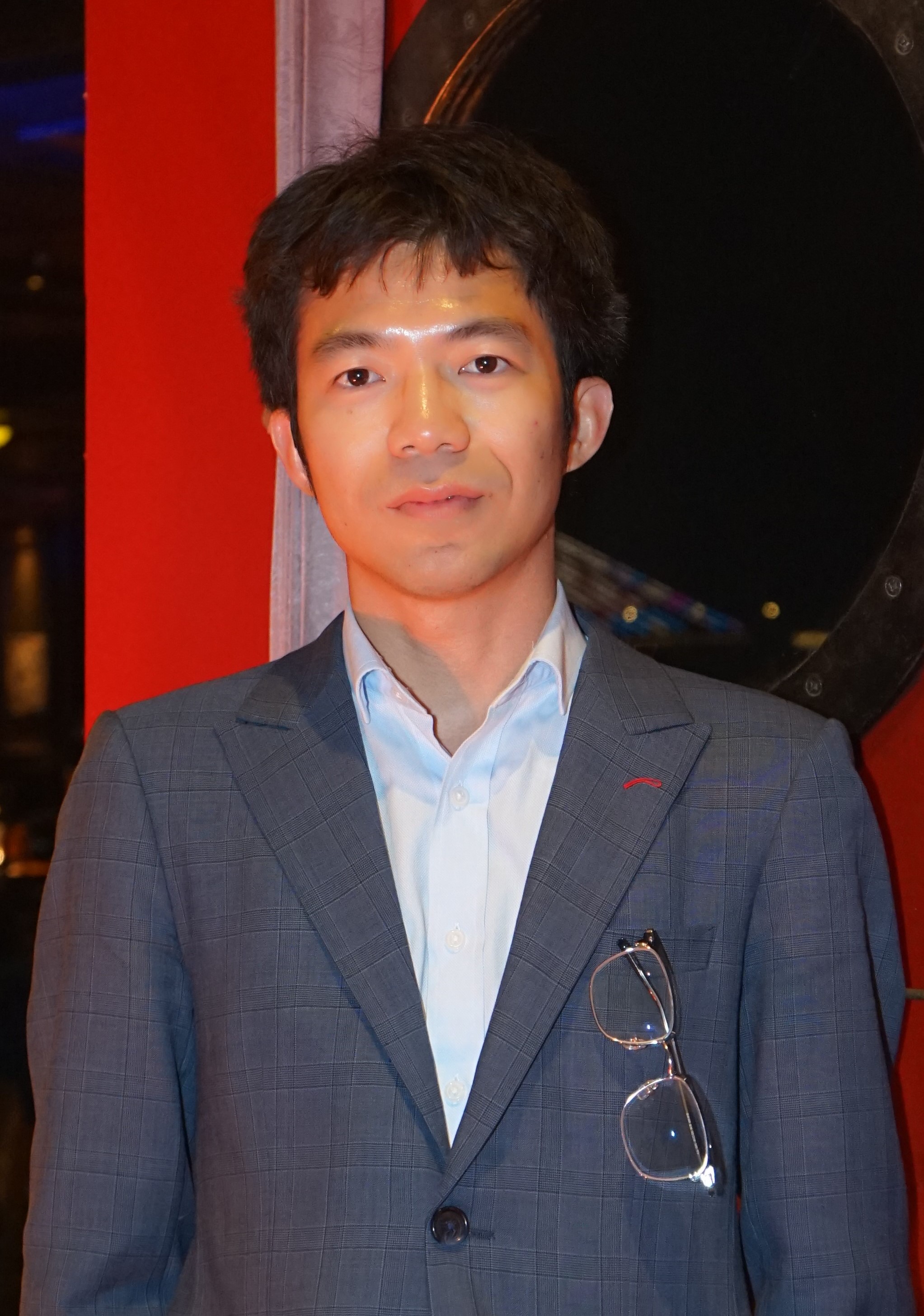}}]
{Fei Wang} (Member, IEEE) received the Ph.D. degree in Computer Science and Technology from Xi’an Jiaotong University, Xi'an, China, in 2020. He was a visiting Ph.D. student with the School of Computer Science, Carnegie Mellon University, Pittsburgh, USA, from 2017 to 2019. He received the 2019 IEEE GLOBECOM Best Paper Award. He is currently an Associate Professor with Xi’an Jiaotong University, Xi'an, China. His research interests include human sensing, mobile computing, and deep learning.
\end{IEEEbiography}


\begin{IEEEbiography}[{\includegraphics[width=1in,height=1.25in,clip,keepaspectratio]{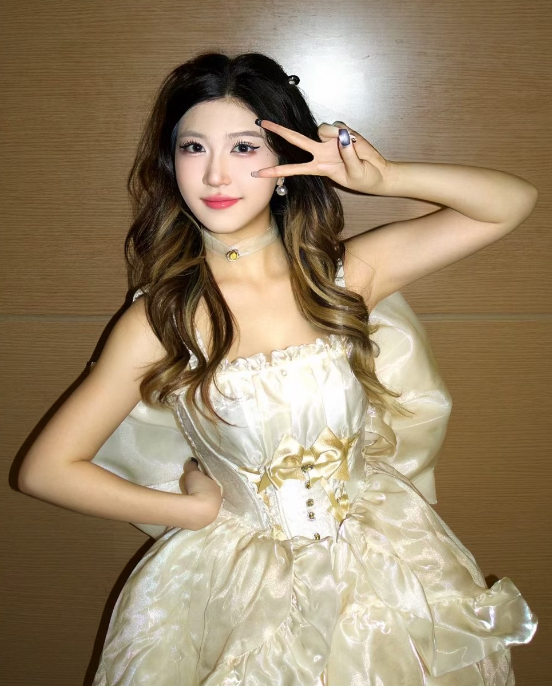}}]
{Tingting Zhang} received the B.E. degree from the School of Information Technology, Xiamen University, Xiamen, China, in 2024. She is currently a master student with the School of Software Engineering, Xi'an Jiaotong University, Xi'an, China. Her research interest is wireless sensing, human-computer interaction, and ubiquitous sensing. 
\end{IEEEbiography}

\begin{IEEEbiography}[{\includegraphics[width=1in,height=1.25in,clip,keepaspectratio]{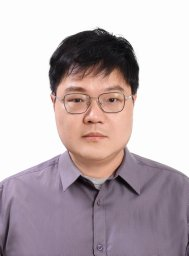}}]
{Wei Xi} (Member, IEEE) received his Ph.D. degree in computer science and technology from Xi'an Jiaotong University in 2014. He is currently a Professor with Xi'an Jiaotong University, Xi'an, China. He is a member of CCF, ACM, and IEEE. His research interests focus on Internet of Things, Wireless Networks, Artificial Intelligence, and Privacy \& Security.
\end{IEEEbiography}

\begin{IEEEbiography}[{\includegraphics[width=1in,clip,keepaspectratio]{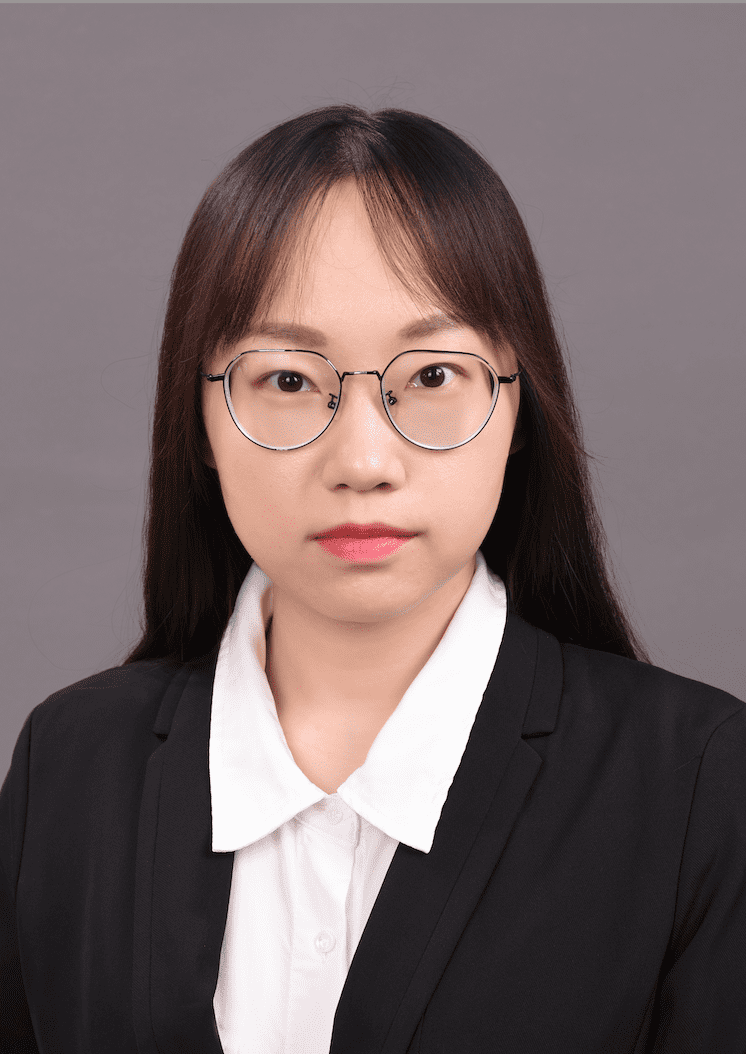}}]
{Han Ding} (Senior Member, IEEE) received the Ph.D. degree in computer science and technology from Xi’an Jiaotong University, Xi’an, China, in 2017.
She is now a Professor with Xi’an Jiaotong University, Xi'an, China. His research interests focus on AIoT, smart sensing, and RFID systems.
\end{IEEEbiography}

\begin{IEEEbiography}[{\includegraphics[width=1in,clip,keepaspectratio]{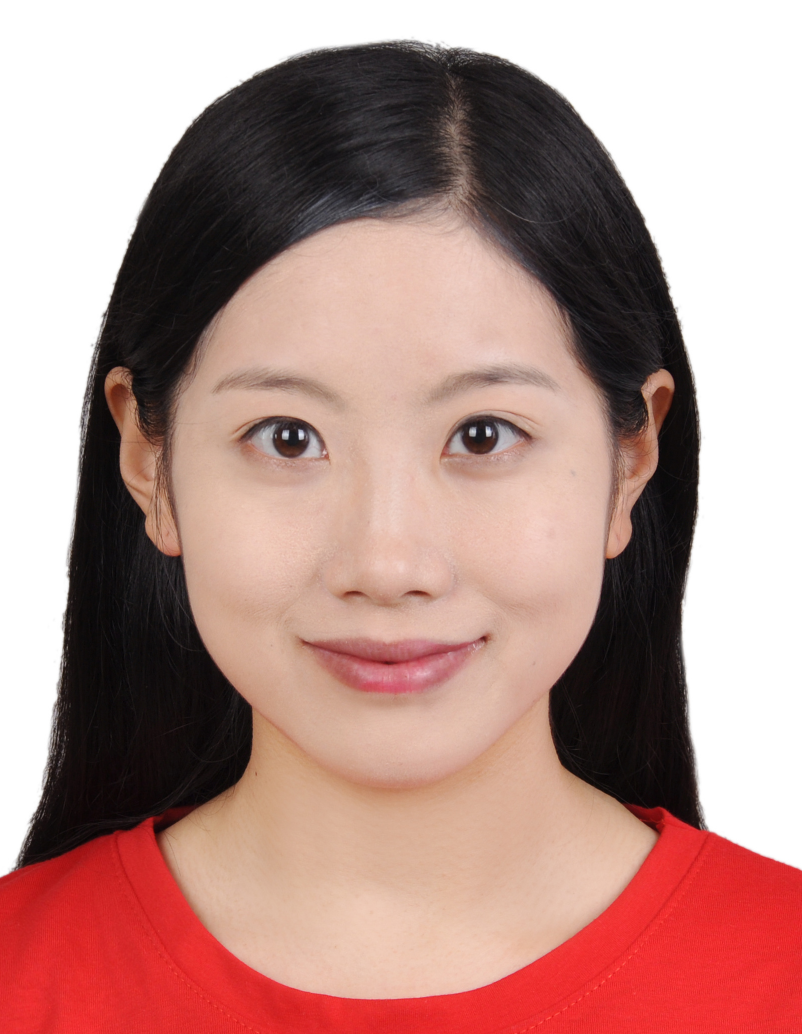}}]
{Ge Wang} (Member, IEEE) is now an Associate Professor at Xi'an Jiaotong University, Xi'an, China. She received her Ph.D. at Xi'an Jiaotong University in 2019. Her research interests include wireless sensor network and mobile computing.
\end{IEEEbiography}

\begin{IEEEbiography}[{\includegraphics[width=1in,clip,keepaspectratio]{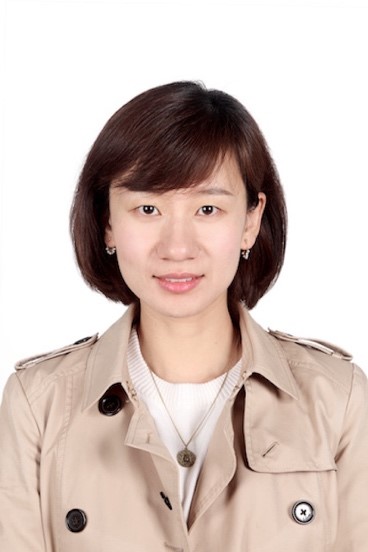}}]
{Di Zhang} is a Ph.D. student at Beijing University of Posts and Telecommunications (BUPT), Beijing 100876, China. She is a reviewer for several journals and a program committee member for international conferences, including the IEEE Journal of Selected Areas in Sensors, the IEEE Internet of Things Journal, and the International Conference on Information Processing and Network Provisioning. Her research interests include integrated sensing and communications (ISAC) and ubiquitous sensing.
\end{IEEEbiography}

\vspace{-20pt}

\begin{IEEEbiography}[{\includegraphics[width=1in,clip,keepaspectratio]{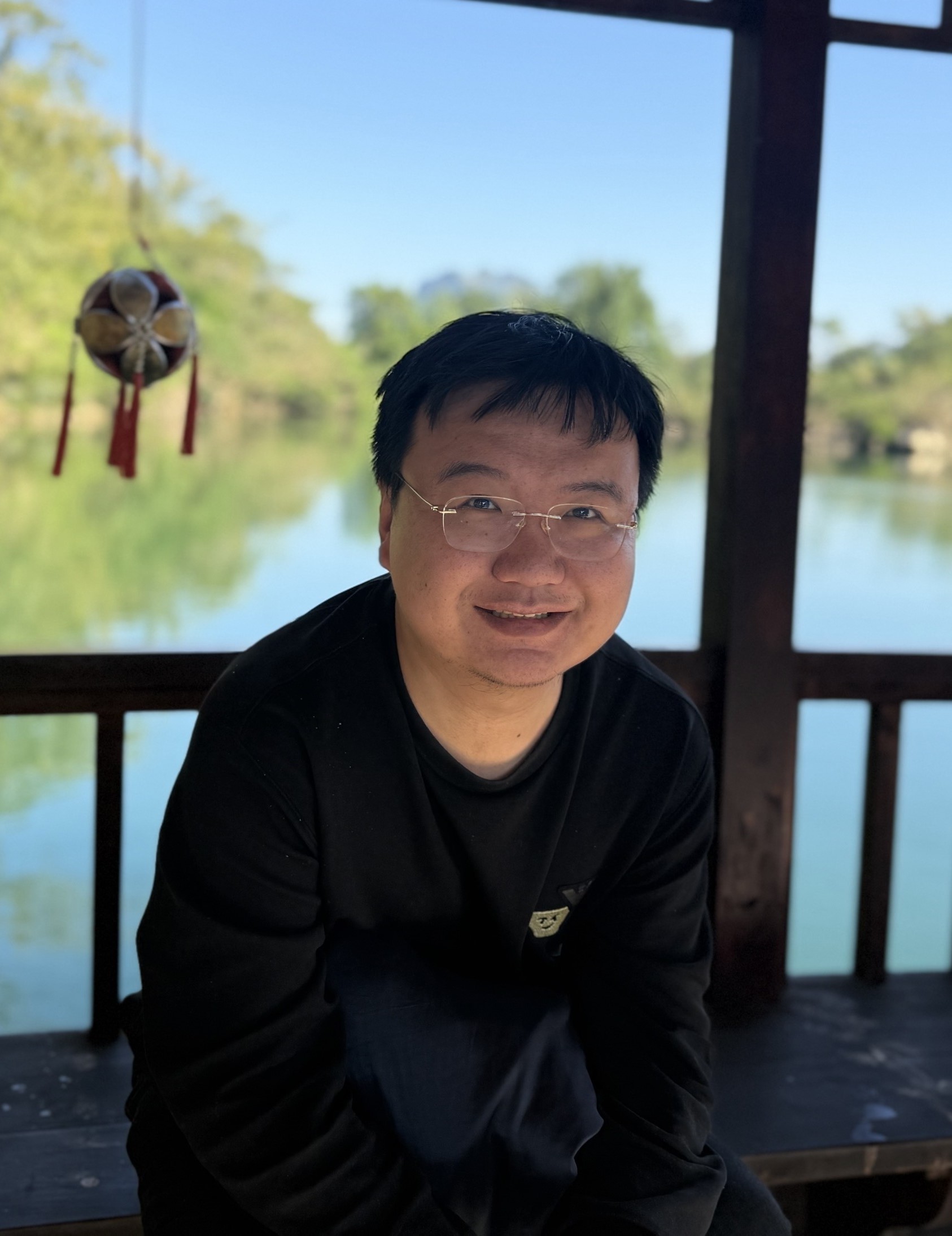}}]
{Yuanhao Cui} (Member, IEEE) is currently an assistant Professor with the School of Information Science and Engineering, Beijing University of Posts and Telecommunications, Beijing, China. Dr. Cui’s research interests lie in the general area of signal processing and wireless communications, and in particular in the area of Integrated Sensing and Communications (ISAC) and Low-Altitude Wireless Network (LAWN). He is the Founding Chair of the IEEE ComSoc Special Interest Group on Low-Altitude Wireless Networks (LAWN-SIG), the founding Secretary of the IEEE ComSoc ISAC Emerging Technology Initiative (ISAC-ETI), and the founding Secretary of the CCF Scientific Communication standing committee. He serves on the editorial board of IEEE Transactions on Mobile Computing, IEEE Vehicular Technology Magazine, IEEE Journal of Internet of Things, and IEEE Journal of Biomedical and Health Informatics. He is a member of the IMT-2030 (6G) ISAC Task Group. He was listed among the World’s Top 2\% Scientists by Stanford University for citation impact from 2023 to 2025. He was a recipient of numerous Best Paper Awards, including the 2025 IEEE Communication Society and Information Theory Society Joint Paper Award, 2024 IEEE Communications Society Asia-Pacific Outstanding Paper Award, 2024 IEEE Globecom Best Paper Award, 2024 IEEE JC\&S Symposium Best Paper Award, 2023 ACM MobiCom Best Paper Award in ISAC, and 2023 IEEE/CIC ICCC 2023 Best Paper Award.
\end{IEEEbiography}

\vspace{-20pt}

\begin{IEEEbiography}[{\includegraphics[width=1in,clip,keepaspectratio]{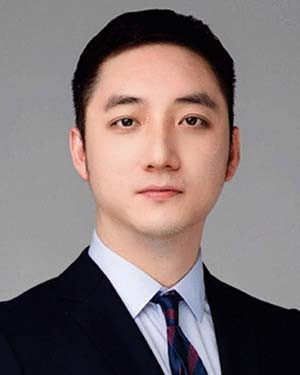}}]
{Fan Liu} (Senior Member, IEEE) received the B.Eng.
and Ph.D. degrees from the Beijing Institute of Technology, Beijing, China, in 2013 and 2018, respectively. He was an Assistant Professor with the Southern University of Science and Technology, Shenzhen,
China, from 2020 to 2024. He held academic positions with the University College London, London,
U.K., as a Visiting Researcher from 2016 to 2018, and
a Marie Curie Research Fellow from 2018 to 2020.
He is currently a Professor with the National Mobile
Communications Research Laboratory, School of Information Science and Engineering, Southeast University, Nanjing, China. His research interests include the general area of signal processing and wireless communications, and in particular in the area of Integrated Sensing and Communications. He is the Founding Academic Chair of the IEEE ComSoc ISAC Emerging
Technology Initiative (ISAC-ETI), the Vice Chair and Founding Member of the
IEEE SPS ISAC Technical Working Group (ISAC-TWG), an Elected Member of
the IEEE SPS Sensor Array and Multichannel Technical Committee (SAM-TC),
an Associate Editor for IEEE TRANSACTIONS ON COMMUNICATIONS, IEEE
TRANSACTIONS ON MOBILE COMPUTING, and IEEE OPEN JOURNAL OF SIGNAL
PROCESSING, and a Guest Editor of the IEEE JOURNAL ON SELECTED AREAS IN
COMMUNICATIONS, IEEE WIRELESS COMMUNICATIONS, and IEEE Vehicular
Technology Magazine. He was a TPC Co-Chair of the 2nd-4th IEEE Joint
Communication and Sensing (JC\&S) Symposium, a Symposium Co-Chair for
the IEEE ICC 2026 and IEEE GLOBECOM 2023, and a Track Co-Chair for the
IEEE WCNC 2024. He is a member of the IMT-2030 (6G) ISAC Task Group.
He was listed among the World’s Top 2\% Scientists by Stanford University
for citation impact from 2021 to 2024, and among the Elsevier Highly-Cited
Chinese Researchers from 2023 to 2024. He was the recipient of numerous Best
Paper Awards, including the 2025 IEEE Communications Society \& Information
Theory Society Joint Paper Award, 2024 IEEE Signal Processing Society Best
Paper Award, 2024 IEEE Signal Processing Society Donald G. Fink Overview
Paper Award, 2024 IEEE Communications Society Asia-Pacific Outstanding
Paper Award, 2023 IEEE Communications Society Stephan O. Rice Prize, and
2021 IEEE Signal Processing Society Young Author Best Paper Award.
\end{IEEEbiography}

\begin{IEEEbiography}[{\includegraphics[width=1in,clip,keepaspectratio]{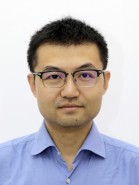}}]
{Jinsong Han} (Senior Member, IEEE) received his Ph.D. degree from Hong Kong University of Science and Technology in 2007. He is now a professor with Zhejiang University, Hangzhou, China. He is a senior member of the ACM and IEEE. His research interests focus on IoT security, smart sensing, wireless and mobile computing.
\end{IEEEbiography}

\begin{IEEEbiography}[{\includegraphics[width=1in,clip,keepaspectratio]{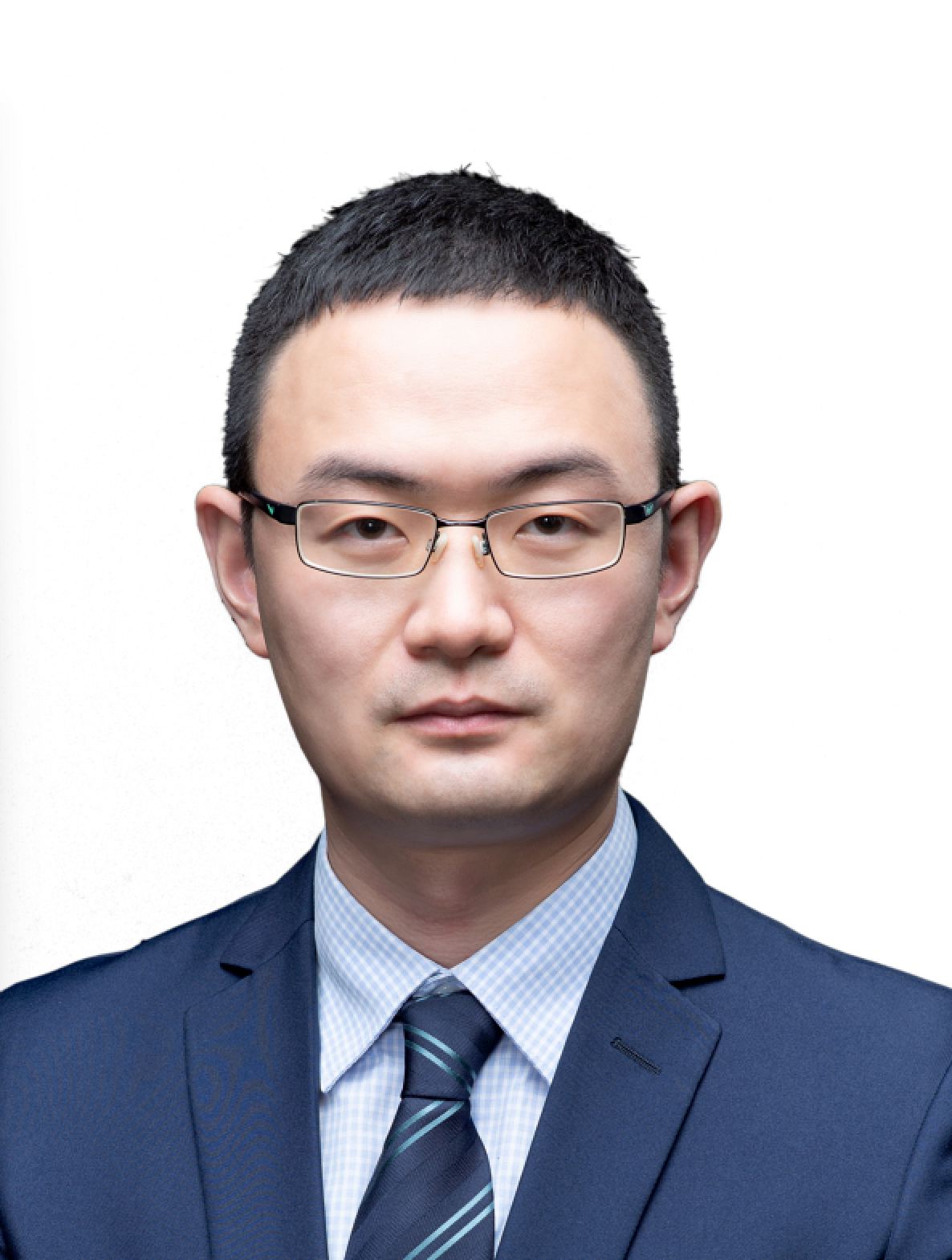}}]
{Jie Xu} (Fellow, IEEE) received the B.E. and Ph.D. degrees from the University of Science and Technology of China. He is currently an Associate Professor (Tenured) with the School of Science and Engineering, the Shenzhen Future Network of Intelligence Institute (FNii-Shenzhen), and the Guangdong Provincial Key Laboratory of Future Networks of Intelligence, The Chinese University of Hong Kong (Shenzhen). His research interests include wireless communications, wireless information and power transfer, UAV communications, edge computing and intelligence, and integrated sensing and communication (ISAC). He was a recipient of the 2017 IEEE Signal Processing Society Young Author Best Paper Award, the IEEE/CIC ICCC 2019 Best Paper Award, the 2019 IEEE Communications Society Asia-Pacific Outstanding Young Researcher Award, and the 2019 Wireless Communications Technical Committee Outstanding Young Researcher Award. He is the Symposium Co-Chair of the IEEE GLOBECOM 2019 Wireless Communications Symposium and the IEEE ICC 2025 Communication Theory Symposium, the workshop co-chair of several IEEE ICC and GLOBECOM workshops, the Tutorial Co-Chair of the IEEE/CIC ICCC 2019/2022, the Chair of the IEEE Wireless Communications Technical Committee (WTC), and the Vice Co-chair of the IEEE Emerging Technology Initiative (ETI) on ISAC. He served or is serving as an Associate Editor-in-Chief of the IEEE Transactions on Mobile Computing, an Editor of the IEEE Transactions on Wireless Communications, IEEE Transactions on Communications, IEEE Wireless Communications Letters, and Journal of Communications and Information Networks, an Associate Editor of IEEE Access, and a Guest Editor of the IEEE Wireless Communications, IEEE Journal on Selected Areas in Communications, IEEE Internet of Things Magazine, Science China Information Sciences, and China Communications. He is a Clarivate Highly Cited Researcher and a Distinguished Lecturer of IEEE Communications Society.
\end{IEEEbiography}

\begin{IEEEbiography}[{\includegraphics[width=1in,clip,keepaspectratio]{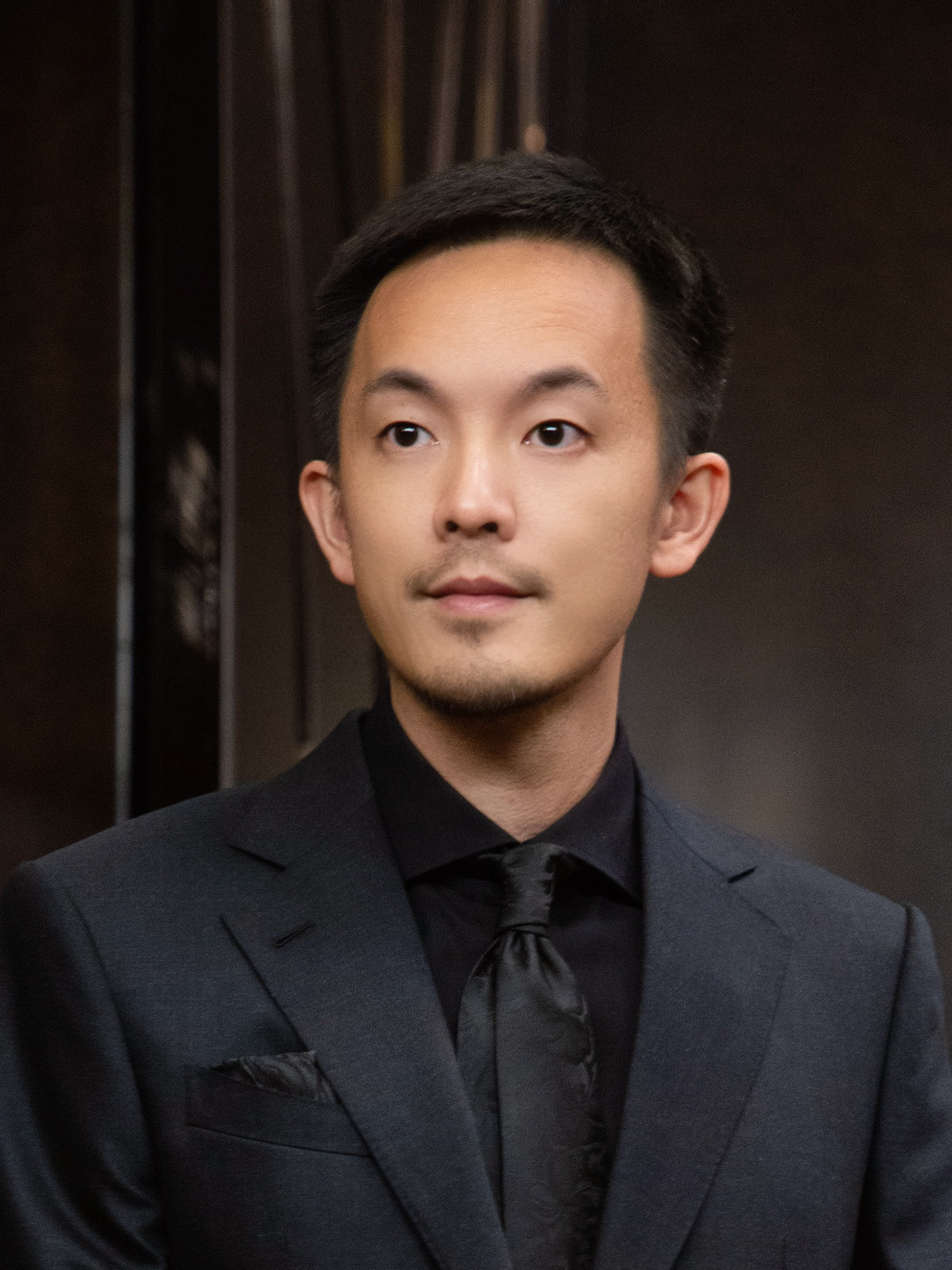}}]
{Tony Xiao Han} (Senior Member, IEEE) is currently a Research Expert and Project Leader with Huawei Technologies Co., Ltd. He received his B.E. degree in Electrical Engineering from Sichuan University, Chengdu, China, and his Ph.D. degree in Communications Engineering from Zhejiang University, Hangzhou, China. He was a Postdoctoral Research Fellow at the National University of Singapore (NUS), Singapore. His research interests include wireless communications, Integrated Sensing and Communication (ISAC), WLAN sensing, and wireless communication standardization.
Dr. Han is a recipient of the 2025 IEEE Standards Association Standards Medallion, and the 2025 IEEE Communications Society \& Information Theory Society Joint Paper Award. He has held several key leadership roles in international standardization bodies and academic communities. He previously served as Chair of the IEEE 802.11 WLAN Sensing Topic Interest Group (TIG) and Study Group (SG). He currently serves as Chair of the IEEE 802.11bf WLAN Sensing Task Group (TG) and Chair of the Wi-Fi Alliance (WFA) Sensing Task Group. In addition, he is the founding Industry Chair of the IEEE ComSoc ISAC Emerging Technology Initiative (ISAC-ETI), Vice Chair of the IEEE Wireless Technical Committee (WTC) Special Interest Group on ISAC, and a Guest Editor of the IEEE Journal on Selected Areas in Communications (JSAC) Special Issue on ``Integrated Sensing and Communications (ISAC).” He has also served as Co-Chair of multiple international workshops, including the IEEE GLOBECOM 2020 Workshop on ISAC.
\end{IEEEbiography}

\end{document}